\documentclass[12pt,draftcls,onecolumn]{IEEEtran}
\usepackage{CJK}
\usepackage{amssymb,amsmath,mathrsfs,amsopn,amsfonts,graphicx,color,amsthm}
\usepackage{algorithmic}
\usepackage{graphicx}
\usepackage{mathtools}
\usepackage{subfig}
\usepackage{float}
\usepackage{multirow}
\usepackage{textcomp}

\newtheorem{remark}{Remark}
\newtheorem{theorem}{Theorem}
\newtheorem{lemma}{Lemma}

\newtheorem{definition}{Definition}
\newtheorem{corollary}{Corollary}
\newtheorem{example}{Example}

\allowdisplaybreaks
\def\A{\mathcal A}
\def\V{\mathcal V}
\def\({\Big(}
\def\){\Big)}

\def\G{\mathcal G}
\def\L{\mathcal L}
\def\H{\mathcal H}

\def\E{\mathbb{E}}
\def\P{\mathbb{P}}
\def\F{\mathscr{F}}
\def\a{\alpha}
\def\b{\beta}
\def\g{\gamma}
\def\ba{\begin{array}}
\def\ea{\end{array}}
\def\ban{\begin{eqnarray*}}
\def\ean{\end{eqnarray*}}
\def\bann{\begin{eqnarray*}}
\def\eann{\end{eqnarray*}}
\def\bd{\begin{description}}
\def\ed{\end{description}}
\def\be{\begin{equation}}
\def\ee{\end{equation}}
\def\bna{\begin{eqnarray}}
\def\ena{\end{eqnarray}}

\def\d{\delta}

\def\ln{\mbox{ln}}

\def\z{\zeta}

\def\ban{\begin{eqnarray*}}
\def\ean{\end{eqnarray*}}
\def\bna{\begin{eqnarray}}
\def\ena{\end{eqnarray}}
\def\bnaa{\begin{eqnarray}}
\def\enaa{\end{eqnarray}}
\def\bann{\begin{eqnarray*}}
\def\eann{\end{eqnarray*}}

\ifCLASSINFOpdf
\else
\fi
\begin{document}
\begin{CJK}{GBK}{song}

\title{Decentralized Online Regularized Learning Over Random Time-Varying Graphs}

%\author{Xiwei Zhang,~Tao Li,~\IEEEmembership{Senior Member,~IEEE}, and Xiaozheng Fu
%\thanks{This work was supported by the National Natural Science Foundation
%of China under Grant 62261136550. \emph{(Corresponding author: Tao Li.)} }
%\thanks{Xiwei Zhang is with the School of Mathematical Sciences, East China Normal University, Shanghai 200241, China (e-mail: xwzhangmath@sina.com).}
%\thanks{Tao Li is with the School of Mathematical Sciences, East China Normal University, Shanghai 200241, China (e-mail: tli@math.ecnu.edu.cn).}
%\thanks{Xiaozheng Fu is with the School of Mathematical Sciences, East China Normal University, Shanghai 200241, China (e-mail: fxz4926@163.com).}}
\author{Xiwei Zhang,~Tao Li,~\IEEEmembership{Senior Member,~IEEE}, and Xiaozheng Fu
\thanks{This work was supported by the National Natural Science Foundation
of China under Grant 62261136550. \emph{(Corresponding author: Tao Li.)} }
\thanks{Xiwei Zhang was with the School of Mathematical Sciences, East China Normal University and now is with the No.2 High School of East China Normal University, Shanghai, 201203, China (e-mail: xwzhangmath@sina.com
).}
\thanks{Tao Li is with the Key Laboratory of Management, Decision and Information Systems, Institute of Systems Science, Academy of Mathematics and Systems Science, Chinese Academy of Sciences,  Beijing 100190, China, and also with School of
Mathematical Sciences, University of Chinese Academy of Sciences, Beijing 100149, China (email: litao@amss.ac.cn).}
\thanks{Xiaozheng Fu is with the School of Mathematics and Statistics,
Ningbo University, Ningbo 315211, China (e-mail: fxz4926@16
3.com).}}


\markboth{Journal of \LaTeX\ Class Files, May~2023}%
{Shell \MakeLowercase{\textit{et al.}}: Bare Demo of IEEEtran.cls for IEEE Journals}



\maketitle

\begin{abstract}
We study the decentralized online regularized linear regression algorithm over random time-varying graphs. At each time step, every node runs an online estimation algorithm consisting of an innovation term processing its own new measurement, a consensus term taking a weighted sum of estimations of its own and its neighbors with additive and multiplicative communication noises and a regularization term preventing over-fitting.
It is not required that the regression matrices and graphs satisfy special statistical assumptions such as mutual independence, spatio-temporal independence or stationarity. We develop the nonnegative supermartingale inequality of the estimation error, and prove that the estimations of all nodes converge to the unknown true parameter vector almost surely if the algorithm gains, graphs and regression matrices jointly satisfy the \emph{sample path spatio-temporal persistence of excitation} condition.
Especially, this condition holds by choosing appropriate algorithm gains if the graphs are \emph{uniformly conditionally jointly connected} and \emph{conditionally balanced}, and the regression models of all nodes are \emph{uniformly conditionally spatio-temporally jointly observable}, under which the algorithm converges in mean square and almost surely.
In addition, we prove that the regret upper bound is $\mathcal O(T^{1-\tau}\ln T)$, where $\tau\in (0.5,1)$ is a constant depending on the algorithm gains.
\end{abstract}


\section{Introduction}
\label{Asec:introduction}
\IEEEPARstart{E}{mpirical}
risk minimization is an important criterion to judge the predictive ability of models in statistical learning. Problems in machine learning are usually ill-posed, and the solutions obtained by using the empirical risk minimization principle are unstable, easily overfitted and usually have large norms. To solve this problem, Poggio, Girosi \emph{et al}. \cite{AEvgeniou}-\cite{APoggio} introduced regularization methods in inverse problems.
Regularization is an effective tool for dealing with the complexity of a model, and its role is to choose a model with less empirical risk and complexity at the same time. In addition, this method can transform the original ill-posed problem into a well-posed one, reduce the norm of estimations of the unknown true parameter vector, and keep the sum of the squared errors small \cite{ATheodoridis}.
Regularization methods are widely used in various research areas, such as classification in machine learning, online state estimation for power systems and image reconstruction \cite{Axue2020}-\cite{Asun2019}. There are two classic regularization methods: Ivanov regularization \cite{Aivanov2002}, which restricts the hypothesis space, and Tikhonov regularization, which restricts certain parameters in the loss function, i.e., adds regularization terms which are also called penalty terms into the algorithm. In learning theory, ridge regression adopts Tikhonov regularization \cite{AVito}.

An offline algorithm usually needs to acquire a finite amount of data generated by an unknown, stationary probability distribution in advance, which heavily relies on the memory of the system especially when the dataset is far too large, while an online algorithm processes infinite data streams that are continuously generated at rapid rates, where the data is discarded after it has been processed, and the unknown data generation process is possibly non-stationary.
So far, the centralized online regularized algorithms have been widely studied in \cite{ACesa}-\cite{AThramp}. In a  centralized algorithm, there is an information fusion center collecting the measurements of all nodes and giving  the global estimation.
In reality, many learning tasks process very large datasets, and thus decentralized parallel processing of data by communicating and computing units in the network is necessary, see e.g. \cite{Ascaman}-\cite{Apredd} and references therein. Besides, if the data contains sensitive private information (e.g. medical and social network data, etc.), they may come from different units in the network, and transmitting all these data subsets to the fusion center may lead to potential privacy risks \cite{ALLB}-\cite{AYSVQ}. Therefore, decentralized learning is needed, which can  improve the efficiency of communication and protect the privacies of users.

At present, the non-regularized decentralized linear regression problems have been widely studied in \cite{Akar2011}-\cite{Awjx}.
Xie and Guo \cite{Axiesiyu1}-\cite{Axiesiyu2} considered the time-varying linear regression with measurement noises, where the cooperative information condition on the conditional expectations of the regression matrices was proposed over a deterministic, undirected and strongly connected graph.
Chen \emph{et al}. \cite{Acheny} proposed a saturated innovation update algorithm for the decentralized estimation under sensor attacks, where the interagent communication is noiseless. They proved that if the communication graph is undirected and fixed, the nodes are locally observable, and the number of attacked nodes is less than half of the total, then all nodes' estimations converge to the unknown true parameter with a polynomial rate.
Wang \emph{et al}. \cite{Awjx} investigated a consensus plus innovation based decentralized linear regression  algorithm over random networks with random regression matrices. Some scholars have also considered both measurement and communication noises among nodes, e.g.
\cite{Akar2012}-\cite{Akar20132} and \cite{Asahu2016}.

Based on the advantages of the decentralized information structure, the online algorithm and the regularization method, we propose a decentralized online regularized algorithm for the linear regression problem over random time-varying graphs. The algorithm of each node contains an innovation term, a consensus term and a regularization term. In each iteration, the innovation term is used to update the node's own estimation, the consensus term is the weighted sum of estimations of its own and its neighbors with additive and multiplicative communication noises, and the regularization term is helpful for constraining the norm of the estimation in the algorithm and preventing the unknown true parameter vector from being overfitted.
Although regularization is an effective method to deal with linear regression problems, it brings essential difficulties to the convergence analysis of the algorithm. Compared with the non-regularized decentralized linear regression algorithm, the estimation error equation of this algorithm contains a non-martingale difference term with the regularization parameter, which cannot be directly analyzed by using martingale convergence theorem as \cite{Alw}.
We no longer require that the sequences of regression matrices and graphs satisfy special statistical properties, such as mutual independence, spatio-temporal independence and stationarity. Compared with the case with i.i.d. data, dependent observations and data contain less information and therefore lead to more unstable learning errors as well as the performance degradation \cite{AZYD}.
Besides, we consider both additive and multiplicative communication noises in the process of the information exchange among nodes. All these challenges make it difficult to analyze the convergence and performance of the algorithm, and the methods in the existing literature are no longer applicable.
For example, the methods in \cite{Akar2011}-\cite{Akar2013} and \cite{Asahu2018} are applicable for the case that the graphs, regression matrices and noises are i.i.d. and mutually independent and it is required that the expectations of the regression matrices be known in \cite{Akar2012}-\cite{Akar20132}.
Liu \emph{et al}. \cite{ALLB} studied the decentralized regularized gossip gradient descent algorithm for linear regression models, where the method is applicable for the case that only two nodes exchange information at each instant. In addition, they require that the graphs be strongly connected and the observation vectors and the noises be i.i.d. and bounded.
Wang \emph{et al}. \cite{Awjx} studied the non-regularized decentralized online algorithm, where the communication noises were not considered, and the method therein is only applicable for the case that the inhomogeneous part of the estimation error equation is a martingale difference sequence.

To overcome the difficulties mentioned above, we develop the nonnegative supermartingale inequality of the estimation error, and further establish the \emph{sample path spatio-temporal persistence of excitation} condition by combining information of the regression matrices, graphs and algorithm gains,
under which sufficient conditions for the convergence of the algorithm are obtained. We prove that if the algorithm gains and the sequences of regression matrices and Laplacian matrices of the graphs jointly satisfy the \emph{sample path spatio-temporal persistence of excitation} condition, then the estimations of all nodes converge to the unknown true parameter vector almost surely.
Furthermore, we give an intuitive sufficient condition of the \emph{sample path spatio-temporal persistence of excitation} condition, i.e., the graphs are \emph{conditionally balanced} and \emph{uniformly conditionally jointly connected}, and the regression models of all nodes are \emph{uniformly conditionally spatio-temporally jointly observable}, which drives the algorithm to converge in mean square and almost surely by the proper choice of the algorithm gains.
%Especially, for the case with Markovian switching graphs and regression matrices, we prove that the \emph{sample path spatio-temporal persistence of excitation} condition holds if the stationary graph is balanced with a spanning tree and regression model is \emph{spatio-temporally jointly observable},  implying that neither \emph{local observability} of each node nor \emph{instantaneous global observability} of all the regression models is necessary.

Historically, Guo \cite{Aguolei} was the first to propose the stochastic persistence of excitation condition for analyzing the centralized Kalman filtering algorithm, which was then refined in \cite{AZJF}. Whereafter, the cooperative information condition on the conditional expectations of the regression matrices over the deterministic connected graph for the decentralized adaptive filtering algorithms was proposed in \cite{Axiesiyu1}-\cite{Axiesiyu2}. The stochastic spatio-temporal persistence of excitation condition on the conditional expectations of the regression matrices and graphs was proposed for the non-regularized  decentralized online estimation algorithm over the random time-varying graphs in \cite{Awjx}.
In this paper, this excitation condition is further weakened to the \emph{sample path spatio-temporal persistence of excitation} condition, which only requires that the infinite series composed of the minimum eigenvalues of the information matrices diverge for almost all sample paths. Meanwhile, we extend the work of \cite{Alw} from distributed averaging to decentralized online learning and generalize the network model of Xie and Guo \cite{Axiesiyu1}-\cite{Axiesiyu2} from deterministic, time-invariant and undirected graphs to random and time-varying digraphs.
Xie and Guo \cite{Axiesiyu1}-\cite{Axiesiyu2} considered tracking time-varying parameters by using constant algorithm gains and obtained the $L_p$-boundedness of the tracking errors. In \cite{Axiesiyu1}-\cite{Axiesiyu2},  the homogeneous part of the estimation error equation is  $L_p$-exponentially stable. Different from \cite{Axiesiyu1}-\cite{Axiesiyu2}, we consider estimating time-invariant parameters. To ensure the strong consistency of the algorithm, we use decaying algorithm gains, which result in that the  homogeneous part of the estimation error equation is not $L_p$-exponentially stable. To ensure not only boundedness but also strong consistency of the algorithm, the decaying rates of the algorithm gains and regularization parameter need to be designed precisely.

We use the regret to evaluate the performance of the decentralized optimization algorithm, which has been investigated in \cite{Aydm} and \cite{AASBedi}-\cite{ARDixit}. Yuan \emph{et al}. \cite{Aydm} studied the non-regularized decentralized online linear regression problem over the fixed graph. Bedi \emph{et al}. \cite{AASBedi} considered the multi-agent stochastic optimization problems with constraints in heterogeneous networks. Dixit \emph{et al}. \cite{ARDixit} studied the decentralized online dynamic optimization problems over deterministic and time-varying graphs. Compared with \cite{Aydm} and
\cite{AASBedi}-\cite{ARDixit}, which assumed the graphs to be deterministic, strongly connected and balanced, we consider the decentralized online regularized linear regression algorithm over the random time-varying graphs, and our techniques are helpful to further study the optimization problems over random time-varying graphs. We prove that the upper bound of the regret is $\mathcal O(T^{1-\tau}\ln T)$, where $\tau\in (0.5,1)$ is a constant depending on the decaying algorithm gains.

The rest of this paper is organized as follows. Section II proposes the algorithm. Section III gives the convergence analysis. Section IV gives some numerical examples. Section V concludes the full paper and gives some future topics.

Notation and symbols:
$\mathbb{R}^n$: the $n$ dimensional real vector space; $\mathbb{R}^{m\times n}$: the $m\times n$ dimensional real matrix space;
$\otimes$: the Kronecker product;
$\text{\textbf{diag}}(A_1,\cdots,A_n)$: the block diagonal matrix with entries being  $A_1,\cdots,A_n$; $\|A\|$: the 2-norm of matrix $A$;
$\lambda_{\min}(A)$: the minimum eigenvalue of real symmetric matrix $A$;
$\lambda_2(A)$: the second smallest eigenvalue of real symmetric matrix $A$;
$A\ge B$: the matrix $A-B$ is positive semi-definite; $A\succeq B$: the matrix $A-B$ is nonnegative;
$I_n$: the $n$ dimensional identity matrix; $\mathbf{O}_{n\times m}$: the $n\times m$ dimensional zero matrix;
$\mathbf{1}_{N}$: the $N$ dimensional vector with all ones;
$\lfloor x\rfloor$: the largest integer less than or equal to $x$;
$\lceil x \rceil$: the smallest integer greater than or equal to $x$;
For a sequence of $n\times n$ dimensional matrices $\{Z(k),k\ge 0\}$, denote
\begin{equation*}
\Phi_Z(j,i)=\begin{cases}
Z(j)\cdots Z(i),   &j\ge i \\
~~~~~~I_n,    &j<i.
\end{cases},~
\prod_{k=i}^jZ(k)=\Phi_Z(j,i).
\end{equation*}

\section{Decentralized Regularized Linear Regression}
Suppose that $x_0 \in \mathbb R^n$ is the unknown true parameter vector. We consider a network modeled by a sequence of random digraphs with $N$ nodes $\{\G(k)=\{\V,\mathcal E_{\G(k)},\A_{\G(k)}\}, k\ge 0\}$, where $\mathcal V$ is the node set, $\mathcal E_{\G(k)}$ is the edge set at instant $k$, and $\A_{\G(k)}$ is the weighted adjacency matrix.
The regression model of node $i$ at instant $k$ is given by
\begin{equation}\label{Ameasurement}
y_i(k)=H_i(k)x_0+v_i(k),~k\ge 0,~i\in \mathcal V,
\end{equation}
where $H_i(k)\in \mathbb R^{n_i\times n}$ is the random regression matrix of node $i$ at instant $k$, $v_i(k)\in \mathbb R^{n_i}$ is the additive measurement noise, and $y_i(k)\in \mathbb R^{n_i}$ is the observation data.

\vskip 0.2cm
\begin{remark}\label{Aexample000}
\rm{The regression model (\ref{Ameasurement}) with random regression matrices over random graphs have been widely studied in \cite{ALLB}, \cite{Akar2011}-\cite{Asahu2016}, \cite{Asahu2018}, \cite{Awjx} and \cite{Azzjf}-\cite{Azhz}. There are various uncertainties in real-word networks, where intermittent sensing failures and packet dropouts may occur at random times \cite{Akar20132}.
(i) Node/link failures can be modeled by a sequence of random communication graphs \cite{Azzjf}. (ii) Node sensing failures or measurement losses can be modeled by a sequence of random observation/regression matrices, for example, $H_i(k)=\frac{1}{p}\mu_i(k)C_i$, where $\{\mu_i(k),k\ge 0\}$ is a sequence of zero-one i.i.d. Bernoulli variables accounting for sensor failures, $p>0$ is the sensing probability, and $C_i$ models the normal operation of the sensor \cite{Akar2012}-\cite{Akar20132}. Besides, in decentralized parameter identification \cite{Awjx}, all the nodes over graphs cooperatively identify the parameters of an auto-regressive (AR) model from noisy data, and each node's measurement equation is given by
\bna\label{Aexample111}
y_i(k)=\sum_{j=1}^dc_jy_i(k-j)+v_i(k),~k\ge 0,~i\in \mathcal V,
\ena
where $\{c_j\in \mathbb R, 1\leq j\leq d\}$ are the model parameters to be identified, and $v_i(k)\in \mathbb R$ is the additive zero-mean white noise. Here, the unknown parameter $x_0=[c_1,\cdots,c_d]^T\in \mathbb R^d$ and the random regression matrix $H_i(k)=[y_i(k-1),\cdots,y_i(k-d)]\in \mathbb R^{1\times d}$. }
\end{remark}

\vskip 0.2cm
\begin{remark}
\rm{In this paper, the network structure is modeled by a  sequence of random digraphs $\G(k,\omega)=\{\V,$ $\mathcal E_{\G(k,\omega)},$ $\A_{\G(k,\omega)}\}$, $k\ge 0$, where $\omega$ is the sample path,
$\A_{\G(k,\omega)}=[w_{ij}(k,\omega)]_{N\times N}$ is an $N$-dimensional random weighted adjacency matrix with zero diagonal elements, and $\mathcal E_{\G(k,\omega)}$ is the edge set, where each edge represents a communication link.
Denote the neighborhood of node $i$ at instant $k$ by $\mathcal N_i(k,\omega)=\{j\in \V|(j,i)\in \mathcal E_{\G(k,\omega)}\}$. There is a one-to-one correspondence between $\G(k,\omega)$  and $\A_{\G(k,\omega)}$.
Here, the sample path $\omega$ is omitted. The in-degree of node $i$ at instant $k$ $\mathrm{deg}^i_{\text{in}}(k)=\sum_{j\in\mathcal V}w_{ij}(k)$ and the out-degree of node $i$  at instant $k$ $\mathrm{deg}^i_{\text{out}}(k)=\sum_{j\in\mathcal V}w_{ji}(k)$. If $\mathrm{deg}^i_{\text{in}}(k)=\mathrm{deg}^i_{\text{out}}(k)$, $\forall~ i\in \mathcal V$, then $\G(k)$ is balanced. We call $\L_{\G(k)}=\mathcal D_{\G(k)}-\mathcal A_{\G(k)}$ the Laplacian matrix of $\G(k)$, where $\mathcal D_{\G(k)}=\textbf{diag}(\mathrm{deg}^1_{\text{in}}(k),\cdots,\mathrm{deg}^N_{\text{in}}(k))$.}
\end{remark}

\vskip 0.2cm
We now provide the following real-world example of decentralized linear regression with random regression matrices over random graphs. %More details

\vskip 0.2cm
\begin{example}
\rm{
In the decentralized multi-area state estimation in power systems \cite{Awjx}, the power grid is partitioned into multiple geographically non-overlapping areas, where the communication topology is modeled by random graphs. The state of the grid $x_0$ to be estimated consists of the amplitude of the voltages and the phase angles of all the buses. The measurement $y_i(k)$ in each region consists of the active power flow and the reactive power flow. After a DC power flow approximation \cite{Awood}, the grid state to be estimated degenerates into a vector of phase angles of all buses. The relationship between the measurement in the $i$-th region and the grid state is represented by \[y_i(k)=s_i(k)\overline{H}_i(k)x_0+v_i(k),~k\ge 0,\] where $\{v_i(k), k\ge 0\}$ is the sensing noise sequence, $\{s_i(k)\in \mathbb R, k\ge 0\}$ is a sequence of i.i.d. Bernoulli variables, which represents the intermittent sensor failures, and $\{\overline{H}_i(k),k\ge 0\}$ is the sequence of observation matrices without sensing failures.}
\end{example}

\vskip 0.2cm
Denote $y(k)=[y_1^T(k),\cdots,y_N^T(k)]^T$,  $H(k)=[H_1^T(k),\cdots,H_N^T(k)]^T$, $\mathcal H(k)=\textbf{diag}(H_1(k),\cdots$ $,H_N(k))$, and $v(k)=[v_1^T(k),\cdots,v_N^T(k)]^T$. Rewrite (\ref{Ameasurement}) by the compact form
\begin{equation}\label{Acompactmodel}
y(k)=H(k)x_0+v(k),~k\ge 0.
\end{equation}
The estimation of $\mathbf{1}_N\otimes x_0$ can be obtained by minimizing the loss function $\Psi(\cdot)$ as
\[
\hat{x}_0=\arg\min\limits_{x\in\mathbb R^{Nn}}\Psi(x)\triangleq\frac{1}{2}\left(\mathbb E\left[\|y(k)-\H(k)x\|^2\right] +\mathbb E\left[\left\langle \left(\L_{\G(k)}\otimes I_n\right)x,x\right\rangle\right]+\lambda \|x\|^2\right),~\lambda>0,
\]
if the sequences of the graphs, regression matrices and noises are identically distributed, respectively,  the mean graph is undirected, and the sequences of regression matrices and noises are mutually independent.
The loss function $\Psi (\cdot)$ consists of three parts: the mean squared error of the estimation $\E[\|y(k)-\H(k)x\|^2]$, the consensus error $\E[\langle (\L_{\G(k)}\otimes I_n)x,x\rangle]$ and the regularization term $\lambda \|x\|^2$. To solve the above optimization problem, we consider the stochastic gradient descent (SGD) algorithm:
\bna\label{Amklsvscx}
x(k+1)=x(k)+a(k)\H^T(k)\left(y(k)-\H(k)x(k)\right)-b(k)\left(\L_{\G(k)}\otimes I_n\right)x(k)-\lambda(k)x(k),\ena
where $\lambda(k)\triangleq\lambda c(k)$. Let $x(k)=[x^T_1(k),\cdots,x^T_N(k)]^T$, then $x_i(k)\in \mathbb R^n$ is the estimation of the node $i$ at instant $k$. From $(\ref{Amklsvscx})$, it follows that
\ban
&&x_i(k+1)=x_i(k)+a(k)H_i^T(k)\left(y_i(k)-H_i(k)x_i(k)\right)
+b(k)\sum_{j\in \mathcal N_i(k)}w_{ij}(k)(x_j(k)-x_i(k))\cr
&&~~~~~~~~~~~~~~-\lambda(k)x_i(k),~i\in \V,
\ean
where $a(k)H_i^T(k)(y_i(k)-H_i(k)x_i(k))$ is the innovation term updating the estimation $x_i(k)$ with the innovation gain $a(k)$, $b(k)\sum_{j\in \mathcal N_i(k)}w_{ij}(k)(x_{j}(k)-x_i(k))$ is the consensus term taking a weighted sum of estimations of its own and its neighbors with the consensus gain $b(k)$, and $\lambda(k)x_i(k)$ is the regularization term constraining the estimation $x_i(k)$
with the regularization gain $\lambda(k)$, which avoids overfitting the estimation of the unknown true parameter vector $x_0$. In the practical algorithm, there are communication noises among nodes. Specifically, node $j$ acquires the estimations of its neighbors by
\bna\label{Aeekksls}
\mu_{ji}(k)=x_j(k)+f_{ji}(x_j(k)-x_i(k))\xi_{ji}(k),~j\in \mathcal N_i(k),
\ena
where $\xi_{ji}(k)$ is the communication noise, and $f_{ji}(x_j(k)-x_i(k))$ is the  noise intensity function.

Therefore, the decentralized online regularization algorithm is given by
\bna\label{Aalgorithm}
&&x_i(k+1)=x_i(k)+a(k)H_i^T(k)(y_i(k)-H_i(k)x_i(k))+b(k)\sum_{j\in \mathcal N_i(k)}w_{ij}(k)(\mu_{ji}(k)-x_i(k))\cr
&&~~~~~~~~~~~~~~-\lambda(k)x_i(k),~i\in \mathcal V.
\ena
Here, by assuming that (i) the sequences of the graphs, regression matrices and noises are identically distributed, respectively; (ii) the mean graph is undirected; (iii) the sequences of regression matrices and noises are mutually independent, we obtain the above algorithm. In fact, even if the mean graphs are directed, and the graphs and regression matrices do not satisfy special statistical properties such as independence and stationarity, the algorithm (\ref{Aalgorithm}) will still be proved to converge.

\vskip 0.2cm
\begin{remark}
\rm{The communication model (\ref{Aeekksls}) with relative-state-dependent communication noises $f_{ji}(x_j(k)-x_i(k))$ $\xi_{ji}(k)$ are reasonable for many realistic applications. (i) In some multiple robots or UAV systems, due to the non-reliability of the communications, the transmission states are more prone to noises whose density functions depend on the distances between the transmitter and receiver, i.e., the relative distances between agents \cite{ADja}. (ii) In consensus problems with quantized measurements of relative states \cite{ADima}, the logarithmic quantized measurement by agent $i$ of $x_j(k)-x_i(k)$ is given by $x_j(k)-x_i(k)+(x_j(k)-x_i(k))\Delta_{ji}(k)$, which is a special case of (\ref{Aeekksls}), where $\Delta_{ji}(k)$ is the quantization uncertainty. (iii) In distributed averaging with Gaussian fading channels \cite{Ajwang}, the measurement of $x_j(k)-x_i(k)$ is given by $z_{ji}(k)=\xi_{ij}(k)(x_j(k)-x_i(k))$, where $\{\xi_{ij}(k),k\ge 0\}$ are independent Gaussian noises with mean value $\gamma_{ij}$. In \cite{Ajwang}, the above equation is transformed into $z_{ji}(k)=\gamma_{ij}(x_j(k)-x_i(k))+\Delta_{ij}(k)(x_j(k)-x_i(k))$ with $\Delta_{ij}(k)=\xi(k)-\gamma_{ij}$ being independent zero-mean Gaussian noises, which is a special case of (\ref{Aeekksls}). }
\end{remark}

\vskip 0.2cm
Denote $\mathscr F(k)=\sigma(\mathcal A_{\mathcal G(s)},H_i(s),v_i(s),\xi_{ji}(s),j,i\in \mathcal V,0\leq s\leq k)$ with $\mathscr F(-1)=\{\Omega,\emptyset\}$, and $\xi(k)=[\xi_{11}(k),\cdots,\xi_{N1}(k),\cdots,\xi_{1N}(k),\cdots,\xi_{NN}(k)]^T$. For the algorithm (\ref{Aalgorithm}), we need the following assumptions.
\vskip 0.2cm
\textbf{(A1)}~For the noise intensity function $f_{ji}(\cdot):\mathbb R^n \to \mathbb R$, there exist constants $\sigma$ and $b$, such that $|f_{ji}(x)|\leq \sigma\|x\|+b$, $\forall~x\in \mathbb R^n$.

\begin{remark}
\rm{Assumption \textbf{(A1)} indicates that the communication noises in (\ref{Aeekksls}) cover both additive and multiplicative noises, where $\sigma$ and $b$ are multiplicative and additive noise intensity coefficients, respectively.}
%Especially, if $f_{ji}(x)=\sigma_{ji}\|x\|+b_{ji}$, where $\sigma_{ji}$ and $b_{ji}$, $i,j\in\mathcal V$ are all constants, then Assumption \textbf{(A1)} holds with $\sigma=\max_{1\leq i,j\leq N}\{|\sigma_{ji}|\}$ and $b=\max_{1\leq i,j\leq N}\{|b_{ji}|\}$.
\end{remark}

\vskip 0.2cm
\textbf{(A2)}~The noises $\{v(k),\mathscr F(k),k\ge 0\}$ and $\{\xi(k),\mathscr F(k),k\ge 0\}$ are both martingale difference sequences and independent of $\{\H(k),$ $\mathcal A_{\mathcal G(k)},k\ge 0\}$. There exists a constant $\beta_v$, such that $\sup_{k\ge 0}$ $\mathbb E[\|v(k)\|^2+\|\xi(k)\|^2|\mathscr F(k-1)]\leq \beta_v~\text{a.s.}$

\vskip 0.2cm
\begin{remark}
\rm{Different from \cite{Akar2011}-\cite{Akar2013}, the measurement noises are only assumed to be a martingale difference sequence and independent of the graphs and regression matrices in Assumption \textbf{(A2)}. In this paper, neither mutual independence nor spatio-temporal independence is assumed on the regression matrices and graphs.
This is applicable to complex scenarios where regression matrices and graphs are spatio-temporal dependent.}
\end{remark}

\vskip 0.2cm
The problems of decentralized online regression over graphs have been investigated in most of the literature, including regression with time-varying unknown parameters (R.T.V.P.) (e.g. \cite{Axiesiyu1}-\cite{Axiesiyu2}) and regression with time-invariant unknown parameters (R.T.I.P.) (e.g. \cite{Akar2011}-\cite{Akar2013}, \cite{Asahu2018}, \cite{Awjx}, \cite{Azzjf} and \cite{Asayed1}-\cite{Apiggott}), where different assumptions of models and excitation conditions have been required. Here, we sum up in Table \ref{Asummarytable} the assumptions, excitation conditions that they required and their main results.

\begin{table}
  \caption{Summary of assumptions, excitation conditions and main results in relevant works.}\label{Asummarytable}
  \centering
  %\begin{tabular*}{\hline}{|c|c|p{4cm}|c|c|}
\scalebox{0.77}{
\begin{tabular}{|c|c|c|c|c|c|}
  \hline
\multicolumn{2}{|c|}{}                & \multicolumn{1}{c|}{\textbf{Regression matrices}}& \multicolumn{1}{c|}{\textbf{Graphs}}  & \textbf{Excitation conditions}\centering &  \multicolumn{1}{c|}{\textbf{Results}}   \\
  \hline
\multicolumn{1}{|c|}{\multirow{1}{*}{ \textbf{R.T.V.P.}}}  &  \multicolumn{1}{c|}{\begin{tabular}[c]{@{}l@{}} \cite{Axiesiyu1}-\cite{Axiesiyu2}\end{tabular}} & \multicolumn{1}{c|}{random and time-varying} &  \multicolumn{1}{c|}{\begin{tabular}[c]{@{}l@{}} deterministic, undirected and time-\\ invariant \end{tabular}} & \multicolumn{1}{c|}{\begin{tabular}[c]{@{}l@{}} connected graph and coope-\\rative information condition  \end{tabular}} & \multicolumn{1}{c|}{\begin{tabular}[c]{@{}l@{}} $L_p$-stable tracking errors\end{tabular}}   \\
\hline
\multicolumn{1}{|c|}{\multirow{7}{*}{\begin{tabular}[c]{@{}l@{}} \\ \\ \\ \\ \\ \\\\ \\ \textbf{R.T.I.P.}\end{tabular}}}  & \cite{Asayed1}\centering & \multicolumn{1}{c|}{spatio-temporally independent}  & \multicolumn{1}{c|}{deterministic and time-invariant} & \multicolumn{1}{c|}{\begin{tabular}[c]{@{}l@{}}stability of the information ~ \\ matrix\end{tabular}}  & \multicolumn{1}{c|}{\begin{tabular}[c]{@{}l@{}}mean and m.s.  convergence\end{tabular}}\\
  \cline{2-6}
   & \cite{Azzjf}\centering & \multicolumn{1}{c|}{i.i.d.} & \multicolumn{1}{c|}{\begin{tabular}[c]{@{}l@{}} balanced digraphs and homogeneo-\\us ergodic Markov chain\end{tabular}} & \multicolumn{1}{c|}{\begin{tabular}[c]{@{}l@{}} global observability and a ~~ \\ spanning tree contained in\\ the union of the graphs\end{tabular}} & \multicolumn{1}{c|}{\begin{tabular}[c]{@{}l@{}}a.s. and m.s.  convergence\end{tabular}} \\
 \cline{2-6}
& \cite{Akar20132}\centering & \multicolumn{1}{c|}{i.i.d.} & \multicolumn{1}{c|}{undirected and i.i.d.} & \multicolumn{1}{c|}{\begin{tabular}[c]{@{}l@{}}global observability and co-\\ nnected mean graph\end{tabular}} & \multicolumn{1}{c|}{ a.s. convergence}  \\
  \cline{2-6}
     & \cite{Apiggott}\centering & \multicolumn{1}{c|}{\begin{tabular}[c]{@{}l@{}}temporally strictly stationary and\\ temporally correlated\end{tabular}} & \multicolumn{1}{c|}{\begin{tabular}[c]{@{}l@{}}deterministic and time-invariant\end{tabular}} & \multicolumn{1}{c|}{\begin{tabular}[c]{@{}l@{}}spatially joint observability\end{tabular}} & \multicolumn{1}{c|}{m.s. convergence} \\
  \cline{2-6}
   & \cite{Asahu2018}\centering & \multicolumn{1}{c|}{\begin{tabular}[c]{@{}l@{}}time-invariant and deterministic\end{tabular}} & \multicolumn{1}{c|}{undirected and i.i.d.}   & \multicolumn{1}{c|}{\begin{tabular}[c]{@{}l@{}}global observability and co-\\ nnected mean graph\end{tabular}} & \multicolumn{1}{c|}{\begin{tabular}[c]{@{}l@{}}a.s.  convergence\end{tabular}}  \\
  \cline{2-6}
  & \cite{Awjx}\centering &  \multicolumn{1}{c|}{random and time-varying} & \multicolumn{1}{c|}{random and time-varying digraphs}  & \multicolumn{1}{c|}{\begin{tabular}[c]{@{}l@{}}stochastic spatio-temporal p-\\ ersistence of excitation  \end{tabular}} & \multicolumn{1}{c|}{\begin{tabular}[c]{@{}l@{}}a.s. and m.s.  convergence\end{tabular}}  \\
 \cline{2-6}
% & \cite{Awjx}\centering &  \multicolumn{1}{c|}{none} & none \centering & stochastic spatio-temporal persistence of excitation condition & almost sure and mean square convergence  \\
% \hline
 & This paper \centering & \multicolumn{1}{c|}{random and time-varying} & \multicolumn{1}{c|}{random and time-varying digraphs}  & \multicolumn{1}{c|}{\begin{tabular}[c]{@{}l@{}}sample path spatio-temporal\\ persistence of excitation \end{tabular}} &  \multicolumn{1}{c|}{\begin{tabular}[c]{@{}l@{}}a.s. and m.s.  convergence\end{tabular}} \\
  \hline
\end{tabular}}
\end{table}

\section{Main Results}
The convergence analysis of the algorithm (\ref{Aalgorithm}) are presented in this section. First, Lemma \ref{Alemma4} gives a nonnegative supermartingale type inequality of the squared estimation error. Based on which, Theorem \ref{ATheorem1} proves the almost sure convergence of the algorithm. Then, Theorem \ref{Atheorem2} gives intuitive convergence conditions for the case with balanced conditional digraphs by Lemma \ref{Alemma5}. Finally, Theorem \ref{Atheorem3}   establishes an upper bound for the regret of the algorithm and Theorem \ref{Atheorem4} gives a non-asymptotic rate for the algorithm. The proofs of theorems are in Appendix \ref{Aprooftheorem}, and those of lemmas in this section are in Appendix \ref{Aprooflemma4}.
%The proofs of Example \ref{Aexample222}, Theorem \ref{Aexample333} and Corollary \ref{Acorollary1} are in \cite{Azlf}.

Denote
$f_i(k)=\textbf{diag}(f_{1i}(x_1(k)-x_i(k)),\cdots,f_{Ni}(x_N(k)-x_i(k)))$;  $Y(k)=\textbf{diag}(f_1(k),\cdots,$ $f_N(k))$; $M(k)=Y(k)\otimes I_n$;  $W(k)=\textbf{diag}(\alpha_1^T(k)\otimes I_n,\cdots,\alpha_N^T(k)\otimes I_n)$, where $\alpha_i^T(k)$ is the $i$-th row of $\mathcal A_{\mathcal G(k)}$. A compact form equivalent to (\ref{Aalgorithm}) is given by
\bna\label{Acompactform}
&&\hspace{-0.3cm}x(k+1)=\big[(1-\lambda(k))I_{Nn}-b(k)\mathcal L_{\mathcal G(k)}\otimes I_n -a(k){\mathcal H}^T(k){\mathcal H}(k)\big]x(k)+a(k){\mathcal H}^T(k)y(k)\cr
&&~~~~~~~~~~~+b(k)W(k)M(k)\xi(k).
\ena
Denote $\delta(k)=x(k)-\textbf{1}_N\otimes x_0$ as the global estimation error. Noting that $(\mathcal L_{\mathcal G(k)}\otimes I_n)(\textbf{1}_N\otimes x_0)=(\mathcal L_{\mathcal G(k)}\textbf{1}_N)\otimes x_0=0$ and $\mathcal H(k)(\textbf{1}_N\otimes x_0)=H(k)x_0$, subtracting $\textbf{1}_N\otimes x_0$ from both sides of $(\ref{Acompactform})$ gives
\bna\label{Aerror}
&&\hspace{-0.4cm}\d(k+1)\cr
&&\hspace{-0.7cm}=\left((1-\lambda(k))I_{Nn}-b(k)\mathcal L_{\mathcal G(k)}\otimes I_n-a(k){\mathcal H}^T(k){\mathcal H}(k)\right)x(k)+a(k)\mathcal H^T(k)y(k) \cr
&&\hspace{-0.4cm}+b(k)W(k)M(k)\xi(k)-\textbf{1}_N\otimes x_0\cr
&&\hspace{-0.7cm}=\Phi_{P}(k,0)\d(0)+\sum_{i=0}^ka(i)\Phi_P(k,i+1)\mathcal H^T(i)v(i)+\sum_{i=0}^kb(i)\Phi_P(k,i+1)W(i)M(i)\xi(i)\cr
&&\hspace{-0.4cm}-\sum_{i=0}^k\lambda(i)\Phi_P(k,i+1)(\textbf{1}_N\otimes x_0),
\ena
where $P(k)=(1-\lambda(k))I_{Nn}-b(k)\mathcal{L}_{\mathcal{G}(k)}\otimes I_n-a(k)\mathcal{H}^{\mathrm{T}}(k)\mathcal{H}(k)$.

Denote $\widehat{\L}_{\G(k)}=\frac{\L_{\G(k)}+\L^T_{\G(k)}}{2}$. For any given positive integers $h$ and $k$, denote
\ban
&&\Lambda_k^h=\lambda_{\min}\Bigg[\sum_{i=kh}^{(k+1)h-1}\Big(b(i)\mathbb E\left[\widehat {\mathcal L}_{\mathcal G(i)}|\F(kh-1)\right]\otimes I_n  +a(i)\E\left[{\mathcal H}^T(i){\mathcal H}(i)|\F(kh-1)\right]\Big)\Bigg].
\ean
Denote $V(k)=\|\d(k)\|^2$. A nonnegative supermartingale type inequality of the squared estimation error $V(k)$ is obtained in the following lemma, which plays a key role in the convergence and performance analysis of the algorithm.
\vskip 0.2cm
\begin{lemma}\label{Alemma4}
\rm{For the algorithm (\ref{Aalgorithm}), if Assumptions \textbf{(A1)}-\textbf{(A2)} hold, the algorithm gains $a(k)$, $b(k)$ and $\lambda(k)$ monotonically decrease to zero, and there exists a positive integer $h$ and a positive constant $\rho_0$, such that
$\sup_{k\ge 0}(\|\mathcal L_{\mathcal G(k)}\|+(\mathbb E[\|{\mathcal H}^T(k){\mathcal H}(k)\|^{2^{\max\{h,2\}}}|\mathscr F(k-1)])^{\frac{1}{2^{\max\{h,2\}}}})\leq \rho_0$ a.s.,
then there exists a positive integer $k_0$, such that
\ban
&&\hspace{-0.7cm}\E[V((k+1)h)|\F(kh-1)]\leq (1+\Omega(k))V(kh)-2\left(\Lambda_k^h+\sum_{i=kh}^{(k+1)h-1}\lambda(i)\right)V(kh)\cr
&&~~~~~~~~~~~~~~~~~~~~~~~~~~~~~~~+\Gamma(k)~\text{a.s.},~k\ge k_0,
\ean
where $\{\Omega(k),k\ge 0\}$ and $\{\Gamma(k),k\ge 0\}$ are nonnegative deterministic real sequences satisfying $\Omega(k)+\Gamma(k)=\mathcal O(a^2(kh)+b^2(kh)+\lambda(kh))$.}
\end{lemma}
\vskip 0.2cm
\begin{remark}
\rm{The entries of $\mathcal A_{\G(k)}$ represent the weights of all edges of the graph. It is reasonable to assume that the weights are uniformly bounded with respect to the sample paths.

It is realistic to assume that the regression matrices are conditionally bounded.
For example, consider the AR model (\ref{Aexample111}) with i.i.d. Gaussian white noises. For this case, $\E[\|H^T_i(k)H_i(k)\||\F(k-1)]$ is bounded but $H_i(k)$ is not uniformly bounded with respect to the sample paths.}
\end{remark}
\vskip 0.2cm

The main results are as follows. Firstly, we analyze the convergence of the algorithm.
\vskip 0.2cm
\begin{theorem}\label{ATheorem1}
\rm{For the algorithm (\ref{Aalgorithm}), if Assumptions \textbf{(A1)}-\textbf{(A2)} hold, the algorithm gains $a(k)$, $b(k)$ and $\lambda(k)$ decrease monotonically, and there exists a positive integer $h$ and a positive constant $\rho_0$, such that (i) $\sum_{k=0}^{\infty}\Lambda_k^h=\infty$ a.s. with $\inf_{k\ge 0}(\Lambda_k^h+\lambda(k))\ge 0$ a.s.;
(ii)  $\sum_{k=0}^{\infty}(a^2(k)+b^2(k)+\lambda(k))<\infty$;
(iii) $\sup_{k\ge 0}(\|\mathcal L_{\mathcal G(k)}\|+(\mathbb E[\|{\mathcal H}^T(k){\mathcal H}(k)\|^{2^{\max\{h,2\}}}|\mathscr F(k-1)])^{\frac{1}{2^{\max\{h,2\}}}})\leq \rho_0$ a.s.,
then $\lim_{k\to\infty}x_i(k)=x_0,~i\in \mathcal V$ a.s.}
\end{theorem}
\vskip 0.2cm
The following proof sketch provides the main steps in the proof of Theorem \ref{ATheorem1}, from which we have solved the limitations of prior works.

\textbf{Proof sketch.} The non-commutative, non-independent and non-stationary random matrices pose intrinsic difficulties for analyzing the convergence of the online algorithm (\ref{Aalgorithm}), for which our proof  mainly consists of the following three steps. (I) We start by introducing the tools of conditional mathematical expectations in probability theory,  and then derive the upper bound of the state transition matrix, i.e., $\|\E[\Phi_P^T((k+1)h-1,kh)\Phi_P((k+1)h-1,kh)|\F(kh-1)]\|\leq 1-2(\Lambda_k^h+\sum_{i=kh}^{(k+1)h-1}\lambda(i))+\mathcal O(a^2(kh)+b^2(kh)+\lambda^2(kh))$ a.s. through the binomial expansion techniques, which helps us to transform the analysis of random matrix products into that of the minimum eigenvalue of the information matrix $\Lambda_k^h$ and we no longer need to separate the product of the random matrices by assuming independence as \cite{Akar2012}-\cite{Akar20132}. (II) Different from the existing techniques in \cite{Akar2012}-\cite{Akar20132} and \cite{Awjx}, which directly analyzed the mean squared error $\E[V(k)]$, we next turn to consider the squared estimation error $V(kh)$. By applying the inequality techniques in probability theory, we reveal the relation between $\E[V((k+1)h)|\F(kh-1)]$ and $V(kh)$ and derive the nonnegative supermartingale type inequality of the squared estimation error in Lemma \ref{Alemma4}. (III) Then, by the probabilistic structure of the estimation error obtained in Lemma \ref{Alemma4} and the convergence theorem for nonnegative supermartingales, we conclude that $V(kh)$ and the infinite series $\sum_{k=0}^{\infty}(\Lambda_k^h+\sum_{i=kh}^{(k+1)h-1}\lambda(i))V(kh)$ almost surely converge by choosing appropriate algorithm gains and the regularization parameter. (IV) Finally,  we prove that $V(kh)\to 0$ a.s. as $k\to\infty$ by the condition (i) in Theorem \ref{ATheorem1}, where we no longer require the minimum eigenvalue of the information matrix $\Lambda_k^h$ has a positive and deterministic lower bound as in \cite{Awjx}.
As a consequence, our results not only remove the reliance on the special statistical properties of regression matrices and graphs, but also weaken the conditions in our previous works.

\vskip 0.2cm
\begin{remark}\label{Aremark666}
\rm{The choices of the algorithm gains $a(k)$, $b(k)$ and the regularization parameter $\lambda(k)$ is crucial for the nodes to estimate the unknown true parameter vector $x_0$.
The conditions (i) and (iii) in Theorem \ref{ATheorem1} imply $\sum_{k=0}^{\infty}(a(k)+b(k))=\infty$, which means that the algorithm gains can not be too small. Besides, to reduce the effects of perturbing the innovations and consensus by the communication noises, the monotonically decreasing algorithm gains and the condition (ii) in Theorem \ref{ATheorem1} ensure that all nodes avoid making excessive changes to the current estimate when acquiring new noisy data. The conditions (i) and (ii) give what need to be satisfied for the algorithm gains.

It is worth noting that the information matrix $\Lambda_k^h$ in the condition (i) is coupled with the gains, graphs and regression matrices. For some special cases, Theorem \ref{Atheorem2} later provides acceptable ranges on the values of these gains.
}
\end{remark}

\vskip 0.2cm
\begin{remark}
\rm{Most existing literature on decentralized online regression suppose that the mean graphs are balanced and strongly connected (e.g. \cite{Akar2012}-\cite{Akar20132} and \cite{Aydm}). Here, the condition (i) in Theorem \ref{ATheorem1} may still hold even if the mean graphs are unbalanced and not strongly connected. For example, consider a fixed digraph $\G=\{\mathcal V=\{1,2\}, \mathcal A_{\G}=[w_{ij}]_{2\times 2}\}$ with $w_{12}=1$ and $w_{21}=0$. Obviously, $\G$ is unbalanced and not strongly connected. Suppose $H_1=0, H_2=1$. Choose $a(k)=b(k)=\frac{1}{k+1}$. We have $\lambda_{\min}(b(k)\widehat{\L}_{\G}+a(k)\H^T\H)=\frac{1}{k+1}\lambda_{\min}(\widehat{\L}_{\G}+\H^T\H)=\frac{1}{2k+2}$. Then, the condition (i) holds with $h=1$ and $\sum_{k=0}^{\infty}\Lambda_k^h=\sum_{k=0}^{\infty}\frac{1}{2k+2}=\infty$.}
\end{remark}

\vskip 0.2cm
For the special case without regularization, we directly obtain the following corollary by Theorem \ref{ATheorem1}.
\vskip 0.2cm
\begin{corollary}\label{Amlnvlllsnrf}
\rm{For the algorithm (\ref{Aalgorithm}) with $\lambda (k)\equiv 0$, if Assumptions \textbf{(A1)}-\textbf{(A2)} hold, and there exists a positive integer $h$ and a positive constant $\rho_0$, such that
(i) $\sum_{k=0}^{\infty}\Lambda_k^h=\infty$ a.s. with $\inf_{k\ge 0}\Lambda_k^h$ $\ge 0$ a.s.;
(ii) $\sum_{k=0}^{\infty}(a^2(k)+b^2(k))<\infty$;
(iii)
$\sup_{k\ge 0}(\|\mathcal L_{\mathcal G(k)}\|+(\mathbb E[\|{\mathcal H}^T(k){\mathcal H}(k)\|^{2^{\max\{h,2\}}}|\mathscr F(k-1)]$ $)^{\frac{1}{2^{\max\{h,2\}}}})\leq \rho_0$ a.s.,
then
$\lim_{k\to\infty}x_i(k)=x_0,~i\in \mathcal V$ a.s.}
\end{corollary}

\vskip 0.2cm
\begin{remark}
\rm{The conditions for the regularized algorithm in Theorem \ref{ATheorem1} are less conservative than those for the non-regularized algorithm in Corollary \ref{Amlnvlllsnrf}. For example, consider the digraph sequence $\{\G(k)=\{\mathcal V=\{1,2\}, \mathcal A_{\G(k)}=[w_{ij}(k)]_{2\times 2}\},k\ge 0\}$ with $w_{12}(k)=\begin{cases}1, & k=2m; \\ \frac{1}{k}, & k=2m+1 \end{cases}$ and $w_{21}(k)=0$, and the regression matrices $H_1(k)=0$, $H_2(k)=\begin{cases} 1, & k=2m; \\ \frac{1}{\sqrt{5k}}, & k=2m+1 \end{cases}$, $m\ge 0$. Choose $a(k)=b(k)=\frac{1}{k+1}$. Then, we have $\sum_{k=0}^{\infty}\Lambda_k^1=\infty$. Obviously, the condition (i) in  Corollary \ref{Amlnvlllsnrf} does not hold since $\inf_{k\ge 0}\Lambda_{k}^1\leq \inf_{k\ge 0}\Lambda_{2k+1}^1=\inf_{k\ge 0}(-\frac{\sqrt{41}-6}{10(2k+1)^2})<0$. Choose $\lambda(k)=\frac{1}{10(k+1)^2}$. Then, the condition (i) in Theorem \ref{ATheorem1} holds with $\inf_{k\ge 0}(\Lambda_k^1+\lambda(k))=0$.
}
\end{remark}

\vskip 0.2cm
Subsequently, we list the conditions on the algorithm gains that may be needed later.

\vskip 0.2cm
\textbf{(C1)} The algorithm gains $\{a(k),k\ge 0\}$, $\{b(k),k\ge 0\}$ and $\{\lambda(k),k\ge 0\}$ are all nonnegative sequences monotonically decreasing, and satisfy $
\sum_{k=0}^{\infty}\min\{a(k),b(k)\}=\infty$, $ \sum_{k=0}^{\infty}(a^2(k)+b^2(k)+\lambda(k))<\infty$, $a(k)=\mathcal O(a(k+1))$, $b(k)=\mathcal O(b(k+1))$ and $\lambda(k)=o(\min\{a(k),b(k)\})$.
\vskip 0.2cm
\textbf{(C2)} The algorithm gains $\{a(k),k\ge 0\}$, $\{b(k),k\ge 0\}$ and $\{\lambda(k),k\ge 0\}$ are all nonnegative sequences monotonically decreasing to zero,  and satisfy $
\sum_{k=0}^{\infty}\min\{a(k),b(k)\}=\infty$, $a(k)=\mathcal O(a(k+1))$, $b(k)=\mathcal O(b(k+1))$ and  $a^2(k)+b^2(k)+\lambda(k)=o(\min\{a(k),b(k)\})$.

\vskip 0.2cm
\begin{remark}
\rm{As mentioned previously, the regularization gain $\lambda(k)$ has the function of constraining the estimation $x_i(k+1)$ in the algorithm (\ref{Aalgorithm}). To ensure that the algorithm can cooperatively estimate the unknown true parameter vector $x_0$ by means of the innovation and consensus terms, Conditions \textbf{(C1)}-\textbf{(C2)} require the regularization gain $\lambda(k)$ to decay faster than the innovation gain $a(k)$ and the consensus gain $b(k)$.}
\end{remark}
\vskip 0.2cm
Next, we consider the sequence of balanced conditional digraphs
\ban
&&\hspace{-0.6cm}\Gamma_1=\big\{\{\G(k),k\ge 0\}|\E\left[\A_{\G(k)}|\F(k-1)\right]\succeq \textbf{O}_{N\times N}~\text{a.s.},~\G(k|k-1)~\text{is balanced}~\text{a.s.},~k\ge 0\big\}.
\ean
Here, $\E[\A_{\G(k)}|\F(m)]$, $m<k$, is called the \emph{conditional generalized weighted adjacency matrix} of $\A_{\G(k)}$ w.r.t. $\F(m)$, and its associated random graph is called the \emph{conditional digraph} of $\G(k)$ w.r.t. $\F(m)$, denoted by $\G(k|m)$, i.e., $\G(k|m)=\{\mathcal V,\E[$ $\A_{\G(k)}|\F(m)]\}$.

For any given positive integers $h$ and $k$, we denote
\[
\widetilde\Lambda_k^h=\lambda_{\min}\Bigg[\sum_{i=kh}^{(k+1)h-1}\mathbb E\Big[\widehat {\mathcal L}_{\mathcal G(i)}\otimes I_n+{\mathcal H}^T(i){\mathcal H}(i)|\F(kh-1)\Big]\Bigg].
\]
Then $\widetilde\Lambda_k^h$ contains information of both the Laplacian matrices of the graphs and regression matrices. As shown in Remark \ref{Aremark666}, the information matrix $\Lambda_k^h$ in the condition (i) in Theorem \ref{ATheorem1} is coupled with the gains, graphs and regression matrices, it is generally difficult to decouple the algorithm gains from $\Lambda_k^h$ for general graphs and regression matrices since $\widehat{\L}_{\G(k)}$ is indefinite. For the case with the conditionally balanced digraphs $\G(k)\in \Gamma_1$, we have $\Lambda_k^h\ge \min\{a(i),b(i),kh\leq i<(k+1)h\}\widetilde\Lambda_k^h$ a.s.  The following lemma gives a lower bound of $\widetilde\Lambda_k^h$, where one can see how the conditionally balanced graphs and regression matrices of all nodes affect the lower bound, respectively.

\vskip 0.2cm
\begin{lemma}\label{Alemma5}
\rm{For the algorithm (\ref{Aalgorithm}), suppose that  $\{\mathcal G(k),k\ge 0\}\in \Gamma_1$, and there exists a positive constant $\rho_0$, such that
\[
\sup_{k\ge 0} \mathbb E\left[\left\|{\mathcal H}^T(k){\mathcal H}(k)\right\||\mathscr F(k-1)\right]\leq \rho_0~\text{a.s.}
\]
Then, for any given positive integer $h$,
\ban
&&\hspace{-0.5cm}\widetilde\Lambda_k^h \ge\frac{\displaystyle\lambda_2\Bigg[\sum_{i=kh}^{(k+1)h-1}\E\left[\widehat \L_{\G(i)}|\F(kh-1)\right]\Bigg]}{\displaystyle 2Nh\rho_0+N\lambda_2\Bigg[\sum_{i=kh}^{(k+1)h-1}\E\left[\widehat \L_{\G(i)}|\F(kh-1)\right]\Bigg]}\cr  &&\hspace{-0.2cm}\times\lambda_{\min}\Bigg(\sum_{i=1}^N\sum_{j=kh}^{(k+1)h-1}\E\left[H_i^T(j)H_i(j)|\F(kh-1)\right]\Bigg)~\text{a.s.}
\ean
}
\end{lemma}

\vskip 0.2cm
\begin{remark}
\rm{If the network structure degenerates into a deterministic, undirected and connected graph $\G$, and $H^T_i(k)H_i(k)\leq \rho_0I_{n}$, $i\in \mathcal V$ a.s., then the inequality in Lemma \ref{Alemma5} degenerates to
\ban
&&\hspace{-0.5cm}\widetilde\Lambda_k^h\ge \frac{\lambda_2\left(\L_{\G}\right)}{2Nh\rho_0+N\lambda_2\left(\L_{\G}\right)}\cr &&\hspace{-0.2cm}\times\lambda_{\min}\Bigg(\sum_{i=1}^N\sum_{j=kh}^{(k+1)h-1}\E\left[H_i^T(j)H_i(j)|\F(kh-1)\right]\Bigg)~\text{a.s.},
\ean
which is given in \cite{Axiesiyu1}-\cite{Axiesiyu2}.}
\end{remark}

\vskip 0.2cm
Then, we give intuitive convergence conditions for the case with balanced conditional digraphs. We first introduce the following definitions.
\vskip 0.2cm
\begin{definition}\label{Adefinition1}
\rm{For the random undirected graph sequence $\{\G(k),k\ge 0\}$, if there exists a positive integer $h$ and a positive constant $\theta$, such that $$\inf_{k\ge 0}\lambda_2\left(\sum_{i=kh}^{(k+1)h-1}\E\left[\L_{\G(i)}|\F(kh-1)\right]\right)\ge \theta~\mathrm{a.s.},$$ then $\{\G(k),k\ge 0\}$ is said to be \emph{uniformly conditionally jointly connected}.}
\end{definition}

\vskip 0.2cm
\begin{definition}\label{Adefinition2}
\rm{For the sequence of regression matrices $\{H_i(k),i=1,\cdots, N, k\ge 0\}$, if there exists a positive integer $h$ and a positive constant $\theta$, such that $$\inf_{k\ge 0}\lambda_{\min}\left(\sum_{i=1}^N\sum_{j=kh}^{(k+1)h-1}\mathbb E\left[H_i^T(j)H_i(j)|\mathscr F(kh-1)\right]\right)\ge \theta~\mathrm{a.s.},$$ then $\{H_i(k),i=1,\cdots, N, k\ge 0\}$ is said to be \emph{uniformly conditionally spatio-temporally jointly observable}.}
\end{definition}
\vskip 0.2cm
Denote the symmetrized graph of $\G(k)$ by $\widehat\G(k)=\{\mathcal V, \mathcal E_{\G(k)}\cup \mathcal E_{\widetilde\G(k)},\frac{\mathcal A_{\G(k)}+\mathcal A^T_{\G(k)}}{2}\}$, where $\widetilde \G(k)$ is the reversed digraph of $\G(k)$ \cite{Alw}.
\vskip 0.2cm
\begin{theorem}\label{Atheorem2}
\rm{For the algorithm (\ref{Aalgorithm}), suppose that  $\{\mathcal G(k),k\ge 0\}\in \Gamma_1$, Assumptions \textbf{(A1)}-\textbf{(A2)} hold, and there exists a positive integer $h$ and a positive constant $\rho_0$, such that
(i) $\{\widehat\G(k),k\ge 0\}$ is \emph{uniformly conditionally jointly connected};
(ii) $\{H_i(k),i\in\mathcal V,k\ge 0\}$ is \emph{uniformly conditionally spatio-temporally jointly observable}; (iii)
$\sup_{k\ge 0}(\|\mathcal L_{\mathcal G(k)}\|+(\mathbb E[\|{\mathcal H}^T(k){\mathcal H}(k)\|^{2^{\max\{h,2\}}}|\mathscr F(k-1)])^{\frac{1}{2^{\max\{h,2\}}}})\leq \rho_0$ a.s.\\
(I). If Condition \textbf{(C1)} holds, then $\lim_{k\to\infty}x_i(k)=x_0,~i\in \mathcal V$ a.s.\\
(II). If Condition \textbf{(C2)} holds, then $\lim_{k\to\infty}\E[\|x_i(k)-x_0\|^2]=0,~i\in \mathcal V$.}%\\
\end{theorem}
\vskip 0.2cm

The combination of the conditions (i)-(ii) in Theorem \ref{Atheorem2} with Condition \textbf{(C1)} or Condition \textbf{(C2)} gives an intuitive sufficient condition for the condition (i) in Theorem \ref{ATheorem1} to hold.

The condition (i) in Theorem \ref{ATheorem1} is called  the \textbf{\emph{sample path spatio-temporal persistence of excitation}} condition, which is an indispensable part of the convergence and performance analysis of the algorithm.
Specifically, \textbf{\emph{spatio-temporal persistence of excitation}} means that the infinite series of the minimum eigenvalues of the information matrices composed of the graphs and regression matrices in fixed-length time intervals diverges for almost all sample paths, i.e., $\sum_{k=0}^{\infty}\Lambda_k^h=\infty$ a.s., which is to avoid the failure of estimating the unknown true parameter vector due to lack of effective measurement information or sufficient information exchange among nodes. To illustrate this, let us  consider the extreme case that the regression matrix of each node at each instant is always zero, then $\Lambda_k^h=0$ for any given positive integer $h$.  In this case, no matter how to design the algorithm gains, one can not obtain any information about the unknown true parameter vector since there is no measurement information and no information interflow. Here,
\textbf{\emph{spatio-temporality}} focuses on the state of the information matrices consisting of the graphs and regression matrices of all nodes over a fixed-length time period rather than the state at each instant, where the temporality is captured by $h$.
From Theorem \ref{Atheorem2}, we know that neither the \emph{locally temporally joint observability} of each node, i.e., $\inf_{k\ge 0}\lambda_{\min}(\E[\sum_{j=kh}^{(k+1)h-1}H_i^T(j)H_i(j)|\F(kh-1)])$ being uniformly bounded away from zero for each node $i$, nor the \emph{instantaneously globally spatially joint observability} of all the regression models, i.e., $\inf_{k\ge 0}\lambda_{\min}(\E[\sum_{i=1}^{N}H_i^T(j)H_i(j)|\F(kh-1)])$ being  uniformly bounded away from zero for each instant $j$, is necessary.

At present, most results on decentralized online linear regression algorithms (e.g. \cite{Akar2012}-\cite{Akar20132} and \cite{Azzjf}-\cite{Azhz}) all require that the regression matrices and graphs satisfy some special statistical properties, such as i.i.d., spatio-temporal independence or stationarity, etc. However, these special statistical assumptions are difficult to be satisfied if the regression matrices are generated by the auto-regressive models. In order to solve this problem, in the past several decades, many scholars have proposed the persistence of excitation condition based on the conditional expectations of the regression matrices.
The stochastic persistence of excitation condition was first proposed in the analysis of the centralized Kalman filter algorithm in \cite{Aguolei} and then refined in \cite{AZJF}. For the decentralized adaptive filtering algorithms in \cite{Axiesiyu1}-\cite{Axiesiyu2}, the cooperative information condition on the conditional expectations of the regression matrices was proposed for the case with deterministic connected graphs. For the decentralized online estimation algorithms over random time-varying graphs in \cite{Awjx}, the stochastic spatio-temporal persistence of excitation condition was proposed.
The stochastic spatio-temporal persistence of excitation condition in \cite{Awjx} means that the minimum eigenvalue of the matrix consisting of the spatial-temporal observation matrices and Laplacian matrices in each time period $[kh,(k+1)h-1]$ has a positive lower bound $c(k)$ independent of the sample paths, i.e., $\Lambda_k^h\ge c(k)$ a.s. with $\sum_{k=0}^{\infty}c(k)=\infty$.
Here, in Corollary \ref{Amlnvlllsnrf}, we show that this is actually not needed. Therefore, the \emph{sample path spatio-temporal persistence of excitation} condition in this paper is more general than the stochastic spatio-temporal persistence of excitation condition in \cite{Awjx}.
%where the \emph{sample path spatio-temporal persistence of excitation} condition means
%$\sum_{k=0}^{\infty}\Lambda_k^h=\infty$ a.s. with $\inf_{k\ge 0}\Lambda_k^h\ge 0$ a.s. for the case without regularization.

We give an example for which the stochastic spatio-temporal persistence of excitation condition does not hold but the \emph{sample path spatio-temporal persistence of excitation} condition hold.

\vskip 0.2cm
\begin{example}\label{Aexample333}
\rm{Consider a fixed digraph $\G=\{\mathcal V=\{1,2\}, \mathcal A_{\G}=[w_{ij}]_{2\times 2}\}$ with $w_{12}=1$ and $w_{21}=0$. Let $H_1=0$ and $H_2=\sqrt{x}$, where the random variable $x$ is uniformly distributed in $(0.25,1.25)$. Choose $a(k)=b(k)=\frac{1}{k+1}$. For any given positive integer $h$, if Assumption \textbf{(A2)} holds, then the \emph{sample path spatio-temporal persistence of excitation} condition holds and there does not exist a positive real sequence $\{c(k),k\ge 0\}$ satisfying $\Lambda_k^h\ge c(k)$ a.s., i.e., the stochastic spatio-temporal persistence of excitation condition in \cite{Awjx} does not hold.}
\end{example}

\vskip 0.2cm
Therefore, the \emph{sample path spatio-temporal persistence of excitation} condition weakens the stochastic spatio-temporal persistence of excitation condition in \cite{Awjx}. To our best knowledge, we have obtained the most general persistence of excitation condition ever.

To appraise the performance of the algorithm (\ref{Aalgorithm}), the regret upper bound of the decentralized online regularized algorithm will be taken into account. The loss function of node $j$ at instant $t$ is defined by
$$l_{j,t}(x)\triangleq \frac{1}{2}\left\|H_j(t)x-y_j(t)\right\|^2,~x\in \mathbb R^n.$$
The performance of the algorithm (\ref{Aalgorithm}) is appraised, at node $i$, through its regret defined by
\ban
&&\text{Regret}_{\text{LMS}}(i,T)\triangleq\sum\limits_{t=0}^T\sum\limits_{j=1}^N\mathbb{E}\left[l_{j,t}(x_i(t))-l_{j,t}\left(x_{\text{LMS}}^*\right)\right],
\ean
where
\[x_{\text{LMS}}^*\triangleq\arg\min\limits_{x\in\mathbb R^n}\sum\limits_{t=0}^T\sum\limits_{j=1}^N
\frac{1}{2}\mathbb{E}\left[\|H_j(t)x-y_j(t)\|^2\right]\]
is the linear optimal estimated parameter.

%For the \emph{uniformly conditionally jointly connected} graphs and \emph{uniformly conditionally spatio-temporally jointly observable} regression models,
The following theorems give an upper bound of $\text{Regret}_{\text{LMS}}(i,T)$, $i\in \mathcal V$, and a non-asymptotic rate for the algorithm.

\vskip 0.2cm
\begin{theorem}\label{Atheorem3}
\rm{For the algorithm (\ref{Aalgorithm}), suppose that $\{\mathcal G(k),k\ge 0\}\in \Gamma_1$ and Assumptions \textbf{(A1)}-\textbf{(A2)} hold. If
$a(k)=b(k)=\frac{c_1}{(k+1)^{\tau}}$ and  $\lambda(k)=\frac{c_2}{(k+1)^{2\tau}}$, $c_1>0$,  $c_2>0$, $0.5<\tau<1$, and there exists a positive integer $h$ and a positive constant $\rho_0$, such that
(i) $\{\widehat\G(k),k\ge 0\}$ is \emph{uniformly conditionally jointly connected};
(ii) $\{H_i(k),i\in\mathcal V,k\ge 0\}$ is \emph{uniformly conditionally spatio-temporally jointly observable}; (iii)
$\sup_{k\ge 0}(\|\mathcal L_{\mathcal G(k)}\|+(\mathbb E[\|{\mathcal H}^T(k){\mathcal H}(k)\|^{2^{\max\{h,2\}}}|\mathscr F(k-1)])^{\frac{1}{2^{\max\{h,2\}}}})\leq \rho_0$ a.s.,
then the regret of node $i$ satisfies
\bna\label{Ayiyiyiy}
\text{Regret}_{\text{LMS}}(i,T)=\mathcal O\left(T^{1-\tau}\ln T\right),~i\in\mathcal V.
\ena
}
\end{theorem}

\vskip 0.2cm
\begin{remark}
\rm{Yuan \emph{et al}. \cite{Aydm} studied the non-regularized decentralized online linear regression algorithm over the fixed graph.
In this paper, we consider the decentralized online regularized linear regression algorithm over the random time-varying graphs.
Theorem $\ref{Atheorem3}$ shows that the upper bound of the regret is $\mathcal O(T^{1-\tau}\ln T)$, where $\tau\in (0.5,1)$ is a constant depending on the decaying algorithm gains, which improves the regret upper bound $\mathcal O(\sqrt{T})$ with the fixed gain in \cite{Aydm}.}
\end{remark}

\vskip 0.2cm

\begin{theorem}\label{Atheorem4}
\rm{For the algorithm (\ref{Aalgorithm}), suppose that $\{\mathcal G(k),k\ge 0\}\in \Gamma_1$ and Assumptions \textbf{(A1)}-\textbf{(A2)} hold. If
$a(k)=b(k)=\frac{1}{(k+1)^{\tau}}$ and  $\lambda(k)=\frac{1}{(k+1)^{2\tau}}$, $0.5<\tau<1$, and there exists a positive integer $h$ and a positive constant $\rho_0$, such that
(i) $\{\widehat\G(k),k\ge 0\}$ is \emph{uniformly conditionally jointly connected};
(ii) $\{H_i(k),i\in\mathcal V,k\ge 0\}$ is \emph{uniformly conditionally spatio-temporally jointly observable}; (iii)
$\sup_{k\ge 0}(\|\mathcal L_{\mathcal G(k)}\|+(\mathbb E[\|{\mathcal H}^T(k){\mathcal H}(k)\|^{2^{\max\{h,2\}}}|\mathscr F(k-1)])^{\frac{1}{2^{\max\{h,2\}}}})\leq \rho_0$ a.s.,
then there exist constants $c_i$, $i=1,\cdots,7$, such that $\mathbb{E}[V(k)]$ is bounded by
\ban
2^h \left(\frac{25c_4\ln (k+1)}{c_5\left(\frac{k}{2h}\right)^{\tau}}+c_6e^{-\frac{c_5\left(\left(\frac{k}{2h}\right)^{1-\tau}-\left(k_0+2\right)^{1-\tau}\right)}{1-\tau}}\right)+\frac{c_7}{k^{2\tau}},
\ean
when $k$ is larger than
\ban
\left\lfloor \left(\frac{12}{c_5(1-\tau)}\ln \left(\frac{4}{c_5(1-\tau)}\right)\right)^{\frac{1}{1-\tau}}+ \frac{4c_1^2}{c_2^2}+c_3^2  \right\rfloor+2.
\ean
}
\end{theorem}

\vskip 0.2cm
The proof of Theorem \ref{Atheorem4} is similar to  Theorem \ref{Atheorem3}, in which we further analyze the nonnegative supermartingale type inequality of $V(k)$ given by Lemma \ref{Alemma4}, and obtain the upper bound of $\E[V(k)]$ by Lemma 2.9 in \cite{ALeonard}. The detailed proof is put in Appendix \ref{Aprooftheorem}.

\section{Numerical Example}
We consider the graphs composed of 10 nodes, and the states are $x_i(k)$, $i=1,\cdots,10$, $k\geq 0$.
Each node estimates the unknown true parameter vector $x_0=[5,4,3]^T$.

\begin{figure}[ht]
\centering
\includegraphics[width=.5\columnwidth]{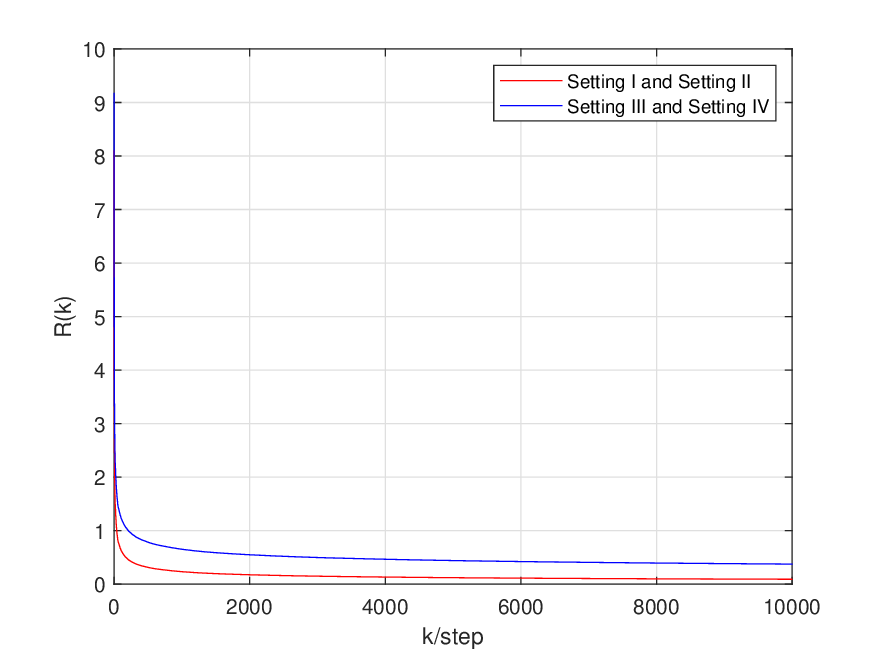}
\caption{The sample paths of $R(k)$ with different settings.}
\label{Apeconditions}
\end{figure}

The local observation  is given by $y_i(k)=H_i(k)x_0+v_i(k)$, $i=1,\cdots,10$. Here, the regression matrices are taken as $$\begin{pmatrix} 0 &0 & 0\\\bar{h}_{s,t,k} &0 & 0 \end{pmatrix}, ~\begin{pmatrix}0 & 0 &\bar{h}_{s+2,t,k} \\0 & 0 & 0\\ 0 & 0 & 0 \end{pmatrix}, ~\begin{pmatrix}0 &\bar{h}_{5,t,k} &0\\ 0 &0 &0\\ \bar{h}_{6,t,k} &0 &0 \end{pmatrix}, ~s=1,2,~t=1,2,$$ where $\bar{h}_{s,t,k}=(-1)^sh_{s,t}(k)+(-1)^t0.5, \bar{h}_{s+2,t,k}=(-1)^sh_{s+2,t}(k)+(-1)^t0.5, \bar{h}_{5,t,k}=(-1)^th_{5,t}(k)$ $+(-1)^t0.5, \bar{h}_{6,t,k}=(-1)^{t+1}h_{6,t}(k)+(-1)^{t+1}0.5$, and $h_{i,j}(k), i=1,2,3,4,5,6, j=1,2$, are independent random variables with uniform distribution on $(0,1)$.
%A direct calculation gives
%\[\inf_{k\ge 0}\lambda_{\min}\left(\sum_{i=1}^3\sum_{j=2k}^{2k+1}\E\left[H_i^T(j)H_i(j)|\F(2k-1)\right]\right)\ge 0.2,\]
%which shows that the regression matrices are uniformly conditionally spatio-temporally jointly observable.
The signal at the receiver of the communication channel is described by (\ref{Aeekksls}), where $f_{ji}(x_j(k)-x_i(k))=0.1||x_j(k)-x_i(k)||+0.1$, $i, j= 1,\cdots,10$. The updating rules of  the estimation states follow algorithm (\ref{Aalgorithm}),
% \ban
% &&\hspace{-0.7cm}x_1(k+1)=x_1(k)+k^{-0.6}H_1^T(k)(y_1(k)-H_1(k)x_1(k))\cr
 %&&\hspace{-0.4cm}+k^{-0.6}\sum_{j=2,3}a_{1j}(k)(x_j(k)-x_1(k)\cr
 %&&\hspace{-0.4cm}+\sigma||x_j(k)-x_1(k)||\xi_{j1}(k)+b\xi_{j1}(k))-k^{-2}x_1(k),
% \ean
% \ban
% &&\hspace{-0.7cm}x_2(k+1)=x_2(k)+k^{-0.6}H_2^T(k)(y_2(k)-H_2(k)x_2(k))\cr
% &&\hspace{-0.4cm}+k^{-0.6}\sum_{j=1,3}a_{2j}(k)(x_j(k)-x_1(k)\cr
% &&\hspace{-0.4cm}+\sigma||x_j(k)-x_2(k)||\xi_{j2}(k)+b\xi_{j2}(k))-k^{-2}x_2(k),
 %\ean
% \ban
% &&\hspace{-0.7cm}x_3(k+1)=x_3(k)+k^{-0.6}H_3^T(k)(y_3(k)-H_3(k)x_3(k))\cr
 %&&\hspace{-0.4cm}+k^{-0.6}\sum_{j=1,2}a_{3j}(k)(x_j(k)-x_3(k)\cr
 %&&\hspace{-0.4cm}+\sigma||x_j(k)-x_3(k)||\xi_{j3}(k)+b\xi_{j3}(k))-k^{-2}x_3(k),
% \ean
where the random weights $\{w_{ij}(k),i,j=1,\cdots,10, k\geq 0\}$ are selected by the following rules. When $k=2m$, $m\ge0$, the random weights are uniformly distributed in the interval $[0,1]$; when $k\not=2m$, $m\ge0$, the random weights are uniformly distributed in $[-0.5,0.5]$. So the random weights may be negative at some time instants. Here, $\{w_{ij}(k)$, $i,j=1,\cdots,10$, $k\geq0\}$  are spatio-temporally independent. Thus, when $k=2m, m\geq 0$, the average graph is balanced and connected, and when $k\neq 2m, m\geq 0$, the average graph is empty.
%Therefore, the average graph in $[2m,2(m+1))$ is jointly connected and uniformly conditionally spatio-temporally jointly connected with
%$$\inf_{k\ge 0}\lambda_2\left(\sum_{i=2k}^{2k+1}\E\left[\widehat\L_{\G(i)}|\F(2k-1)\right]\right)\ge 0.5.$$
Suppose that the measurement noises $\{v_i(k),i=1,\cdots,10, k\geq 0\}$ and the communication noises $\{\xi_{ji}(k),i,j=1,\cdots,10, k\geq 0\}$ are normally distributed r.v.s and independent of the random graphs. Different settings of the algorithm gains are shown in Table \ref{ATabf}, where we take $a(k)=b(k)$ for simplicity.
\begin{table}
\caption{Settings of the algorithm gains}\label{ATabf}
\centering
\begin{tabular}{|c|c|c|c|}
  \hline
   & $a(k)$ & $b(k)$ & $\lambda(k)$ \\
     \hline
  Setting I & $(k+1)^{-0.6}$ & $(k+1)^{-0.6}$ & $(k+1)^{-2}$ \\
    \hline
    Setting II & $(k+1)^{-0.6}$ & $(k+1)^{-0.6}$ & $0.1(k+1)^{-3}$ \\
      \hline
    Setting III & $(k+1)^{-0.8}$ & $(k+1)^{-0.8}$ & $(k+1)^{-2}$ \\
      \hline
      Setting IV & $(k+1)^{-0.8}$ & $(k+1)^{-0.8}$ & $0$ \\
  \hline
%    Setting V & $(k+1)^{-0.6}$ & $(k+1)^{-0.6}$ & $0$ \\
%    \hline
%      Setting VI & $(k+1)^{-0.8}$ & $(k+1)^{-0.8}$ & $0$ \\
%    \hline
\end{tabular}
\end{table}
To verify the \emph{sample path spatio-temporal persistence of excitation} condition, Fig. \ref{Apeconditions} presents the curves of the quantity $R(k)\triangleq(\sum_{i=0}^k\Lambda_i^2)^{-1}=(\sum_{i=0}^k\lambda_{\min}(\sum_{j=2i}^{2i+1}(b(j)\E[\widehat\L_{\G(j)}|\F(2i-1)]\otimes I_3+a(j)\E[\H^T(j)\H(j)|\F(2i-1)])))^{-1}$ over time $k$ with different settings in Table \ref{ATabf}.

\begin{figure}[htbp]
\centering
\subfloat[Setting I]{\includegraphics[width=.5\columnwidth]{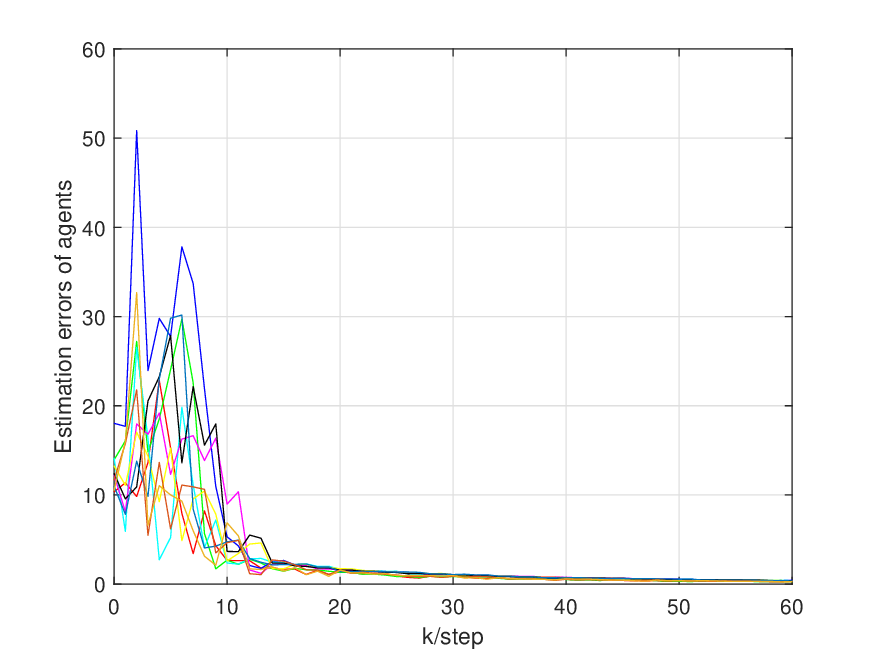}}\hspace{5pt}
\subfloat[Setting II]{\includegraphics[width=.5\columnwidth]{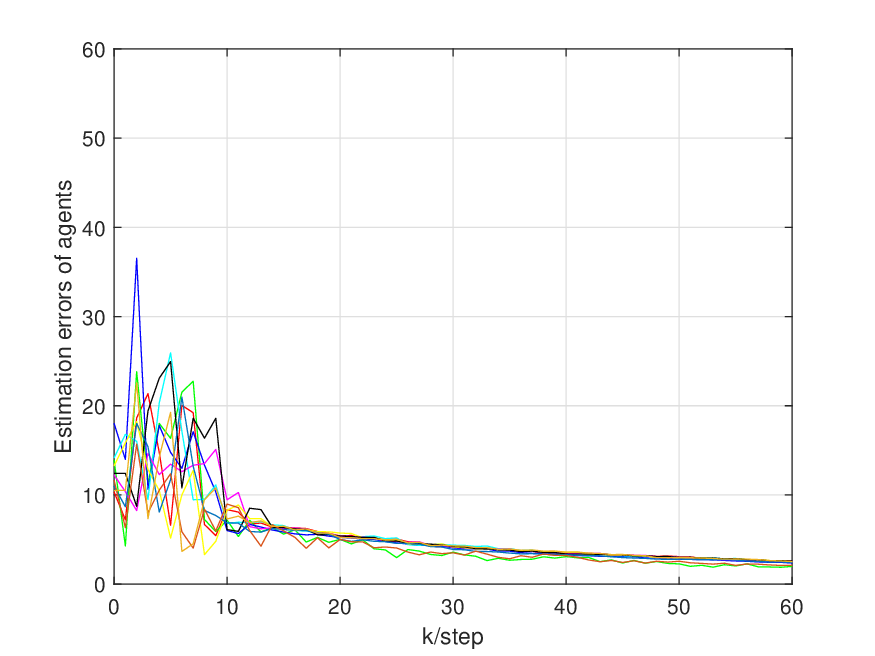}}\vspace{1pt}
\subfloat[Setting III]{\includegraphics[width=.5\columnwidth]{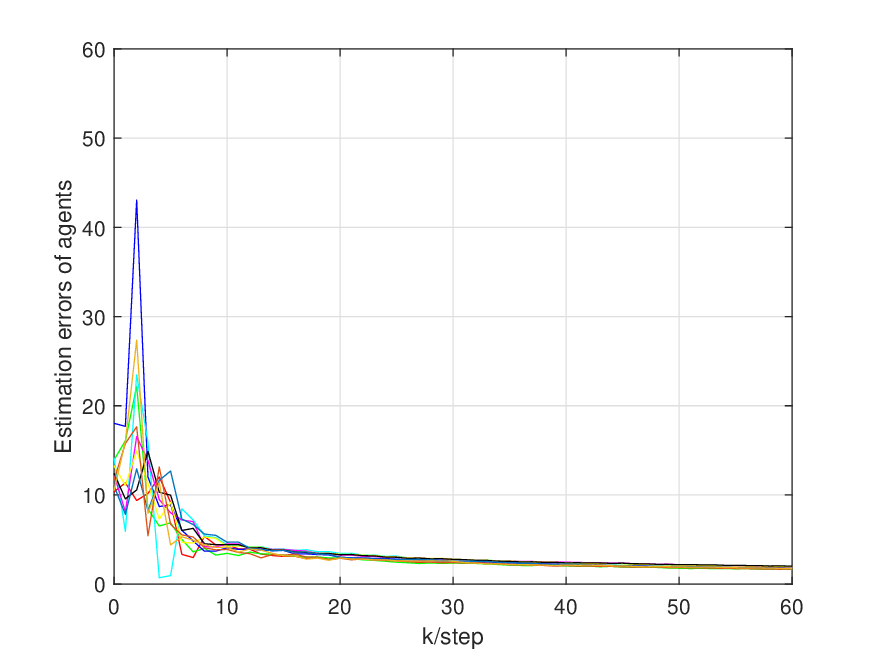}}\vspace{1pt}
\caption{The sample paths of estimation errors with different settings.}\label{AFigure1}
\end{figure}

Here, all the conditions of Theorem \ref{ATheorem1} hold with $h=2$ and $\rho_0=5$. The trajectories of the estimation errors are shown in Fig. \ref{AFigure1}. It can be seen that as time goes on, the estimations of each node's algorithms with different algorithm gains and regularization parameters in Table \ref{ATabf} converge to $x_0$ almost surely. Fig. \ref{Apeconditions} and Fig. \ref{AFigure1} show that (i) for the same gains $a(k)$ and $b(k)$, it is possible that the algorithm with larger regularization parameter $\lambda(k)$ converges faster than that with smaller one; (ii) for the same regularization parameter, larger $\sum_{i=0}^k\Lambda_i^2$ leads to faster convergence of the algorithm.

\begin{figure}[ht]
 \centering
 \includegraphics[width=.5\columnwidth]{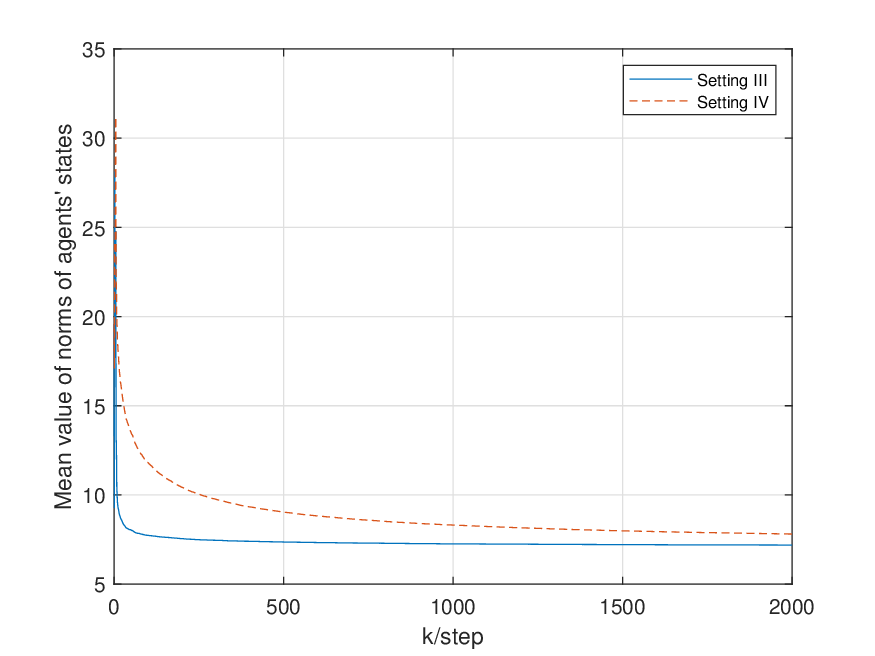}
 \caption{Mean value of norms of $10$ nodes' states.}\label{AFigure555}
\end{figure}

Meanwhile, the mean values of the norms of $10$ nodes' states are plotted in Fig. \ref{AFigure555}. We can see that compared with the non-regularized algorithms, the regularization is effective for reducing the magnitudes of estimations of the unknown true vector, which helps to reduce the model complexity and to avoid overfitting.

\begin{figure}[ht]
 \centering
 \includegraphics[width=.5\columnwidth]{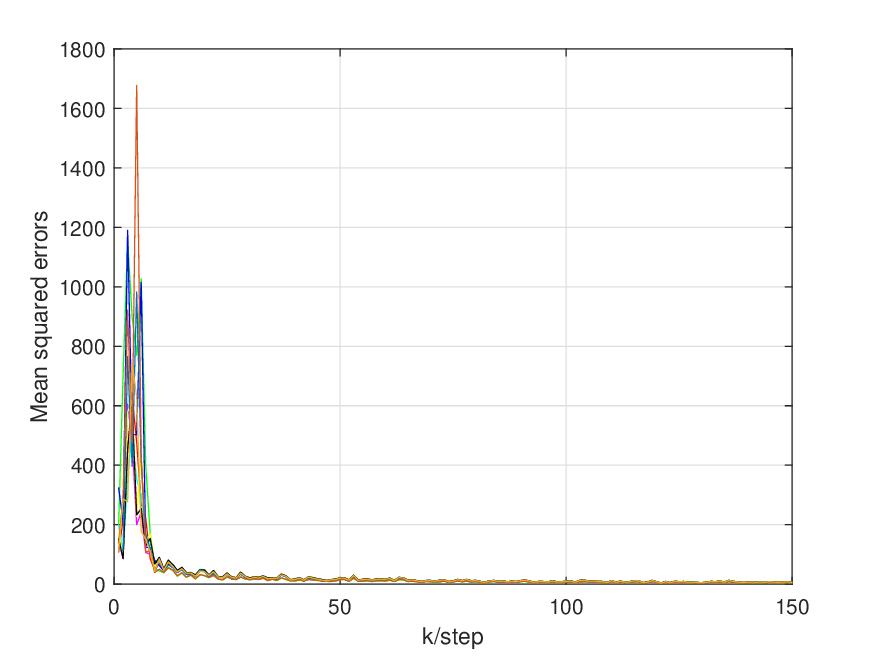}
 \caption{Mean square estimation errors of nodes for Setting III.}\label{AFigure2}
\end{figure}

Finally, all the conditions of Theorem \ref{Atheorem2} hold with Setting III in Table \ref{ATabf}. Fig. \ref{AFigure2} shows that the estimate of each node converges to $x_0$ in mean square.

\section{Conclusions}
We study the decentralized online regularized linear regression algorithm over random time-varying graphs.
For the generalized regression models and graphs, we obtain the sufficient conditions for the almost sure and mean square convergences of the algorithm. We prove that if the algorithm gains, the sequences of regression matrices and Laplacian matrices of the graphs jointly satisfy the \emph{sample path spatio-temporal persistence of excitation} condition, then the estimations of all nodes  converge to the unknown true parameter vector almost surely. Especially, if the graphs are \emph{conditionally balanced} and \emph{uniformly conditionally jointly connected}, and the regression models of all nodes  are \emph{uniformly conditionally spatio-temporally jointly observable}, then the \emph{sample path spatio-temporal persistence of excitation} condition holds and the estimations of all nodes converge to the unknown true parameter vector almost surely and in mean square by properly choosing the algorithm gains. To appraise the performance of the algorithm, we prove that the regret upper bound is $\mathcal O(T^{1-\tau}\ln T)$, where $\tau\in (0.5,1)$ is a constant depending on the algorithm gains.

There are a number of interesting open problems. For examples, (i) how to design the appropriate regularization parameter to achieve the optimal convergence rate for decentralized online regularized algorithm; (ii) It is interesting to develop tools for the case with more general state-dependent communication noises, i.e. the multiplicative term does not go to zero in communication model (\ref{Aeekksls}); (iii) it is interesting to study the  case with time-varying unknown parameters by combining our method with those in Xie and Guo \cite{Axiesiyu1}-\cite{Axiesiyu2}; (iv) the case with random time delays as in Wang \emph{et al.} \cite{Awjx}.

%There are a number of interesting problems for further investigation. For examples, (i) how to design the appropriate regularization parameter to achieve the optimal convergence rate for decentralized online regularized algorithm; (ii) for the case that the relative-state difference $x_j(k)-x_i(k)$ goes to zero but the multiplicative term does not go to zero in communication model (\ref{Aeekksls}), the method in this paper may not be applicable. It is interesting to develop effective tools to analyze the decentralized online learning algorithm with more general state-dependent communication noises; (iii) it is interesting to study the  decentralized online regression problems over random time-varying graphs with time-varying unknown parameters by combining our method with those in Xie and Guo \cite{Axiesiyu1}-\cite{Axiesiyu2}; (iv) it is possible to study the regularized algorithm with time delays by the techniques in Wang \emph{et al.} \cite{Awjx} and this paper.

\begin{appendices}
\section{Proofs of Theorems \ref{ATheorem1}-\ref{Atheorem4} and Verification of Example \ref{Aexample333}}\label{Aprooftheorem}
\setcounter{equation}{0}
\def\theequation{A.\arabic{equation}}
%\subsection{Proof of Theorem \ref{ATheorem1}}
\noindent\textbf{Proof of Theorem \ref{ATheorem1}.}
%Noting that $\Lambda_k^h\in \F(kh-1)$ and $V(kh)\in \F(kh-1)$, by the condition (i), we know that $\{V(kh),\F(kh-1),k\ge 0\}$ and $\{\Lambda_k^h-\sum_{i=kh}^{(k+1)h-1}\lambda^{\a}(i),\F(kh-1),k\ge 0\}$ are both nonnegative adaptive sequences. Then, by Lemma $\ref{Alemma4}$, there exists a positive integer $k_0$ and nonnegative deterministic real sequences $\{\Omega(k),k\ge 0\}$ and $\{\Gamma_{\a}(k),k\ge 0\}$ satisfying $\Omega(k)=\mathcal O(a^2(kh)+b^2(kh)+\lambda^2(kh))$ and  $\Gamma_{\a}(k)=\mathcal O(a^2(kh)+b^2(kh)+\lambda^{2-\a}(kh))$, such that
It follows from the condition (iii) and Lemma \ref{Alemma4} that there exists a positive integer $k_0$, such that
\ban
&&\hspace{-0.7cm}\E[V((k+1)h)|\F(kh-1)]\leq (1+\Omega(k))V(kh)-2\left(\Lambda_k^h+\sum_{i=kh}^{(k+1)h-1}\lambda(i)\right)V(kh)\cr
&&~~~~~~~~~~~~~~~~~~~~~~~~~~~~~~~+\Gamma(k)~\text{a.s.},~k\ge k_0,
\ean
where $\Omega(k)+\Gamma(k)=\mathcal O(a^2(kh)+b^2(kh)+\lambda(kh))$ is nonnegative.
It follows from the condition (i) and $V(k)=\|\d(k)\|^2$ that $\Lambda_k^h+\sum_{i=kh}^{(k+1)h-1}\lambda(i)$ and $V(kh)$ are both nonnegative and adapted to $\F(kh-1)$ for $k\ge k_0$.
By the condition (ii), we have
$$
  \sum_{k=0}^{\infty}\Omega(k)+\sum_{k=0}^{\infty}\Gamma(k)<\infty,
$$
which together with above and Theorem 1 in \cite{Arobbinssiegmund} leads to
$$
    \sum_{k=0}^{\infty}\left(\Lambda_k^h+\sum_{i=kh}^{(k+1)h-1}\lambda(i)\right)V(kh)<\infty~\text{a.s.},
$$
and $V(kh)$ converges almost surely. This along with the condition (i)   gives
$\liminf_{k\to \infty}V(kh)$ $=0~\text{a.s.}$, which further leads to
\bna\label{Aoolwlfw}
\lim_{k\to\infty}\d(kh)=0~\text{a.s.}
\ena
Denote $m_k=\lfloor \frac{k}{h} \rfloor$. For any $\varepsilon>0$, by Markov inequality and the conditions (ii)-(iii), we have
$$
\sum_{k=0}^{\infty}\mathbb P\{\|b(k)\mathcal L_{\mathcal G(k)}\otimes I_n+a(k){\mathcal H}^T(k){\mathcal H}(k)\|\ge\varepsilon\}
<\infty~\text{a.s.},
$$
which together with Borel-Cantelli Lemma gives  $$\mathbb P\{\|b(k)\mathcal L_{\mathcal G(k)}\otimes I_n+a(k){\mathcal H}^T(k){\mathcal H}(k)\|\ge \varepsilon~\text{i.o.}\}=0,$$ i.e., $\|b(k)\mathcal L_{\mathcal G(k)}\otimes I_n+a(k){\mathcal H}^T(k){\mathcal H}(k)\|$ converges to zero almost surely,   which gives $$\sup_{k\ge 0}\|b(k)\mathcal L_{\mathcal G(k)}\otimes I_n+a(k){\mathcal H}^T(k){\mathcal H}(k)\|<\infty~\text{a.s.}$$ Noting that $0\leq k-m_kh<h$, we have
\bna\label{Aldkjvn}
\Phi_0\triangleq\sup_{k\ge 0}\sup_{m_kh\leq i\leq k}\|\Phi_P(k-1,i)\|<\infty ~\text{a.s.}
\ena
By definitions of $W(k)$ and $M(k)$, we know that $\|W(k)\|\leq \sqrt{N}\|\A_{\G(k)}\|$ and $\|M(k)\|\leq \sqrt{4\sigma^2V(k)+2b^2}\leq 2\sigma \d(k)+\sqrt{2}b$, it follows from the estimation error equation (\ref{Aerror}) and  (\ref{Aldkjvn}) that
$$
\|\d(k)\|\leq f(k)+\sum_{i=m_kh}^{k-1}g(i)\|\d(i)\|,
$$
where \ban
\begin{cases}
f(k)=\Phi_0(\|\d(m_kh)\|+\displaystyle\sum_{i=m_kh}^{k-1}(a(i)\|v(i)\|\|\mathcal H^T(i)\|+b\sqrt{2N}b(i)\|\xi(i)\|\|\mathcal A_{\mathcal G(i)}\|+\sqrt{N}\|x_0\|\lambda(i))),\\
g(i)=2\sigma\sqrt{N}\Phi_0 b(i)\|\xi(i)\|\|\mathcal A_{\mathcal G(i)}\|.
\end{cases}
\ean
By Gronwall inequality, we get
  \bna\label{Afwqqqwww}
  \|\d(k)\|\leq f(k)+\sum_{i=m_kh}^{k-1}f(i)g(i)\prod_{j=i+1}^{k-1}(1+g(j)).
  \ena
For any $\varepsilon>0$, it follows from Assumption \textbf{(A2)}, the conditions (ii)-(iii) and Markov inequality that  $$\sum_{k=0}^{\infty}\mathbb P\{b(k)\|\xi(k)\|\|\mathcal A_{\mathcal G(k)}\|\ge \varepsilon\}<\infty ~\text{a.s.},$$ which shows that $b(k)\|\xi(k)\|\|\mathcal A_{\mathcal G(k)}\|$ converges to zero almost surely by Borel-Cantelli Lemma. Following the same way,  we know that $$\lim_{k\to\infty}a(k)\|v(k)\|\|\mathcal H^T(k)\|=0~\text{a.s.},$$
  %\ban
%  &&\hspace{-0.4cm}\sum_{k=0}^{\infty}\(\mathbb P\{a(k)\|v(k)\|\|\mathcal H^T(k)\|\ge \varepsilon\}\cr &&~~~~+\mathbb P\{b(k)\|\xi(k)\|\|\mathcal A_{\mathcal G(k)}\|\ge \varepsilon\}\)\cr &&\hspace{-0.7cm}\leq \sum_{k=0}^{\infty}\(\frac{\E[a^2(k)\|v(k)\|^2\|\mathcal H^T(k)\|^2]}{\varepsilon^2}\cr &&~~~~+\frac{\E[b^2(k)\|\xi(k)\|^2\|\mathcal A_{\mathcal G(k)}\|^2]}{\varepsilon^2}\)\cr
%  &&\hspace{-0.7cm}\leq \frac{\b_v\rho_0}{\varepsilon^2}\sum_{k=0}^{\infty}a^2(k)+\frac{N\b_v\rho^2_0}{\varepsilon^2}\sum_{k=0}^{\infty}b^2(k)\cr
%  &&\hspace{-0.7cm}<\infty~\text{a.s.},
%  \ean
which together with (\ref{Aoolwlfw})-(\ref{Aldkjvn}) gives $\lim_{k\to\infty}f(k)=0$ a.s. By (\ref{Aldkjvn}), we also have $\lim_{k\to\infty}$ $g(k)=0$ a.s. It follows from the above and  (\ref{Afwqqqwww}) that  $$\lim_{k\to\infty}\d(k)=0~\rm{a.s.}$$

\vskip 0.2cm
%\subsection{Proof of Theorem \ref{Atheorem2}}

\noindent\textbf{Proof of Theorem \ref{Atheorem2}.}
It follows from the conditions (i)-(ii) and Definitions \ref{Adefinition1}-\ref{Adefinition2}  that there exist positive constants $\theta_1$ and $\theta_2$, such that $\forall~k\ge 0$,  $$\lambda_2\left(\sum_{j=kh}^{(k+1)h-1}\E\left[\widehat\L_{\G(j)}|\F(kh-1)\right]\right)\ge \theta_1~\rm{a.s.}$$ and
$$\lambda_{\min}\left(\sum_{i=1}^N\sum_{j=kh}^{(k+1)h-1}\mathbb E\left[H_i^T(j)H_i(j)|\mathscr F(kh-1)\right]\right)\ge \theta_2~\rm{a.s.}$$ Denote
$$T_k=\frac{\theta_1\theta_2 \min\{a((k+1)h),b((k+1)h)\}}{2Nh\rho_0+N\theta_1}+\sum_{i=kh}^{(k+1)h-1}\lambda(i).
$$
By Condition \textbf{(C1)}, we have
$$
\sum_{k=0}^{\infty}T_k=\infty.
$$
Hence, by  Condition \textbf{(C1)}  and (I) of Lemma \ref{Alemma6}, we have proved (I) of Theorem $\ref{Atheorem2}$.

Next, we will prove (II) of Theorem $\ref{Atheorem2}$.
It follows from Condition \textbf{(C2)} that
$$ \sum_{k=0}^{\infty}T_k=\infty.$$
Noting that $a(k)=\mathcal O(a(k+1))$ and $b(k)=\mathcal O(b(k+1))$, we know that there exist positive constants $C_1$ and $C_2$, such that $$a(k)\leq C_1a(k+1),\quad b(k)\leq C_2b(k+1).$$
Denote $C=\max\{C_1,C_2\}$, we have $$\min\{a(k),b(k)\}\leq C\min\{a(k+1),b(k+1)\},$$ which leads to
\ban
&&~~~\frac{\min\{a(kh),b(kh)\}}{\min\{a((k+1)h),b((k+1)h)\}}\cr &&=\prod_{i=0}^{h-1}\left[\frac{\min\{a(kh+i),b(kh+i)\}}{\min\{a(kh+i+1),b(kh+i+1)\}}\right]\leq C^h,
\ean
thus, we obtain  %$(\min\{a((k+1)h),b((k+1)h)\})^{-1}(a^2(kh)+b^2(kh)+\lambda(kh))\leq (\min\{a((k+1)h),b((k+1)h)\})^{-1}\min\{a(kh),b(kh)\}(\min\{a(kh),b(kh)\})^{-1}(a^2(kh)+b^2(kh)+\lambda(kh))\leq C^h(\min\{a(kh),b(kh)\})^{-1}(a^2(kh)+b^2(kh)+\lambda(kh))$,
$$
\min\{a(kh),b(kh)\}\leq C^h\min\{a((k+1)h),b((k+1)h)\}
,$$
%which together with $(\ref{Appwddfff})$ gives $T_k^{-1}(a^2(kh)+b^2(kh)+\lambda(kh))=\mathcal O((\min\{a(kh),b(kh)\})^{-1}(a^2(kh)+b^2(kh)+\lambda(kh)))$,
%  \bna\label{Avnkdkllwe}
%  &&\hspace{-0.6cm}\frac{a^2(kh)+b^2(kh)+\lambda(kh)}{T_k}\cr
%  &&\hspace{-1cm}\leq \frac{4Nh\rho_0+2N\theta_1}{\theta_1\theta_2}\times\frac{a^2(kh)+b^2(kh)+\lambda(kh)}{\min\{a((k+1)h),b((k+1)h)\}}\cr
%  &&\hspace{-1cm}\leq \frac{C^h(4Nh\rho_0+2N\theta_1)}{\theta_1\theta_2}\times \frac{a^2(kh)+b^2(kh)+\lambda(kh)}{\min\{a(kh),b(kh)\}}.
%  \ena
which together with Condition \textbf{(C2)} gives
$$a^2(kh)+b^2(kh)+\lambda(kh)=o(T_k).$$
It follows from (II) of Lemma \ref{Alemma6} that $$\lim_{k\to\infty}\E[V(k)]=0.$$
\vskip 0.2cm
%\subsection{Proof of Corollary \ref{Acorollary1}}
\noindent\textbf{Verification of Example \ref{Aexample333}.}
By Assumption \textbf{(2)}, we have $\E[\H^T\H|\F(kh-1)]=\E[\H^T\H|\sigma(\H)]=\H^T\H$ a.s. For any given positive integer $h$, by the definition of Laplacian matrix, we get
\bna\label{Acxk1}
\Lambda_k^h=\sum_{i=kh}^{(k+1)h-1}\frac{x+1-\sqrt{x^2-2x+2}}{2(i+1)}~\mathrm{a.s.},~\forall k\ge 0.
\ena
Noting that $x$ is uniformly distributed in $(0.25,1.25)$, we obtain $x+1-\sqrt{x^2-2x+2}>0$ a.s., which together with (\ref{Acxk1}) gives
\ban
\sum_{k=0}^{\infty}\Lambda_k^h=
\left(x+1-\sqrt{x^2-2x+2}\right)\sum_{k=0}^{\infty}\sum_{i=kh}^{(k+1)h-1}\frac{1}{2(i+1)}=\infty~\text{a.s.}
\ean
Suppose that there exists a positive real sequence $\{c(k),k\ge 0\}$ satisfying $\Lambda_k^h\ge c(k)$ a.s. For any given integer $k_0>0$, denote $\mu=\frac{(k_0h+1)c(k_0)}{h}$. It follows from $\Lambda_k^h\ge c(k)$ a.s. that
\bna\label{Acxk2}
0<\mu\leq \frac{x+1-\sqrt{x^2-2x+2}}{2}<1~\mathrm{a.s.},
\ena
which leads to $x\ge \mu+\frac{1}{4(1-\mu)}$ a.s. Hence, by (\ref{Acxk2}), we have
\ban
\P\left\{\Lambda_{k_0}^h\ge c(k_0)\right\}&\leq& \P\left\{\mu+\frac{1}{4(1-\mu)}\leq x\leq \frac{5}{4}\right\}\cr
&=&\frac{5}{4}-\mu-\frac{1}{4(1-\mu)}<1,
\ean
which is contradictory to $\Lambda_k^h\ge c(k)$ a.s.
\vskip 0.2cm

\noindent\textbf{Proof of Theorem \ref{Atheorem3}.}
It follows from $\{\mathcal G(k),k\ge 0\}\in \Gamma_1$ that
\ban
&&~~\sum_{i=kh}^{(k+1)h-1}\mathbb E\left[b(i)\widehat {\mathcal L}_{\mathcal G(i)}\otimes I_n+a(i){\mathcal H}^T(i){\mathcal H}(i)|\F(kh-1)\right]\cr
&&\ge \min\{a((k+1)h),b((k+1)h)\}\sum_{i=kh}^{(k+1)h-1}\mathbb E\left[\widehat {\mathcal L}_{\mathcal G(i)}\otimes I_n+{\mathcal H}^T(i){\mathcal H}(i)|\F(kh-1)\right]~ \text{a.s.}
\ean
It follows from the conditions (i)-(iii), Definitions \ref{Adefinition1}-\ref{Adefinition2} and Lemma $\ref{Alemma5}$  that there exist positive constants $\theta_1$ and $\theta_2$, such that
$$
\Lambda_k^h \ge (2Nh\rho_0+N\theta_1)^{-1}\theta_1\theta_2\min\{a((k+1)h),b((k+1)h)\}~\text{a.s.}
$$
Denote
$$L(k)=(2Nh\rho_0+N\theta_1)^{-1}\theta_1\theta_2\min\{a((k+1)h),b((k+1)h)\}+\sum_{i=kh}^{(k+1)h-1}\lambda(i),$$
we have
$$
\Lambda_k^h+\sum_{i=kh}^{(k+1)h-1}\lambda(i)\ge L(k)~ \text{a.s.},$$
which together with Lemma \ref{Alemma4} shows that there exists a positive integer $k_0$, such that
\bna\label{Awwddvbbgg}
\E[V((k+1)h)|\F(kh-1)]\leq \left(1+\Omega(k)\right)V(kh)-2L(k)V(kh)+\Gamma(k)~\text{a.s.},~k\ge k_0,
\ena
where $\Omega(k)+\Gamma(k)=\mathcal O(a^2(kh)+b^2(kh)+\lambda(kh))$, which together with the choice of the algorithm gains shows that there exists a positive constant $u$ such that $$\Gamma(k)\leq u(kh+1)^{-2\tau}, ~k\ge 0.$$ Noting that $\Omega(k)=o(L(k))$ and $L(k)=o(1)$, without loss of generality, we suppose that $0<\Omega(k)\leq L(k)<1$, $k\ge k_0$. Denote $$v=(4Nh\rho_0+2N\theta_1)^{-1}c_1\theta_1\theta_2,$$ it follows that $$L(k)\ge
v(kh+h+1)^{-\tau}, ~k\ge k_0.$$ By taking mathematical expectation on both sides of (\ref{Awwddvbbgg}), we get
\bna\label{ARA7}
&&\hspace{-0.8cm}\mathbb{E}[V((k+1)h)]\cr
%&&\hspace{-0.7cm}\leq (1-L(k))\mathbb{E}[V(kh)]+\Gamma(k)\cr
%&&\hspace{-0.7cm}\leq \prod_{i=k_0}^k(1-L(i))\mathbb{E}[V(k_0h)]+\sum_{i=k_0}^k\Gamma(i)\prod_{j=i+1}^k(1-L(j))\cr
%&&\hspace{-0.7cm}\leq \prod_{i=k_0+1}^k(1-L(i))\mathbb{E}[V(k_0h)]+\sum_{i=k_0}^k\Gamma(i)\prod_{j=i+1}^k(1-L(j))\cr
&&\hspace{-1.5cm}\leq \prod_{i=k_0+1}^k\left(1-\frac{v}{(ih+h+1)^{\tau}}\right)\sup_{k\ge 0}\E[V(k)]\cr &&\hspace{-1.5cm}+\sum_{i=k_0}^k\frac{u}{(ih+1)^{2\tau}} \prod_{j=i+1}^k\left(1-\frac{v}{(jh+h+1)^{\tau}}\right),~k>k_0.
\ena
Noting that
\ban
&&~~\sum_{j=i+1}^k\frac{v}{(jh+h+1)^{\tau}}\cr
&&\ge \int_{i+1}^{k}\frac{v}{(xh+h+1)^{\tau}}dx\cr
&&=\frac{v}{h(1-\tau)}\left((kh+h+1)^{1-\tau}-(ih+2h+1)^{1-\tau}\right),~ i\ge 0,
\ean
we have
\bna\label{Afnmvmmcc}
&&\hspace{-0.35cm}\prod_{j=i+1}^k\left(1-\frac{v}{(jh+h+1)^{\tau}}\right)\cr
&&\hspace{-0.7cm}\leq \exp\bigg(-\frac{v}{h(1-\tau)}\big((kh+h+1)^{1-\tau}-(ih+2h+1)^{1-\tau}\big)\bigg),~ i \leq k.
\ena
By Theorem \ref{Atheorem2}, we have $\lim_{k\to\infty}\E[V(k)]=0$, which shows that $$\sup_{k\ge 0}\E[V(k)]<\infty.$$ It follows from (\ref{Afnmvmmcc}) that $$\exp\left(-\frac{v}{h(1-\tau)}(kh+h+1)^{1-\tau}\right)=o\left((kh+h)^{-\tau}\ln(kh+h)\right),$$ thus, we have
\bna\label{Aeewddd}
\prod_{i=k_0+1}^k\left(1-\frac{v}{(ih+h+1)^{\tau}}\right)\sup_{k\ge 0}\E[V(k)]=o\left((kh+h)^{-\tau}\ln(kh+h)\right).
\ena
Denote $$\epsilon (k)=\left\lceil \frac{2}{v}(kh+h+1)^{\tau}\ln (kh+h+1)\right\rceil,$$ then we have $\epsilon(k)=o(k)$. Noting that $\epsilon (k)\to \infty$ as $k\to \infty$, without loss of generality, we suppose that $k_0<\epsilon (k)\leq 2\epsilon (k)\leq k$. On one hand, for the case with $k_0\leq i \leq k-1-\epsilon(k)$, we know that
\bna\label{Axllsddd}
ih+2h+1\leq kh+h+1-\epsilon_1(k),
\ena
where $\epsilon_1(k)=\lceil \frac{2h}{v}(kh+h+1)^{\tau}\ln (kh+h+1)\rceil$,
which directly gives
$$
(kh+h+1)^{1-\tau}-(kh+h+1-\epsilon_1(k))^{1-\tau}\ge (kh+h+1)^{-\tau}\epsilon_1(k)(1-\tau)\ge v^{-1}2h(1-\tau)\ln (kh+h+1).
$$
This together with (\ref{Axllsddd}) gives
$$
(h(1-\tau))^{-1}v\left((kh+h+1)^{1-\tau}-(ih+2h+1)^{1-\tau}\right)\ge 2\ln (kh+h+1).
$$
Then, it follows from (\ref{Afnmvmmcc}) that
\bna\label{Aankkdd}
\sum_{i=k_0}^{k-1-\epsilon(k)}\frac{u}{(ih+1)^{2\tau}}\prod_{j=i+1}^k\left(1-\frac{v}{(jh+h+1)^{\tau}}\right)=\mathcal O\left(k^{-1}\right).
\ena
On the other hand, for the case with $k-\epsilon(k)\leq i\leq k$, we have $k\leq 2k-2\epsilon(k)\leq 2i$, which shows that
$$
\frac{u}{(ih+1)^{2\tau}}\leq \frac{4^{\tau}u}{(kh+2)^{2\tau}},~k-\epsilon(k)\leq i\leq k.
$$
Then it follows from $$\prod_{j=i+1}^k\left(1-\frac{v}{(jh+h+1)^{\tau}}\right)\leq 1$$ that
\bna\label{Ackllss}
\sum_{i=k-\epsilon(k)}^k\frac{u}{(ih+1)^{2\tau}}\prod_{j=i+1}^k\left(1-\frac{v}{(jh+h+1)^{\tau}}\right)=\mathcal O\left((kh+h)^{-\tau}\ln (kh+h)\right).
\ena
Thus, by (\ref{Aankkdd})-(\ref{Ackllss}), we obtain
\bna\label{Aiiekke}
\sum_{i=k_0}^k\frac{u}{(ih+1)^{2\tau}}\prod_{j=i+1}^k\left(1-\frac{v}{(jh+h+1)^{\tau}}\right)=\mathcal O\left((kh+h)^{-\tau}\ln (kh+h)\right).
\ena
Combining (\ref{ARA7}), (\ref{Aeewddd}) and (\ref{Aiiekke}) directly shows that
$$
\mathbb E[V(kh)]=\mathcal O((kh+h)^{-\tau}\ln (kh+h)).
$$
By Lemma \ref{Alemma2}, there exists a positive integer $k_1$ such that
$$
\mathbb{E}[V(k)]\leq 2^h\mathbb{E}[V(m_kh)]+h2^h\gamma(m_kh),~k\ge k_1,
$$
where $\gamma(k)=\mathcal O((k+1)^{-2\tau})$ and $m_k=\lfloor \frac{k}{h}\rfloor$. Thus, we get
$$
\sum_{t=0}^T\E[V(t)]=\mathcal O\left(\sum_{t=0}^T(t+1)^{-\tau}\ln (t+1)\right).
$$
Noting that $$\sum_{t=0}^T(t+1)^{-\tau}\ln (t+1)=\mathcal O\left(\int_{0}^T(x+1)^{-\tau}\ln (x+1)dx\right),$$ we get (\ref{Ayiyiyiy}).
\vskip 0.2cm

\noindent\textbf{Proof of Theorem \ref{Atheorem4}.}
It follows from $\{\mathcal G(k),k\ge 0\}\in \Gamma_1$ that $\E[\widehat \L_{\G(k)}|\F(k-1)]$ is positive semi-definite. Noting that $$\E\left[\widehat \L_{\G(k)}|\F(mh-1)\right]=\E\left[\E\left[\widehat \L_{\G(k)}|\F(k-1)\right]|\F(mh-1)\right], ~k\ge mh,$$ we know that $\E[\widehat \L_{\G(k)}|\F(mh-1)]$ is also positive semi-definite, which shows that
\ban
&&~~\sum_{i=kh}^{(k+1)h-1}\mathbb E\left[b(i)\widehat {\mathcal L}_{\mathcal G(i)}\otimes I_n+a(i){\mathcal H}^T(i){\mathcal H}(i)|\F(kh-1)\right]\cr
&&\ge \min\{a((k+1)h),b((k+1)h)\}\sum_{i=kh}^{(k+1)h-1}\mathbb E\left[\widehat {\mathcal L}_{\mathcal G(i)}\otimes I_n+{\mathcal H}^T(i){\mathcal H}(i)|\F(kh-1)\right]~ \text{a.s.}
\ean
It follows from the conditions (i)-(ii) and Definitions \ref{Adefinition1}-\ref{Adefinition2}  that there exist positive constants $\theta_1$ and $\theta_2$, such that $$\inf_{k\ge 0}\lambda_2\left(\sum_{j=kh}^{(k+1)h-1}\E\left[\widehat\L_{\G(j)}|\F(kh-1)\right]\right)\ge \theta_1~\rm{a.s.}$$ and
$$\inf_{k\ge 0}\lambda_{\min}\left(\sum_{i=1}^N\sum_{j=kh}^{(k+1)h-1}\mathbb E\left[H_i^T(j)H_i(j)|\mathscr F(kh-1)\right]\right)\ge \theta_2~\rm{a.s.},$$ respectively, which together with the condition (iii) and Lemma \ref{Alemma5} gives
$$
\Lambda_k^h \ge (2Nh\rho_0+N\theta_1)^{-1}\theta_1\theta_2\min\{a((k+1)h),b((k+1)h)\}~\text{a.s.}
$$
Denote
$$L(k)=(2Nh\rho_0+N\theta_1)^{-1}\theta_1\theta_2\min\{a((k+1)h),b((k+1)h)\}+\sum_{i=kh}^{(k+1)h-1}\lambda(i),$$
we have
\bna\label{A9999}
\Lambda_k^h+\sum_{i=kh}^{(k+1)h-1}\lambda(i)\ge L(k)~ \text{a.s.}
\ena
By the proof of Lemma \ref{Alemma4}, we have
\ban
\Omega(k)&=&(h+1)(9^h-1-4h)(\rho_0a(kh)+\lambda(kh))^2+2^{h+3}hN^2\b_v\sigma^2\rho^2_0a^2(kh)\cr &&+2^{h+5}hN^2\b_v\sigma^2\rho^2_0a^2(kh)+\sum_{i=kh}^{(k+1)h-1}\lambda(i)\cr
&\leq& \frac{(9^h-4h-1)(h+1)(\rho_0+1)^2+2^{h+3}hN^2\beta_v\sigma^2\rho_0^2+2^{h+5}hN^2\beta_v\sigma^2\rho_0^2+h}{(kh+1)^{2\tau}}\cr &\triangleq& \frac{c_1}{(kh+1)^{2\tau}},~\forall~k\ge 0.
\ean
Noting that
\[L(k)\ge \frac{\theta_1\theta_2}{(2Nh\rho_0+N\theta_1)((k+1)h+1)^{\tau}}\triangleq \frac{c_2}{((k+1)h+1)^{\tau}},~\forall~k\ge 0,\]
and
\[L(k)\leq \frac{\theta_1\theta_2}{(2Nh\rho_0+N\theta_1)(kh+1)^{\tau}}+\frac{h}{(kh+1)^{\tau}}\triangleq \frac{c_3}{(kh+1)^{\tau}},~\forall~k\ge 0.\]
Denote
\[k_0=\left\lfloor \frac{4c_1^2}{c_2^2}+c_3^2  \right\rfloor+1.\]
We can verify that $0<\Omega(k)\leq L(k)<1$, $k\ge k_0$. By (\ref{A9999}) and the proof of Lemma \ref{Alemma4}, we obtain
\bna\label{Awwddvbbggg}
\E[V((k+1)h)|\F(kh-1)]\leq \left(1+\Omega(k)\right)V(kh)-2L(k)V(kh)+\Gamma(k)~\text{a.s.},~k\ge k_0,
\ena
where
\ban
\Gamma(k)&=&2h\rho_0\b_va^2(kh)+(2^{h+3}h\sigma^2+4b^2)hN^2\b_v\rho^2_0 b^2(kh)+8ha(kh)b(kh)(\b_v\rho_0 \cr &&+N^2\b_v\rho^2_0(2^{h+2}h\sigma^2+2b^2))+2hN\|x_0\|^2\lambda(kh)+2Nh^2\|x_0\|^2\lambda^2(kh)\cr
&\leq&2h\rho_0\b_va^2(kh)+(2^{h+3}h\sigma^2+4b^2)hN^2\b_v\rho^2_0 a^2(kh)+8ha^2(kh)(\b_v\rho_0 \cr &&+N^2\b_v\rho^2_0(2^{h+2}h\sigma^2+2b^2))+2hN\|x_0\|^2a^2(kh)+2Nh^2\|x_0\|^2a^2(kh)\cr
&\triangleq&\frac{c_4}{(kh+1)^{2\tau}},~\forall~k\ge 0.
\ean
Denote $$c_5=((2Nh{\max\{\rho_0,1\}}+N\theta_1)h^{\tau})^{-1}{\min\{\theta_1,1\}}{\min\{\theta_2,1\}},$$ it follows that $$L(k)\ge
c_5(kh+h+1)^{-\tau}, ~k\ge k_0.$$ By taking mathematical expectation on both sides of (\ref{Awwddvbbggg}), we get
\bna\label{ARA77}
\mathbb{E}[V((k+1)h)]\leq \left(1-\frac{c_5}{(kh+h+1)^{\tau}}\right)\mathbb{E}[V(kh)]+\frac{c_4}{(kh+1)^{2\tau}},~\forall~k>k_0.
\ena
By Theorem \ref{Atheorem2}, we have $\lim_{k\to\infty}\E[V(k)]=0$, which shows that there exists a constant $c_6>0$, such that $\sup_{k\ge 0}\E[V(k)]\leq c_6$, which together with Lemma 2.9 in \cite{ALeonard} gives
\[\E[V(kh)]\leq \frac{25c_4\ln (k+1)}{c_5(k+1)^{\tau}}+c_6e^{-\frac{c_5\left((k+1)^{1-\tau}-(k_0+2)^{1-\tau}\right)}{1-\tau}},\]
where
\[k\ge \left\lfloor \left(\frac{12}{c_5(1-\tau)}\ln \left(\frac{4}{c_5(1-\tau)}\right)\right)^{\frac{1}{1-\tau}}\right\rfloor+1+k_0.\]
Denote $m_k=\lfloor \frac{k}{h}\rfloor$, $\forall~k\ge 0$. Noting that $m_kh\leq k<(m_k+1)h$, then for
\ban
k\ge \left\lfloor \left(\frac{12}{c_5(1-\tau)}\ln \left(\frac{4}{c_5(1-\tau)}\right)\right)^{\frac{1}{1-\tau}}+ \frac{4c_1^2}{c_2^2}+c_3^2  \right\rfloor+2,
\ean
it follows from (\ref{ARA77}) and Lemma \ref{Alemma2} that
\ban
\mathbb{E}[V(k)]&\leq& 2^h\mathbb{E}[V(m_kh)]+h2^h\gamma(m_kh)\cr
&\leq&2^h \left(\frac{25c_4\ln ( m_k+1)}{c_5(m_k+1)^{\tau}}+c_6e^{-\frac{c_5\left((m_k+1)^{1-\tau}-(k_0+2)^{1-\tau}\right)}{1-\tau}}\right)+h2^h\gamma(m_kh)\cr
&=&2^h \left(\frac{25c_4\ln ( m_k+1)}{c_5(m_k+1)^{\tau}}+c_6e^{-\frac{c_5\left((m_k+1)^{1-\tau}-(k_0+2)^{1-\tau}\right)}{1-\tau}}\right)+h2^h\big(4b^2a^2(m_kh)N^2\rho^2_0\b_v\cr
&&+2a^2(m_kh)\rho_0\b_v+(N\|x_0\|^2+a^2(m_kh)\rho^2_0)\lambda(m_kh)+N\|x_0\|^2\lambda^2(m_kh)\big)\cr
&\leq&2^h \left(\frac{25c_4\ln ( m_k+1)}{c_5(m_k+1)^{\tau}}+c_6e^{-\frac{c_5\left((m_k+1)^{1-\tau}-(k_0+2)^{1-\tau}\right)}{1-\tau}}\right)+h2^h\big(4b^2a^2(m_kh)N^2\rho^2_0\b_v\cr
&&+2a^2(m_kh)\rho_0\b_v+(N\|x_0\|^2+\rho^2_0)a^2(m_kh)+N\|x_0\|^2a^2(m_kh)\big)\cr
&\leq&2^h \left(\frac{25c_4\ln (k+1)} {c_5\left(\frac{k}{2h}\right)^{\tau}}+c_6e^{-\frac{c_5\left(\left(\frac{k}{2h}\right)^{1-\tau}-(k_0+2)^{1-\tau}\right)}{1-\tau}}\right)+h2^h\Bigg(4b^2a^2\left(\frac{k}{2}\right)N^2\rho^2_0\b_v\cr
&&+2a^2\left(\frac{k}{2}\right)\rho_0\b_v+(N\|x_0\|^2+\rho^2_0)a^2\left(\frac{k}{2}\right)+N\|x_0\|^2a^2\left(\frac{k}{2}\right)\Bigg)\cr
&\triangleq& 2^h \left(\frac{25c_4\ln (k+1)} {c_5\left(\frac{k}{2h}\right)^{\tau}}+c_6e^{-\frac{c_5\left(\left(\frac{k}{2h}\right)^{1-\tau}-\left(k_0+2\right)^{1-\tau}\right)}{1-\tau}}\right)+\frac{c_7}{(k+2)^{2\tau}}\cr
&\leq& 2^h \left(\frac{25c_4\ln (k+1)} {c_5\left(\frac{k}{2h}\right)^{\tau}}+c_6e^{-\frac{c_5\left(\left(\frac{k}{2h}\right)^{1-\tau}-\left(k_0+2\right)^{1-\tau}\right)}{1-\tau}}\right)+\frac{c_7}{k^{2\tau}}.
\ean

\section{Proofs of Lemmas \ref{Alemma4}-\ref{Alemma5}}\label{Aprooflemma4}
\setcounter{equation}{0}
\def\theequation{B.\arabic{equation}}
%\subsection{Proof of Lemma \ref{Alemma4}}
\noindent\textbf{Proof of Lemma \ref{Alemma4}.}
By the estimation error equation (\ref{Aerror}), we have
  \bna\label{Aerror2}
  \E[V((k+1)h)|\F(kh-1)]=\E\left[\sum_{i=1}^4A_i^T(k)A_i(k)+2\sum_{1\leq i< j\leq 4}A^T_i(k)A_j(k)|\F(kh-1)\right],
  \ena
  where
  $A_1(k)=\Phi_P((k+1)h-1,kh)\d(kh)$,
  $A_2(k)=\sum_{i=kh}^{(k+1)h-1}a(i)\Phi_P((k+1)h-1,i+1)\mathcal H^T(i)v(i)$,
  $A_3(k)=\sum_{i=kh}^{(k+1)h-1}b(i)\Phi_P((k+1)h-1,i+1)W(i)M(i)\xi(i)$, and
  $A_4(k)=\sum_{i=kh}^{(k+1)h-1}\lambda(i)\Phi_P((k+1)h-1,i+1)(\textbf{1}_{N}\otimes x_0)$.

We now consider the RHS of (\ref{Aerror2}) term by term. It follows from Assumption \textbf{(A2)} that  $\Phi^T_P((k+1)h-1,kh)\Phi_P((k+1)h-1,i+1)\H^T(i)$ and $v(i)$ are independent, $kh\leq i\leq (k+1)h-1$, which further shows that $\Phi^T_P((k+1)h-1,kh)\Phi_P((k+1)h-1,i+1)\H^T(i)$ and $v(i)$ are conditionally independent w.r.t. $\F(kh-1)$ by Lemma A.1 in \cite{Alw}. Noting that  $\d(kh)\in \F(kh-1)$, we have
  \bna\label{Ajkddsss}
\E\left[A_1^T(k)A_2(k)|\F(kh-1)\right]=0.
  \ena
  Similarly, we also have
  \bna\label{Aedsdddd}
\E\left[A_1^T(k)A_3(k)+A_2^T(k)A_4(k)+A_3^T(k)A_4(k)|\F(kh-1)\right]=0.
  \ena
By Lemma \ref{Alemma3}, there exist positive integers $k_2$ and $k_3$, such that
 \bna\label{Afffwwsffsefefe}
  &&\hspace{-0.35cm}\E\left[A_1^T(k)A_4(k)|\F(kh-1)\right]\cr
  &&\hspace{-0.7cm}\leq \frac{1}{2}\left(\sum_{i=kh}^{(k+1)h-1}\lambda(i)\right)(1+p(k))V(kh)+Nh\|x_0\|^2\lambda(kh)~\text{a.s.},~k\ge \max\{k_2,k_3\},
  \ena
  where $p(k)=(9^h-1-4h)(\rho_0\max\{a(kh),b(kh)\}+\lambda(kh))^2$. For $kh\leq i\neq j \leq (k+1)h-1$, by Assumption \textbf{(A2)} and Lemma A.1 in \cite{Alw}, we know that
\bna\label{Anxccdd}
  &&\hspace{-0.35cm}\E\big[v^T(i)\H(i)\Phi_P^T((k+1)h-1,i+1)\cr
  &&\hspace{-0.4cm}\times\Phi_P((k+1)h-1,j+1) W(j)M(j)\xi(j)|\F(kh-1)\big]\cr
%  &&\hspace{-0.7cm}=\E\big[\E\big[v^T(i)\H(i)\Phi_P^T((k+1)h-1,i+1)\cr
%  &&\hspace{-0.4cm}\times\Phi_P((k+1)h-1,j+1) W(j)M(j)\cr
%  &&\hspace{-0.4cm}\times\xi(j)|\F(j)\big]|\F(kh-1)\big]\cr
%  &&\hspace{-0.7cm}=\E\big[\E\big[v^T(i)\H(i)\Phi_P^T((k+1)h-1,i+1)\cr
%  &&\hspace{-0.4cm}\times\Phi_P((k+1)h-1,j+1)|\F(j)\big] W(j)M(j)\cr
%  &&\hspace{-0.4cm}\times\xi(j)|\F(kh-1)\big]\cr
%  &&\hspace{-0.7cm}=\E\big[\E\big[v^T(i)|\F(j)]\E[\H(i)\Phi_P^T((k+1)h-1,i+1)\cr
%  &&\hspace{-0.4cm}\times\Phi_P((k+1)h-1,j+1)|\F(j)\big]W(j)M(j)\cr
%  &&\hspace{-0.4cm}\times\xi(j)|\F(kh-1)\big]\cr
  &&\hspace{-0.7cm}=0.
\ena
Therefore, by (\ref{Anxccdd}), Assumptions \textbf{(A1)}-\textbf{(A2)}, Lemmas \ref{Alemma2}-\ref{Alemma3} and Lemma A.1 in \cite{Alw}, there exists a positive integer $k_1$, such that
  \bna\label{A23}
  &&\hspace{-1cm}\E\big[A_2^T(k)A_3(k)|\F(kh-1)\big]\cr
%  &&\hspace{-0.7cm}=\sum_{i,j=kh}^{(k+1)h-1}a(i)b(j)\E\big[v^T(i)\H(i)\Phi_P^T((k+1)h-1,i+1)\cr &&\hspace{-0.4cm}\times\Phi_P((k+1)h-1,j+1)W(j)M(j)\xi(j)|\F(kh-1)\big]\cr
%  &&\hspace{-0.7cm}=\Bigg(\sum_{j<i}+\sum_{i<j}+\sum_{i=j}\Bigg)a(i)b(j)\E\big[v^T(i)\H(i)\cr
%  &&\hspace{-0.4cm}\times\Phi_P^T((k+1)h-1,i+1) \Phi_P((k+1)h-1,j+1)\cr
%  &&\hspace{-0.4cm}\times W(j)M(j)\xi(j)|\F(kh-1)\big]\cr
%  &&\hspace{-0.7cm}=\sum_{i=kh}^{(k+1)h-1}a(i)b(i)\E\big[v^T(i)\H(i)\Phi_P^T((k+1)h-1,i+1)\cr &&\hspace{-0.4cm}\times\Phi_P((k+1)h-1,i+1)W(i)M(i)\xi(i)|\F(kh-1)\big]\cr
%  &&\hspace{-0.7cm}\leq \sum_{i=kh}^{(k+1)h-1}a(i)b(i) \E\Big[\big\|\E\big[\Phi_P^T((k+1)h-1,i+1)\cr
%  &&\hspace{-0.4cm}\times\Phi_P((k+1)h-1,i+1)|\F(i)\big]\big\|\cr
%  &&\hspace{-0.4cm}\times\big\|\H^T(i)v(i)+W(i)M(i)\xi(i)\big\|^2|\F(kh-1)\Big]\cr
%  &&\hspace{-0.7cm}=4 \sum_{i=kh}^{(k+1)h-1}a(i)b(i)\E\big[\E\big[\|\H(i)\|^2|\F(i-1)\big]\cr
%  &&\hspace{-0.4cm}\times \E\big[\|v(i)\|^2|\F(i-1)\big]+\|M(i)\|^2\E[\|W(i)\|^2|\F(i-1)\big]\cr
%  &&\hspace{-0.4cm}\times\E\big[\|\xi(i)\|^2|\F(i-1)]|\F(kh-1)\big]\cr
%  &&\hspace{-0.7cm}\leq 4\sum_{i=kh}^{(k+1)h-1}a(i)b(i)\big(\b_v\rho_0+N^2\b_v\rho^2_0\cr
%  &&\hspace{-0.4cm}\times\left(4\sigma^2\E[V(i)|\F(kh-1)]+2b^2\right)\big)\cr
%  &&\hspace{-0.7cm}\leq 4\sum_{i=kh}^{(k+1)h-1}a(i)b(i)\big(\b_v\rho_0+N^2\b_v\rho^2_0\big(2^{h+2}\sigma^2V(kh)\cr
%  &&\hspace{-0.4cm}+2^{h+2}h\sigma^2+2b^2\big)\big)\cr
&&\hspace{-1.4cm}\leq 4 \sum_{i=kh}^{(k+1)h-1}a(i)b(i)\E\big[\|\H(i)\|^2\|v(i)\|^2+\|W(i)\|^2\|M(i)\|^2\|\xi(i)\|^2|\F(kh-1)\big]\cr
  &&\hspace{-1.4cm}\leq 4h\big(\b_v\rho_0+N^2\b_v\rho^2_0\big(2^{h+2}\sigma^2V(kh)+2^{h+2}h\sigma^2+2b^2\big)\big)a(kh)b(kh)~\text{a.s.},~k\ge \max\{k_1,k_3\}.
  \ena
Similar to (\ref{Anxccdd}), we get
  \ban
  \E\big[v^T(i)\H(i)\Phi_P^T((k+1)h-1,i+1)\Phi_P((k+1)h-1,j+1)\H^T(j)v(j)|\F(kh-1)\big]
  =0, \cr kh\leq j\neq i \leq (k+1)h-1,
  \ean
which together with Assumption \textbf{(A2)}, Lemma \ref{Alemma3} and Lemma A.1 in \cite{Alw} gives
  \bna\label{Afjjfjfkdlsjffj}
\E\big[A^T_2(k)A_2(k)|\F(kh-1)\big]
%  &&\hspace{-0.7cm}=\sum_{i=kh}^{(k+1)h-1}a^2(i)\E\big[v^T(i)\H(i)\Phi^T_P((k+1)h-1,i+1)\cr
%  &&\hspace{-0.4cm}\times\Phi_P((k+1)h-1,i+1)\H^T(i)v(i)|\F(kh-1)\big]\cr
%  &&\hspace{-0.7cm}\leq \sum_{i=kh}^{(k+1)h-1}a^2(i)\big\|\E\big[\E\big[v^T(i)\H(i)\Phi^T_P((k+1)h-1,i+1)\cr
%  &&\hspace{-0.4cm}\times\Phi_P((k+1)h-1,i+1)\H^T(i)v(i)|\F(i)\big]|\F(kh-1)\big]\big\|\cr
%  &&\hspace{-0.7cm}= \sum_{i=kh}^{(k+1)h-1}a^2(i)\big\|\E\big[v^T(i)\H(i)\E\big[\Phi^T_P((k+1)h-1,i+1)\cr
%  &&\hspace{-0.4cm}\times\Phi_P((k+1)h-1,i+1)|\F(i)\big]\H^T(i)v(i)|\F(kh-1)\big]\big\|\cr
%  &&\hspace{-0.7cm}\leq \sum_{i=kh}^{(k+1)h-1}a^2(i)\E\Big[\big\|\E\big[\Phi^T_P((k+1)h-1,i+1)\cr
%  &&\hspace{-0.4cm}\times\Phi_P((k+1)h-1,i+1)|\F(i)\big]\big\|\big\|\H^T(i)\big\|^2\cr
%  &&\hspace{-0.4cm}\times\|v(i)\|^2|\F(kh-1)\Big]\cr
%  &&\hspace{-0.7cm}=2\sum_{i=kh}^{(k+1)h-1}a^2(i)\E\big[\big\|\H^T(i)\H(i)\big\||\F(kh-1)\big]\cr
%  &&\hspace{-0.4cm}\times\E\big[\|v(i)\|^2|\F(kh-1)\big]\cr
%  &&\hspace{-0.7cm}\leq 2\b_v\rho_0\sum_{i=kh}^{(k+1)h-1}a^2(i)\cr
\leq 2h\b_v\rho_0a^2(kh)~\text{a.s.},
% \E\big[A^T_2(k)A_2(k)|\F(kh-1)\big]\leq 2h\b_v\rho_0a^2(kh)~\text{a.s.},
  \ena
for $k\ge k_3$. Following the same way as (\ref{Anxccdd}),
  which along with Assumption \textbf{(A2)}, Lemmas \ref{Alemma2}-\ref{Alemma3} and Lemma A.1 in \cite{Alw} leads to
  \bna\label{Afffffffddsaf}
  &&\hspace{-0.35cm}\E\big[A^T_3(k)A_3(k)|\F(kh-1)\big]\cr
%  &&\hspace{-0.7cm}=\sum_{i=kh}^{(k+1)h-1}b^2(i)\E\big[\xi^T(i)M^T(i) W^T(i)\Phi^T_P((k+1)h-1,\cr
%  &&\hspace{-0.4cm}i+1)\Phi_P((k+1)h-1,i+1)W(i)M(i)\xi(i)|\F(kh-1)\big]\cr
%  &&\hspace{-0.7cm}\leq \sum_{i=kh}^{(k+1)h-1}b^2(i)\big\|\E\big[\E\big[\xi^T(i)M^T(i)W^T(i)\cr
%  &&\hspace{-0.4cm}\times\Phi^T_P((k+1)h-1,i+1)\Phi_P((k+1)h-1,i+1)\cr
%  &&\hspace{-0.4cm}\times W(i)M(i)\xi(i)|\F(i)\big]|\F(kh-1)\big]\big\|\cr
%  &&\hspace{-0.7cm}=\sum_{i=kh}^{(k+1)h-1}b^2(i)\big\|\E\big[\xi^T(i)M^T(i)W^T(i)\cr
%  &&\hspace{-0.4cm}\times\E\big[\Phi^T_P((k+1)h-1,i+1)\Phi_P((k+1)h-1,i+1)\cr
%  &&\hspace{-0.4cm}|\F(i)\big]W(i)M(i)\xi(i)|\F(kh-1)\big]\big\|\cr
%  &&\hspace{-0.7cm}\leq \sum_{i=kh}^{(k+1)h-1}b^2(i)\E\big[\big\|\E[\Phi^T_P((k+1)h-1,i+1)\cr
%  &&\hspace{-0.4cm}\times\Phi_P((k+1)h-1,i+1)|\F(i)\big]\big\|\|W(i)\|^2\|M(i)\|^2\cr
%  &&\hspace{-0.4cm}\times\|\xi(i)\|^2|\F(kh-1)\big]\cr
%  &&\hspace{-0.7cm}=2 \sum_{i=kh}^{(k+1)h-1}b^2(i)\E\big[\E\big[\|W(i)\|^2|\F(i-1)\big]\|M(i)\|^2\cr
%  &&\hspace{-0.4cm}\times\E\big[\|\xi(i)\|^2|\F(i-1)\big]|\F(kh-1)\big]\cr
%  &&\hspace{-0.7cm}\leq 2N^2\b_v\rho^2_0\sum_{i=kh}^{(k+1)h-1}b^2(i)\big(4\sigma^2\E[V(i)|\F(kh-1)]+2b^2\big)\cr
  &&\hspace{-0.7cm}\leq 2hN^2\b_v\rho^2_0\big(2^{h+2}\sigma^2V(kh)+2^{h+2}h\sigma^2+2b^2\big)b^2(kh)~\text{a.s.},
  \ena
for $k\ge \max\{k_1,k_3\}$. By Lemma \ref{Alemma3}, we know that
  \bna\label{Afwwijfiwjf}
  &&\hspace{-0.35cm}\E\big[A^T_4(k)A_4(k)|\F(kh-1)\big]\leq 2Nh^2\|x_0\|^2\lambda^2(kh)~\text{a.s.},
  \ena
    for $k\ge k_3$.
By Lemma \ref{Alemma3}, we have
  \bna\label{Afjcnvx}
\E\big[A^T_1(k)A_1(k)|\F(kh-1)\big]\leq \Bigg(1-2\Lambda_k^h-2\sum_{i=kh}^{(k+1)h-1}\lambda(i)+p(k)\Bigg)V(kh)~\text{a.s.},
  \ena
    for $k\ge k_2.$
Noting that $\lambda(k)$ converges to zero, we know that there exists a positive integer $k_4$, such that $\lambda(k)\leq 1$, $k\ge k_4$. Substituting (\ref{Ajkddsss})-(\ref{Afffwwsffsefefe}) and (\ref{A23})-(\ref{Afjcnvx}) into (\ref{Aerror2}) leads to
  \bann
  &&\hspace{-0.35cm}\E[V((k+1)h)|\F(kh-1)]\cr
%  &&\hspace{-0.7cm}\leq \left(1-2\Lambda_k^h-2\sum_{i=kh}^{(k+1)h-1}\lambda(i)+\sum_{i=kh}^{(k+1)h-1}\lambda(i)+\Omega(k)\right)\cr
%  &&~~\times V(kh)+\Gamma(k)\cr
 &&\hspace{-0.7cm}\leq \left(1+\Omega(k)\right)V(kh)-\left(2\Lambda_k^h+2\sum_{i=kh}^{(k+1)h-1}\lambda(i)\right)V(kh)+\Gamma(k)~\text{a.s.},~k\ge k_0,
  \eann
where \ban
\begin{cases}
k_0=\max\{k_1,k_2,k_3,k_4\},\\
\Omega(k)=p(k)+2^{h+3}hN^2 \b_v\sigma^2\rho^2_0b^2(kh)+2^{h+5}hN^2\b_v\sigma^2\\
~~~~~~~~\times\rho^2_0a(kh)b(kh)+hp(k)+\displaystyle\sum_{i=kh}^{(k+1)h-1}\lambda(i),\\
\Gamma(k)=2h\rho_0\b_va^2(kh)+(2^{h+3}h\sigma^2+4b^2)hN^2\b_v\rho^2_0 b^2(kh)\\~~~~~~~~+8ha(kh)b(kh)(\b_v\rho_0 +N^2\b_v\rho^2_0(2^{h+2}h\sigma^2+2b^2))\\~~~~~~~~+2hN\|x_0\|^2\lambda(kh)+2Nh^2\|x_0\|^2\lambda^2(kh).
\end{cases}
\ean
  Noting that
  $p(k)=\mathcal O(a^2(kh)+b^2(kh)+\lambda^2(kh))$, we get $\Omega(k)+\Gamma(k)=\mathcal O(a^2(kh)+b^2(kh)+\lambda(kh))$.

\vskip 0.2cm

%\section{Proof of Lemma \ref{Alemma5}}\label{Aprooflemma5}
\noindent
%\subsection{Proof of Lemma \ref{Alemma5}}
\textbf{Proof of Lemma \ref{Alemma5}.}
It follows from $\L_{\G(k)}=\mathcal D_{\G(k)}-\A_{\G(k)}$ that $\textbf{1}_N$ is the eigenvector corresponding to the zero eigenvalue of $\L_{\G(k)}$.  $\E[\widehat\L_{\G(k)}|\F(k-1)]$ has a unique zero eigenvalue with eigenvector $\textbf{1}_N$ since $\G(k)$ is conditionally balanced. Therefore,  $\E[\sum_{i=kh}^{(k+1)h-1}\widehat\L_{\G(i)}\otimes I_n|\F(kh-1)]$ has only $n$ zero eigenvalues, whose corresponding eigenvectors consist of $
s_1(k)=\frac{1}{\sqrt{N}}\textbf{1}_N\otimes e_1,\cdots,s_n(k)=\frac{1}{\sqrt{N}}\textbf{1}_N\otimes e_n,
$
where $e_i$ is the standard orthonormal basis in $\mathbb R^n$. Suppose that the remaining eigenvalues of $\E[\sum_{i=kh}^{(k+1)h-1}\widehat\L_{\G(i)}\otimes I_n|\F(kh-1)]$ are $\gamma_{n+1}(k),\cdots,\gamma_{Nn}(k)$, and the corresponding unit orthogonal eigenvectors are $s_{n+1}(k),\cdots,s_{Nn}(k)$. Then, given a positive integer $k$, for any unit vector $\eta\in \mathbb R^{Nn}$, there exist $r_i(k)\in \mathbb R,1\leq i \leq Nn$, such that
$$
\eta= \eta_1(k)+\eta_2(k),
$$
where
\ban
\begin{cases}
\eta_1(k)=\displaystyle\sum_{i=1}^nr_i(k)s_i(k),\\ \eta_2(k)=\displaystyle\sum_{i=n+1}^{Nn}r_i(k)s_i(k),\\
\displaystyle\sum_{i=1}^{Nn}r^2_i(k)=1.
\end{cases}
\ean
Denote  $H_k=\textbf{diag}(H_{k,1},\cdots,H_{k,N}),$ where
\ban
\begin{cases}
H_{k,i}=\displaystyle\sum_{j=kh}^{(k+1)h-1}\E[H_i^T(j)H_i(j)|\F(kh-1)],\\  \L_k=\displaystyle\sum_{i=kh}^{(k+1)h-1}\E[\widehat\L_{\G(i)}|\F(kh-1)],\\  u_{st}(k)=\eta^T_s(k)H_k\eta_t(k),\\
\widetilde w_{st}(k)=\eta^T_s(k)(\L_k\otimes I_n)\eta_t(k),~s,t=1,2.
\end{cases}
\ean
It follows that
\bna\label{Anbkdkbkm}
&&\hspace{-0.4cm}\eta^T\Bigg(\sum_{i=kh}^{(k+1)h-1}\mathbb E\left[\widehat {\mathcal L}_{\mathcal G(i)}\otimes I_n+{\mathcal H}^T(i){\mathcal H}(i)|\F(kh-1)\right]\Bigg)\eta\cr
&&\hspace{-0.7cm}=\sum_{i=1}^2\left(u_{ii}(k)+\widetilde w_{ii}(k)\right)+2u_{12}(k)+2\widetilde w_{12}(k).
\ena
Denote
$$
\z(k)=\frac{2h\rho_0}{2h\rho_0+\lambda_2(\L_k)}.
$$
Noting that $H_k$ is positive semi-definite, by Cauchy inequality, we have
\bna\label{Afjvnnkdv}
&&\hspace{-1.4cm}2\left|\eta_1^T(k)H_k\eta_2(k)\right|=2\left|\eta_1^T(k)H^{\frac{1}{2}}_kH^{\frac{1}{2}}_k\eta_2(k)\right|\leq \z(k) \eta_1^T(k)H_k\eta_1(k)+\frac{1}{\z(k)}\eta_2^T(k)H_k\eta_2(k).
\ena
Substituting (\ref{Afjvnnkdv}) into (\ref{Anbkdkbkm}) leads to
  \bna\label{Awowowwlfff}
  &&\hspace{-0.35cm}\eta^T\Bigg(\sum_{i=kh}^{(k+1)h-1}\mathbb E\left[\widehat {\mathcal L}_{\mathcal G(i)}\otimes I_n+{\mathcal H}^T(i){\mathcal H}(i)|\F(kh-1)\right]\Bigg)\eta\cr
  &&\hspace{-0.7cm}\ge (1-\z(k))u_{11}(k)+\left(1-\frac{1}{\z(k)}\right)u_{22}(k)+\widetilde w_{11}(k)+\widetilde w_{22}(k)+2\widetilde w_{12}(k).
  \ena
  We now consider the RHS of  (\ref{Awowowwlfff}) item by item. Denote $A_k=[r_1(k),\cdots,r_n(k)]^T$ and $B_k=[s_1(k),\cdots,s_n(k)]$, we have
  \bna\label{Afhslweff}
  &&\hspace{-1.4cm}u_{11}(k)=\eta_1^T(k)H_k\eta_1(k)=A^T_kB^T_kH_kB_kA_k.
  \ena
  Noting that  $H_k=\textbf{diag}(H_{k,1},\cdots,H_{k,N})$ and  $s_i(k)=\frac{1}{\sqrt{N}}\textbf{1}_N$ $\otimes e_i$, $1\leq i\leq n$, %we have
%\ban
%  &&\hspace{-0.4cm}H_k^{\frac{1}{2}}B_k=\frac{1}{\sqrt{N}}
%  \begin{pmatrix}
%  H_{k,1}^{\frac{1}{2}}\\
%  H_{k,2}^{\frac{1}{2}}\\
%  \vdots \\
%  H_{k,N}^{\frac{1}{2}}
%  \end{pmatrix},
%\ean
%it follows from  $B_k^TH_kB_k=(H_k^{\frac{1}{2}}B_k)^T(H_k^{\frac{1}{2}}B_k)$
it follows that
$$
B_k^TH_kB_k=\frac{1}{N}\sum_{i=1}^N\sum_{j=kh}^{(k+1)h-1}\E\left[H_i^T(j)H_i(j)|\F(kh-1)\right].
$$
By (\ref{Afhslweff}), we get
  \bna\label{Afjwejfoowqqw}
u_{11}(k)\ge \frac{1}{N}\lambda_{\min}\Bigg(\sum_{i=1}^N\sum_{j=kh}^{(k+1)h-1}\E\left[H_i^T(j)H_i(j)|\F(kh-1)\right]\Bigg)\Bigg(\sum_{i=1}^nr_i^2(k)\Bigg).
  \ena
  It can be verified that
  \bna\label{Ajowofjlf}
  u_{22}(k)\leq \|H_k\|\|\eta_2(k)\|^2\leq h\rho_0 \sum_{i=n+1}^{Nn}r_i^2(k)~\text{a.s.}
  \ena
  Noting that $1-\frac{1}{\z(k)}<0$, by (\ref{Ajowofjlf}), we have
\bna
\left(1-\frac{1}{\z(k)}\right)u_{22}(k)\ge h\rho_0\left(1-\frac{1}{\z(k)}\right)\sum_{i=n+1}^{n}r_i^2(k).
\ena
It follows from $$\left(\E\left[\widehat{\L}_{\G(i)}|\F(m_ih-1)\right]\otimes I_n\right)(\textbf{1}_N\otimes e_j)=\left(\E\left[\widehat{\L}_{\G(i)}|\F(m_ih-1)\right]\textbf{1}_N\right)\otimes e_j=0$$ that
  \bna
  \widetilde w_{11}(k)=\widetilde w_{12}(k)=0.
  \ena
  Since $\G(k|k-1)$ is balanced, then $$\lambda_{n+1}\left(\L_k\otimes I_n\right)=\lambda_2\left(\L_k\right)>0,$$ where $\lambda_{n+1}(\cdot)$ denotes the $(n+1)$-th smallest eigenvalue. Noting that $\{s_{i}(k),n+1\leq i\leq Nn\}$ is an orthonormal system,  we know that
  \bna\label{Ajfkwjfowjfw}
  &&\widetilde w_{22}(k)\ge \lambda_2(\L_k)\Bigg(1-\sum_{i=1}^{n}r^2_i(k)\Bigg)~\rm{a.s.}
  \ena
Denote
\ban
&&~~~F_k(x)\cr
&&=\Bigg(\frac{1-\z(k)}{N}\lambda_{\min}\Bigg(\sum_{i=1}^N\sum_{j=kh}^{(k+1)h-1}\E\big[H_i^T(j)H_i(j)|\F(kh-1)\big]\Bigg)-\lambda_2(\L_k)-h\rho_0+\frac{h\rho_0}{\z(k)}\Bigg)x\cr
&&~~~+\lambda_2(\L_k)+h\rho_0-\frac{h\rho_0}{\z(k)},~ x\in \mathbb R.
\ean
Then, by (\ref{Awowowwlfff}) and (\ref{Afjwejfoowqqw})-(\ref{Ajfkwjfowjfw}), we have
  \bna\label{Aoowwffcww}
\eta^T\Bigg(\sum_{i=kh}^{(k+1)h-1}\mathbb E\big[\widehat {\mathcal L}_{\mathcal G(i)}\otimes I_n+{\mathcal H}^T(i){\mathcal H}(i)|\F(kh-1)\big]\Bigg)\eta \ge F_k\Bigg(\sum_{i=1}^nr_i^2(k)\Bigg)~~\text{a.s.},
  \ena
which along with $\z(k)=\frac{2h\rho_0}{2h\rho_0+\lambda_2(\L_k)}$ gives $$\frac{dF_k(x)}{dx}\leq 0~ \text{a.s.}$$
%  \ban
%  &&\hspace{-0.35cm}\frac{dF_k(x)}{dx}\cr
%  &&\hspace{-0.7cm}=\lambda_{\min}\Bigg(\sum_{i=1}^N\sum_{j=kh}^{(k+1)h-1}\E\big[H_i^T(j)H_i(j)|\F(kh-1)\big]\Bigg)\cr
%  &&\hspace{-0.35cm}\times \frac{1-\z(k)}{N}-\lambda_2(\L_k)-h\rho_0+\frac{h\rho_0}{\z(k)} \cr
%  &&\hspace{-0.7cm}=\lambda_{\min}\Bigg(\sum_{i=1}^N\sum_{j=kh}^{(k+1)h-1}\E\big[H_i^T(j)H_i(j)|\F(kh-1)\big]\Bigg)\cr
%  &&\hspace{-0.35cm}\times \frac{\lambda_2(\L_k)}{N(2h\rho_0+\lambda_2(\L_k))}-\frac{1}{2}\lambda_2(\L_k)\cr
%  &&\hspace{-0.7cm}\leq \Bigg\|\sum_{i=1}^N\sum_{j=kh}^{(k+1)h-1}\E\big[H_i^T(j)H_i(j)|\F(kh-1)\big]\Bigg\|\cr
%  &&\hspace{-0.35cm}\times \frac{\lambda_2(\L_k)}{N(2h\rho_0+\lambda_2(\L_k))}-\frac{1}{2}\lambda_2(\L_k)\cr
%  &&\hspace{-0.7cm}\leq \sum_{i=1}^N\sum_{j=kh}^{(k+1)h-1}\E\big[\big\|H_i^T(j)H_i(j)\big\||\F(kh-1)\big]\cr
%  &&\hspace{-0.35cm}\times \frac{\lambda_2(\L_k)}{N(2h\rho_0+\lambda_2(\L_k))}-\frac{1}{2}\lambda_2(\L_k)\cr
%  %&&\hspace{-0.7cm}\leq \frac{\lambda_2(\L_k)}{N(2h\rho_0+\lambda_2(\L_k))}\sum_{i=1}^N\sum_{j=kh}^{(k+1)h-1}\E[\|\H^T(j)\H(j)\|\cr
%  %&&\hspace{-0.4cm}|\F(kh-1)]-\frac{1}{2}\lambda_2(\L_k)\cr
%  &&\hspace{-0.7cm}\leq \frac{Nh\rho_0\lambda_2(\L_k)}{N(2h\rho_0+\lambda_2(\L_k))}-\frac{1}{2}\lambda_2(\L_k)\cr
%  &&\hspace{-0.7cm}=-\frac{N\lambda^2_2(\L_k)}{2N(2h\rho_0+\lambda_2(\L_k))}\cr
%  &&\hspace{-0.7cm}\leq 0~\text{a.s.}
%  \ean
Hence, the function $F_k(x)$ is monotonically decreasing. It follows from (\ref{Aoowwffcww}) and $$0\leq \sum_{i=1}^nr_i^2(k)\leq 1$$ that
$$\eta^T\left(\sum_{i=kh}^{(k+1)h-1}\mathbb E\left[\widehat {\mathcal L}_{\mathcal G(i)}\otimes I_n+{\mathcal H}^T(i){\mathcal H}(i)|\F(kh-1)\right]\right)\eta \ge F_k(1)
~ \text{a.s.},$$ which completes the proof.
%  \ban
%  &&\hspace{-0.7cm}\widetilde\Lambda_k^h=\min_{\|\eta\|=1}\eta^T\Bigg(\sum_{i=kh}^{(k+1)h-1}\mathbb E\big[\widehat {\mathcal L}_{\mathcal G(i)}\otimes I_n\cr
%  &&~~~~~~~~~~~~~~~~+{\mathcal H}^T(i){\mathcal H}(i)|\F(kh-1)\big]\Bigg)\eta\cr  &&\hspace{-0.25cm}\ge \lambda_{\min}\Bigg(\sum_{i=1}^N\sum_{j=kh}^{(k+1)h-1}\E\big[H_i^T(j)H_i(j)|\F(kh-1)\big]\Bigg)\cr
%  &&~~~~\times \frac{\lambda_2(\L_k)}{2Nh\rho_0+N\lambda_2(\L_k)}~\text{a.s.}
%  \ean

\section{Key Lemmas}
\setcounter{lemma}{0}
\def\thelemma{A.\arabic{lemma}}
\setcounter{equation}{0}
\def\theequation{C.\arabic{equation}}

\begin{lemma}\label{Alemma2}
\rm{For the algorithm (\ref{Aalgorithm}), if Assumptions \textbf{(A1)}-\textbf{(A2)} hold, the algorithm gains $a(k)$, $b(k)$ and $\lambda(k)$ monotonically decrease to zero, and there exists a positive constant $\rho_0$, such that
$\sup_{k\ge 0}(\|\mathcal L_{\mathcal G(k)}\|+(\mathbb E[\|\H^T(k)\H(k)\|^{2}|\mathscr F(k-1)])^{\frac{1}{2}})\leq \rho_0$ a.s.,
then there exist nonnegative deterministic sequences $\{\a(k),k\ge 0\}$  and $\{\gamma(k),k\ge 0\}$ satisfying $\a(k)=o(1)$ and $\gamma(k)=\mathcal O(a^2(k)+b^2(k)+\lambda(k))$, such that
$\E[V(k)|\F(m_kh-1)]\leq \prod_{i=m_kh}^{k-1}(1+\a(i))V(m_kh)+\sum_{i=m_kh}^{k-1}\g(i)\prod_{j=i+1}^{k-1}(1+\a(j))$ a.s., $h\ge 1$,
where $m_k=\lfloor \frac{k}{h}\rfloor$, especially, there exists a positive integer $k_1$, such that
$\E[V(k)|\F(m_kh-1)]\leq 2^hV(m_kh)+h2^h\gamma(m_kh)~\text{a.s.},~k\ge k_1$.  }
\end{lemma}
\vskip 0.2cm
\begin{proof}
By the estimation error equation $(\ref{Aerror})$, we have
\bna\label{Alpojjkjk}
&&\hspace{-0.35cm}V(k+1)\cr
&&\hspace{-0.7cm}=(1-\lambda(k))^2V(k)-(1-\lambda(k))\d^T(k)\left(D(k)+D^T(k)\right)\d(k)+\d^T(k)D^T(k)D(k)\d(k)\cr
&&\hspace{-0.35cm}+S^T(k)S(k)+2S^T(k)\left((1-\lambda(k))I_{Nn}-D(k)\right)\d(k)+\lambda^2(k)\left\|\textbf{1}_N\otimes x_0\right\|^2\cr
&&\hspace{-0.35cm}-2\lambda(k)\d^T(k)\big((1-\lambda(k))I_{Nn}-D^T(k)\big)(\textbf{1}_N\otimes x_0)-2\lambda(k)S^T(k)(\textbf{1}_N\otimes x_0)\cr
&&\hspace{-0.7cm}\leq (1-\lambda(k))^2V(k)+|1-\lambda(k)|\big|\d^T(k)\left(D(k)+D^T(k)\right)\d(k)\big|+\|D(k)\|^2\|\d(k)\|^2+\|S(k)\|^2\cr &&\hspace{-0.35cm}-2\lambda(k)S^T(k)(\textbf{1}_N\otimes x_0)+2S^T(k)((1-\lambda(k))I_{Nn}-D(k))\d(k)+N\lambda^2(k)\|x_0\|^2\cr
&&\hspace{-0.35cm}+2\lambda(k)\left|\d^T(k)\left((1-\lambda(k))I_{Nn}-D^T(k)\right)(\textbf{1}_N\otimes x_0)\right|,
\ena
where $S(k)=a(k)\H^T(k)v(k)+b(k)W(k)M(k)\xi(k)$, $D(k)=b(k)\mathcal{L}_{\mathcal{G}}(k)\otimes I_n+a(k)\mathcal{H}^{\mathrm{T}}(k)\mathcal{H}(k)$. We now consider the conditional mathematical expectation of each term on the RHS of (\ref{Alpojjkjk}) w.r.t. $\F(m_kh-1)$. Noting that $k-1\ge m_kh-1$, by Assumption \textbf{(A2)}, we get
\bna\label{Alemma211111}
&&\hspace{-0.35cm}\E\left[S^T(k)((1-\lambda(k))I_{Nn}-D(k))\d(k)|\F(m_kh-1)\right]\cr
&&\hspace{-0.7cm}=\E\left[a(k)v^T(k)\H(k)((1-\lambda(k))I_{Nn}-D(k))\d(k)|\F(m_kh-1)\right]\cr
&&\hspace{-0.4cm}+\E\left[b(k)\xi^T(k)M^T(k)W^T(k)((1-\lambda(k))I_{Nn}-D(k))\d(k)|\F(m_kh-1)\right]\cr
&&\hspace{-0.7cm}=\E\left[\E\left[a(k)v^T(k)\H(k)((1-\lambda(k))I_{Nn}-D(k))\d(k)|\F(k-1)\right]|\F(m_kh-1)\right]\cr &&\hspace{-0.4cm}+\E\left[\E\left[b(k)\xi^T(k)M^T(k)W^T(k)((1-\lambda(k))I_{Nn}-D(k))\d(k)|\F(k-1)\right]|\F(m_kh-1)\right]\cr
&&\hspace{-0.7cm}=a(k)\E\left[\E\left[v^T(k)|\F(k-1)\right]\E[\H(k)((1-\lambda(k))I_{Nn}-D(k))|\F(k-1)]\d(k)|\F(m_kh-1)\right]\cr
&&\hspace{-0.4cm}+b(k)\E\big[\E\left[\xi^T(k)|\F(k-1)\right]M^T(k)\cr
&&\hspace{-0.4cm}\times\E\left[W^T(k)((1-\lambda(k))I_{Nn}-D(k))|\F(k-1)\right]\d(k)|\F(m_kh-1)\big]\cr
&&\hspace{-0.7cm}=0,
\ena
where the penultimate equality is due to $\d(k)\in \F(k-1)$, $M(k)\in \F(k-1)$ and Lemma A.1 in \cite{Alw}. Following the same way as above, we have
\bna\label{Afwwlmnc}
&&\hspace{-0.35cm}\E\left[S^T(k)(\textbf{1}_N\otimes x_0)|\F(m_kh-1)\right]\cr
&&\hspace{-0.7cm}=\E\left[a(k)v^T(k)\H(k)+b(k)\xi^T(k)M^T(k)W^T(k)|\F(m_kh-1)\right](\textbf{1}_N\otimes x_0)\cr
&&\hspace{-0.7cm}=a(k)\E\left[\E\left[v^T(k)|\F(k-1)\right]\E[\H(k)|\F(k-1)]|\F(m_kh-1)\right](\textbf{1}_N\otimes x_0)\cr
&&\hspace{-0.4cm}+b(k)\E\left[\E\left[\xi^T(k)|\F(k-1)\right]M^T(k)\E\left[W^T(k)|\F(k-1)\right]|\F(m_kh-1)\right](\textbf{1}_N\otimes x_0)\cr
&&\hspace{-0.7cm}=0.
\ena
Denote $q(k)=\max\{a(k),b(k)\}$. Noting that $V(k)\in \F(k-1)$, we obtain
\bna\label{Alemma211112}
&&\hspace{-0.35cm}\E\left[\|D(k)\|^2\|\d(k)\|^2|\F(m_kh-1)\right]\cr
&&\hspace{-0.7cm}\leq \E\left[\left(b(k)\left\|\L_{\G(k)}\right\|+a(k)\left\|\H^T(k)\H(k)\right\|\right)^2\|\d(k)\|^2|\F(m_kh-1)\right]\cr
&&\hspace{-0.7cm}\leq q^2(k)\E\left[\E\left[\left(\left\|\L_{\G(k)}\right\|+\left\|\H^T(k)\H(k)\right\|\right)^2V(k)|\F(k-1)\right]|\F(m_kh-1)\right]\cr
&&\hspace{-0.7cm}=q^2(k)\E\left[\E\left[\left(\left\|\L_{\G(k)}\right\|+\left\|\H^T(k)\H(k)\right\|\right)^2|\F(k-1)\right]V(k)|\F(m_kh-1)\right]\cr
&&\hspace{-0.7cm}\leq 2q^2(k)\rho^2_0\E[V(k)|\F(m_kh-1)]~\text{a.s.}
\ena
It follows from  the definition of $M(k)$ that
$
\E[\|M(k)\|^2|\F(m_kh-1)]\leq 4\sigma^2\E[V(k)|\F(m_kh-1)]+2b^2
$, which together with Assumption \textbf{(A2)} and Lemma A.1 in \cite{Alw} gives
\bna\label{Alemma211113}
&&\hspace{-0.35cm}\E\left[\|S(k)\|^2|\F(m_kh-1)\right]\cr
&&\hspace{-0.7cm}\leq 2q^2(k)\E\left[\left\|\H^T(k)\right\|^2\|v(k)\|^2+\|W(k)\|^2\|M(k)\|^2\|\xi(k)\|^2|\F(m_kh-1)\right]\cr
&&\hspace{-0.7cm}\leq 2q^2(k)\E\left[\rho_0\b_v+\E\left[\|W(k)\|^2\|M(k)\|^2\|\xi(k)\|^2|\F(k-1)\right]|\F(m_kh-1)\right]\cr
&&\hspace{-0.7cm}= 2q^2(k)\E\left[\rho_0\b_v+\E\left[\|W(k)\|^2\|\xi(k)\|^2|\F(k-1)\right]\|M(k)\|^2|\F(m_kh-1)\right]\cr
&&\hspace{-0.7cm}= 2q^2(k)\E\left[\rho_0\b_v+\E\left[\|W(k)\|^2|\F(k-1)\right]\E\left[\|\xi(k)\|^2|\F(k-1)\right]\|M(k)\|^2|\F(m_kh-1)\right]\cr
&&\hspace{-0.7cm}\leq 2q^2(k)\rho_0\b_v+2q^2(k)N^2\rho^2_0\b_v(4\sigma^2\E[V(k)|\F(m_kh-1)]+2b^2)~\text{a.s.}
\ena
Noting that
\bna
&&\hspace{-0.35cm}\E\left[\left|\d^T(k)(D(k)+D^T(k))\d(k)\right||\F(m_kh-1)\right]\cr
&&\hspace{-0.7cm}\leq 2\E\left[V(k)\left(b(k)\left\|\L_{\G(k)}\right\|+a(k)\left\|\H^T(k)\H(k)\right\|\right)|\F(m_kh-1)\right]\cr
&&\hspace{-0.7cm}=2\E\left[\E\left[V(k)(b(k)\left\|\L_{\G(k)}\right\|+a(k)\left\|\H^T(k)\H(k)\right\|)|\F(k-1)\right]|\F(m_kh-1)\right]\cr
&&\hspace{-0.7cm}\leq 2q(k)\E\left[\E\left[\left\|\L_{\G(k)}\right\|+\left\|\H^T(k)\H(k)\right\||\F(k-1)\right]V(k)|\F(m_kh-1)\right]\cr
&&\hspace{-0.7cm}\leq 2q(k)\rho_0\E[V(k)|\F(m_kh-1)]~\text{a.s.},
\ena
by mean value inequality, we have
\bna\label{Aoeoeoeo}
&&\hspace{-0.35cm}\E\left[\d^T(k)\left((1-\lambda(k))I_{Nn}-D^T(k)\right)(\textbf{1}_N\otimes x_0)|\F(m_kh-1)\right]\cr
&&\hspace{-0.7cm}\leq \E\left[\|\d(k)\|\left\|(1-\lambda(k))I_{Nn}-D^T(k)\right\|\|\textbf{1}_N\otimes x_0\||\F(m_kh-1)\right]\cr
&&\hspace{-0.7cm}\leq \E[\|\d(k)\|\left(|1-\lambda(k)|+\|D(k)\|\right)\|\textbf{1}_N\otimes x_0\||\F(m_kh-1)]\cr
&&\hspace{-0.7cm}\leq \E\left[\|\d(k)\|\left(|1-\lambda(k)|+b(k)\left\|\L_{\G(k)}\right\|+a(k)\left\|\H^T(k)\H(k)\right\|\right)\|\textbf{1}_N\otimes x_0\||\F(m_kh-1)\right]\cr
&&\hspace{-0.7cm}\leq \E\left[\|\d(k)\|\|\textbf{1}_N\otimes x_0\|+\|\d(k)\|\|\textbf{1}_N\otimes x_0\|q(k)\left(\left\|\L_{\G(k)}\right\|+\left\|\H^T(k)\H(k)\right\|\right)|\F(m_kh-1)\right]\cr
&&\hspace{-0.7cm}\leq \frac{1}{2}\left(\E[V(k)|\F(m_kh-1)]+N\|x_0\|^2\right)\cr &&\hspace{-0.5cm}+\frac{1}{2}\left(N\|x_0\|^2\E[V(k)|\F(m_kh-1)]+q^2(k)\rho^2_0\right)~\text{a.s.},
\ena
which together with (\ref{Alpojjkjk})-(\ref{Aoeoeoeo}) leads to
\bna\label{Aaaaaa}
\E[V(k+1)|\F(m_kh-1)]\leq (1+\a(k))\E[V(k)|\F(m_kh-1)]+\gamma(k)~\text{a.s.},
\ena
where   $\a(k)=\lambda^2(k)+\lambda(k)(1+N\|x_0\|^2)+2q^2(k)\rho^2_0+8q^2(k)N^2\rho^2_0\b_v\sigma^2+2\rho_0|1-\lambda(k)|q(k)$ and $ \gamma(k)=4b^2q^2(k)N^2\rho^2_0\b_v+2q^2(k)\rho_0\b_v+(N\|x_0\|^2+q^2(k)\rho^2_0)\lambda(k)+N\|x_0\|^2\lambda^2(k)$. Thus, by (\ref{Aaaaaa}), we get
\bna\label{Afscnzsa}
\E[V(k)|\F(m_kh-1)]\leq \prod_{i=m_kh}^{k-1}(1+\a(i))V(m_kh)+\sum_{i=m_kh}^{k-1}\g(i)\prod_{j=i+1}^{k-1}(1+\a(j))~\text{a.s.}
\ena
Since $a(k)$, $b(k)$ and $\lambda(k)$ converge to $0$, then $\a(k)\to 0$ and  $\gamma(k)\to 0$ as $k\to\infty$, from which we know that there exists a positive integer $k_1$ such that $\a(i)\leq 1$, $i\ge k_1$. Noting that $0\leq k-m_kh<h$, the lemma is proved by (\ref{Afscnzsa}).
\end{proof}
\vskip 0.2cm

%The following lemma proves that the norm of the conditional expectation of $\Phi^T_P((k+1)h-1,kh)\Phi_P((k+1)h-1,kh)$ w.r.t. $\F(kh-1)$ has an upper bound strictly less than 1 in the time interval $[kh,(k+1)h-1]$ by performing binomial expansion and inequality estimation on the homogeneous part of the estimation error equation (\ref{Aerror}), which plays an important role in the analysis of the convergence of the algorithm.
%\vskip 0.2cm
%
\begin{lemma}\label{Alemma3}
\rm{For the algorithm (\ref{Aalgorithm}), if the algorithm gains $a(k)$, $b(k)$ and $\lambda(k)$ monotonically decrease satisfying $\sum_{k=0}^{\infty}($ $a^2(k)+b^2(k)+\lambda^2(k))<\infty$, and there exists a positive integer $h$ and a positive constant $\rho_0$, such that
$\sup_{k\ge 0}(\|\mathcal L_{\mathcal G(k)}\|+(\mathbb E[\|{\mathcal H}^T(k){\mathcal H}(k)\|^{2^{\max\{h,2\}}}|\mathscr F(k-1)])^{\frac{1}{2^{\max\{h,2\}}}})\leq \rho_0~\text{a.s.},$
then there exits a positive integer $k_2$, such that
$\|\mathbb E[\Phi_P^T((k+1)h-1,kh)\Phi_P((k+1)h-1,kh)|\F(kh-1)]\|\leq 1-2\Lambda_k^h-2\sum_{i=kh}^{(k+1)h-1}\lambda(i)+p(k)~\text{a.s.},~k\ge k_2$,
where $p(k)=(9^h-1-4h)(\rho_0\max\{a(kh),b(kh)\}+\lambda(kh))^2$, especially, there exists a positive integer $k_3$, such that
$\|\mathbb E[\Phi_P^T((k+1)h-1,i+1)\Phi_P((k+1)h-1,i+1)|\F(kh-1)]\|\leq 2~\text{a.s.}$, $k\ge k_3$, $\forall~ i\in [kh,(k+1)h-1].$
}
\end{lemma}
\vskip 0.2cm

\begin{proof}
By the definitions of $D(k)$ and $P(k)$, it follows that
\bna\label{A0302}
&&\hspace{-0.35cm}\big\|\mathbb E\big[\Phi^T_P((k+1)h-1,kh)\Phi_P((k+1)h-1,kh)|\mathscr F(kh-1)\big]\big\|\cr
&&\hspace{-0.7cm}=\big\|\mathbb E\big[\big((1-\lambda(kh))I_{Nn}-D^T(kh)\big)\times\cdots\times\big((1-\lambda((k+1)h-1))I_{Nn}-D^T((k+1)h-1)\big)\cr
&&\hspace{-0.35cm}\times((1-\lambda((k+1)h-1))I_{Nn}-D((k+1)h-1))\times\cdots\cr
&&\hspace{-0.35cm}\times((1-\lambda(kh))I_{Nn}-D(kh))|\mathscr F(kh-1)\big]\big\|\cr
&&\hspace{-0.7cm}=\left\|I_{Nn}-\sum_{i=kh}^{(k+1)h-1}\mathbb E\left[D^T(i)+D(i)+2\lambda(i)I_{Nn}|\mathscr F(kh-1)\right]+\mathbb E\left[\sum_{i=2}^{2h}M_i(k)|\mathscr F(kh-1)\right]\right\|\cr
&&\hspace{-0.7cm}\le\left\|I_{Nn}-\sum_{i=kh}^{(k+1)h-1}\mathbb E\left[D^T(i)+D(i)+2\lambda(i)I_{Nn}|\mathscr F(kh-1)\right]\right\|+\left\|\mathbb E\left[\sum_{i=2}^{2h}M_i(k)|\mathscr F(kh-1)\right]\right\|,\cr
&&\,
\ena
where $M_i(k),i=2,\cdots,2h$ represent the $i$-th order terms in the binomial expansion of $\Phi^T_P((k+1)h-1,kh)\Phi_P((k+1)h-1,kh)$. By the definition of spectral radius, we have
\bna\label{A0303}
&&\hspace{-0.35cm}\max_{1\leq i \leq Nn}\lambda_i\left(\sum_{j=kh}^{(k+1)h-1}\mathbb E\left[D^T(j)+D(j)+2\lambda(j)I_{Nn}|\mathscr F(kh-1)\right]\right)\cr
&&\hspace{-0.7cm}\leq \left\|\sum_{j=kh}^{(k+1)h-1}\mathbb E\left[D^T(j)+D(j)+2\lambda(j)I_{Nn}|\mathscr F(kh-1)\right]\right\|\cr
&&\hspace{-0.7cm}\leq 2\sum_{j=kh}^{(k+1)h-1}\mathbb E\left[b(j)\left\|\L_{\G(j)}\right\|+a(j)\left\|\H^T(j)\H(j)\right\|+\lambda(j)|\F(kh-1)\right]\cr
&&\hspace{-0.7cm}\leq 2\sum_{j=kh}^{(k+1)h-1}\max\{a(j),b(j)\}\rho_0+2\sum_{j=kh}^{(k+1)h-1}\lambda(j)\cr
&&\hspace{-0.7cm}\leq 2\sum_{j=kh}^{(k+1)h-1}(\rho_0a(j)+\rho_0b(j)+\lambda(j))~\text{a.s.}
\ena
Noting that algorithm gains decrease to zero, we know that the RHS of the last inequality of (\ref{A0303}) converges to zero as $k\to\infty$, which is independent of the sample paths. Hence, there exists a positive integer $l_1$, such that
\ban
&&\hspace{-0.7cm}\lambda_i\Bigg(\sum_{j=kh}^{(k+1)h-1}\mathbb E\big[D^T(j)+D(j)+2\lambda(j)|\mathscr F(kh-1)\big]\Bigg)\leq 1,~i=1,\cdots,Nn,~k\ge l_1,~\text{a.s.},
\ean
from which we get
\bna\label{A0305}
&&\hspace{-0.35cm}\left\|I_{Nn}-\sum_{j=kh}^{(k+1)h-1}\mathbb E\left[D^T(j)+D(j)+2\lambda(j)I_{Nn}|\mathscr F(kh-1)\right]\right\|\cr
&&\hspace{-0.7cm}=\rho\left(I_{Nn}-\sum_{j=kh}^{(k+1)h-1}\mathbb E\left[D^T(j)+D(j)+2\lambda(j)I_{Nn}|\mathscr F(kh-1)\right]\right)\cr
&&\hspace{-0.7cm}=\max_{1\leq i \leq Nn}\left|\lambda_i\left(I_{Nn}-\sum_{j=kh}^{(k+1)h-1}\mathbb E\left[D^T(j)+D(j)+2\lambda(j)I_{Nn}|\mathscr F(kh-1)\right]\right)\right|\cr
&&\hspace{-0.7cm}=1-\lambda_{\min}\left(\sum_{j=kh}^{(k+1)h-1}\mathbb E\left[D^T(j)+D(j)+2\lambda(j)I_{Nn}|\mathscr F(kh-1)\right]\right)\cr
&&\hspace{-0.7cm}=1-2\lambda_{\min}\left(\sum_{j=kh}^{(k+1)h-1}\mathbb E\left[b(j)\widehat{\L}_{\G(j)}\otimes I_n+a(j)\H^T(j)\H(j)+\lambda(j)I_{Nn}|\F(kh-1)\right]\right)\cr
&&\hspace{-0.7cm}=1-2\Lambda_k^h-2\sum_{j=kh}^{(k+1)h-1}\lambda(j),~k\ge l_1.
\ena
By Cr-inequality, we have
\ban
&&\hspace{-0.35cm}\left[\E\left[\|D(j)+\lambda(j)I_{Nn}\|^{2^r}|\F(kh-1)\right]\right]^{\frac{1}{2^r}}\cr
&&\hspace{-0.7cm}\leq \left[\E\left[\left(\left\|b(j)\L_{\G(j)}\otimes I_{Nn}+a(j)\H^T(j)\H(j)\right\|+\lambda(j)\right)^{2^r}|\F(kh-1)\right]\right]^{\frac{1}{2^r}}\cr
&&\hspace{-0.7cm}\leq \left[\E\left[2^{2^{r}-1}\left(\left\|b(j)\L_{\G(j)}\otimes I_{Nn}+a(j)\H^T(j)\H(j)\right\|^{2^r}+\lambda^{2^r}(j)\right)|\F(kh-1)\right]\right]^{\frac{1}{2^r}}\cr
&&\hspace{-0.7cm}\leq 2\left(\rho_0^{2^r}q^{2^r}(j)+\lambda^{2^r}(j)\right)^{\frac{1}{2^r}}\cr
&&\hspace{-0.7cm}\leq 2(\rho_0q(j)+\lambda(j)),~1\leq r\leq h,~kh\leq j \leq (k+1)h-1,
\ean
which together with conditional H\"{o}lder inequality and conditional Lyapunov inequality leads to
\bna\label{A0307}
&&\hspace{-0.35cm}\E\left[\left\|\prod_{j=1}^r(D(n_j)+\lambda(n_j)I_{Nn})\right\||\mathscr F(kh-1)\right]\cr
&&\hspace{-0.7cm}\leq \left[\E\left[\left\|\prod_{j=1}^{r-1}(D(n_j)+\lambda(n_j)I_{Nn})\right\|^2|\mathscr F(kh-1)\right]\right]^{\frac{1}{2}}\left[\E\left[\|D(n_r)+\lambda(n_r)I_{Nn}\|^2|\mathscr F(kh-1)\right]\right]^{\frac{1}{2}}\cr
&&\hspace{-0.7cm}\leq
2(\rho_0q(n_r)+\lambda(n_r))\left[\E\left[\left\|\prod_{j=1}^{r-1}(D(n_j)+\lambda(n_j)I_{Nn})\right\|^2|\mathscr F(kh-1)\right]\right]^{\frac{1}{2}}\cr
&&\hspace{-0.7cm}\leq 2^r\prod_{j=1}^{r}(\rho_0q(n_j)+\lambda\left(n_j\right))~\text{a.s.},~1\leq r \leq h,
\ena
where $kh\leq n_1\leq \cdots \leq n_r \leq (k+1)h-1$. On one hand, we have
\bna\label{Auiojk}
&&\hspace{-0.35cm}\E[\|M_i(k)\||\mathscr F(kh-1)]\cr
&&\hspace{-0.7cm}=\E\left[\left\|\sum_{s+t=i}\prod_{l=1}^s\left(D^T(n_l)+\lambda(n_l)I_{Nn}\right)\prod_{w=1}^t(D(v_{t+1-w})+\lambda(v_{t+1-w})I_{Nn})\right\||\mathscr F(kh-1)\right]\cr
&&\hspace{-0.7cm}\leq \sum_{s+t=i}\E\left[\left\|\prod_{l=1}^s\left(D^T(n_l)+\lambda(n_l)I_{Nn}\right)\right\|\left\|\prod_{w=1}^t(D(v_{t+1-w})+\lambda(v_{t+1-w})I_{Nn})\right\||\mathscr F(kh-1)\right] \cr
&&\hspace{-0.7cm}\leq \sum_{s+t=i}2^{s+t}\prod_{l=1}^{s}(\rho_0q(n_l)+\lambda(n_l))\prod_{w=1}^{t}(\rho_0q(v_w)+\lambda(v_w))~\text{a.s.},~2\leq i \leq h,
\ena
where $kh\leq n_1\leq \cdots \leq n_s \leq (k+1)h-1$, $kh\leq v_1\leq \cdots \leq v_t \leq (k+1)h-1$. Noting that there exists a positive integer $l_2$, such that $\rho_0q(k)+\lambda(k) \leq 1, k\ge l_2$. By (\ref{Auiojk}), we get
\bna\label{Afjsc}
\mathbb E[\|M_i(k)\||\F(kh-1)]
\leq C_{2h}^i2^i(\rho_0q(kh)+\lambda(kh))^{2}~\text{a.s.},~2\leq i \leq h,~k\ge l_2.
\ena
On the other hand, for the case with $h<i\leq 2h$, each term of $M_i(k)$ is multiplied by at most $i$ ($2\leq i \leq h$) different elements, where the matrix and its transpose are regarded as the same element. Noting that $\|A^TA\|=\|A\|^2$ for any matrix $A$. By (\ref{A0307}), we get
\bna\label{Afjswefwefwefwc}
\mathbb E[\|M_i(k)\||\F(kh-1)]\leq C_{2h}^i2^i(\rho_0q(kh)+\lambda(kh))^{2}~\text{a.s.},~h<i\leq 2h,~k\ge l_2.
\ena
By (\ref{Afjsc}) and (\ref{Afjswefwefwefwc}), we have
\bna\label{Ajkljkjk}
&&\hspace{-0.9cm}\left\|\E\left[\sum_{i=2}^{2h}M_i(k)|\mathscr F(kh-1)\right]\right\|\cr
&&\hspace{-1.2cm}\leq
\sum_{i=2}^{2h}C_{2h}^i2^i(\rho_0q(kh)+\lambda(kh))^{2}\cr
&&\hspace{-1.2cm}=(9^h-1-4h)(\rho_0q(kh)+\lambda(kh))^{2}
~\text{a.s.},~k\ge l_2.
\ena
Denote the RHS of (\ref{Ajkljkjk}) by $p(k)$, we know that  $\sum_{k=0}^{\infty}p(k)$ $<\infty$. Denote $k_2=\max\{l_1,l_2\}$. By (\ref{A0302}), (\ref{A0305}) and (\ref{Ajkljkjk}), we get
\ban
&&~~~\left\|\mathbb E\left[\Phi_P^T((k+1)h-1,kh)\Phi_P((k+1)h-1,kh)|\F(kh-1)\right]\right\|\cr
&&\leq 1-2\Lambda_k^h-2\sum_{i=kh}^{(k+1)h-1}\lambda(i)+p(k)~\text{a.s.},~k\ge k_2.
\ean
Furthermore, denote $G_j(k,i)$ the $j$-th order terms in the binomial expansion of $\Phi_P^T((k+1)h-1,i+1)\Phi_P((k+1)h-1,i+1)$. Similar to (\ref{Afjswefwefwefwc}), we get
$\E[\|G_j(k,i)\||\F(i)]\leq C_{2((k+1)h-i-1)}^j2^j(\rho_0q(kh)$ $+\lambda(kh))^2~\text{a.s.},~k\ge k_2,~kh\leq i \leq (k+1)h-1,~2\leq j \leq 2((k+1)h-i-1),
$
which along with (\ref{A0303}) gives
\ban
&&\hspace{-0.35cm}\left\|\mathbb E\left[\Phi_P^T((k+1)h-1,i+1)\Phi_P((k+1)h-1,i+1)|\F(i)\right]\right\|\cr
&&\hspace{-0.7cm}= \left\|I_{Nn}-\sum_{j=i+1}^{(k+1)h-1}\mathbb E\left[D^T(j)+D(j)+2\lambda(j)I_{Nn}|\mathscr F(i)\right]+\mathbb E\left[\sum_{j=2}^{2((k+1)h-i-1)}G_j(k,i)|\mathscr F(i)\right]\right\|\cr
&&\hspace{-0.7cm}\leq \left\|I_{Nn}-\sum_{j=i+1}^{(k+1)h-1}\mathbb E\left[D^T(j)+D(j)+2\lambda(j)I_{Nn}|\mathscr F(i)\right]\right\|+\left\|\mathbb E\left[\sum_{j=2}^{2((k+1)h-i-1)}G_j(k,i)|\mathscr F(i)\right]\right\|\cr
&&\hspace{-0.7cm}\leq 1+\left\|\sum_{j=i+1}^{(k+1)h-1}\mathbb E\left[D^T(j)+D(j)+2\lambda(j)I_{Nn}|\mathscr F(i)\right]\right\|+\sum_{j=2}^{2((k+1)h-i-1)}\mathbb E\left[\|G_j(k,i)\||\mathscr F(i)\right]\cr
&&\hspace{-0.7cm}\leq 1+\sum_{j=kh}^{(k+1)h-1}\left\|\mathbb E\left[D^T(j)+D(j)+2\lambda(j)I_{Nn}|\mathscr F(i)\right]\right\|+\sum_{j=2}^{2((k+1)h-i-1)}\mathbb E[\|G_j(k,i)\||\mathscr F(i)]\cr
&&\hspace{-0.7cm}\leq 1+2\sum_{j=kh}^{(k+1)h-1}(\rho_0a(j)+\rho_0b(j)+\lambda(j))+\sum_{j=2}^{2((k+1)h-i-1)}C_{2((k+1)h-i-1)}^j2^j(\rho_0q(kh)+\lambda(kh))^2 \cr
&&\hspace{-0.7cm}\leq 1+2\sum_{j=kh}^{(k+1)h-1}(\rho_0a(j)+\rho_0b(j)+\lambda(j))+\sum_{j=2}^{2h}C_{2h}^j2^j(\rho_0q(kh)+\lambda(kh))^2 \cr
&&\hspace{-0.7cm}=1+2\sum_{j=kh}^{(k+1)h-1}(\rho_0a(j)+\rho_0b(j)+\lambda(j))+p(k)~\text{a.s.},~k\ge k_2,
\ean
which further shows that there exists a positive integer $k_3$, such that
$\|\mathbb E[\Phi_P^T((k+1)h-1,i+1)\Phi_P((k+1)h-1,i+1)|\F(i)]\|\leq 2~\text{a.s.},~k\ge k_3.$
\end{proof}
\vskip 0.2cm

\begin{lemma}\label{Alemma6}
\rm{For the algorithm (\ref{Aalgorithm}), suppose that $\{\mathcal G(k),k\ge 0\}\in \Gamma_1$,   Assumptions \textbf{(A1)}-\textbf{(A2)} hold, the algorithm gains $a(k)$, $b(k)$ and $\lambda(k)$ monotonically decrease to zero, and there exist positive constants $\rho_0$, $\theta_1$ and $\theta_2$ and a positive integer $h$, such that
(i) $\inf_{k\ge 0}\lambda_2(\sum_{j=kh}^{(k+1)h-1}\E[\widehat\L_{\G(j)}|$ $\F(kh-1)])\ge \theta_1~\rm{a.s.}$;
(ii) $\inf_{k\ge 0}\lambda_{\min}(\sum_{i=1}^N\sum_{j=kh}^{(k+1)h-1}\mathbb E[H_i^T(j)H_i(j)|\mathscr F(kh-1)])\ge \theta_2~\rm{a.s.}$;
(iii) $\sup_{k\ge 0}$ $(\|\mathcal L_{\mathcal G(k)}\|+(\mathbb E[\|{\mathcal H}^T(k){\mathcal H}(k)$ $\|^{2^{\max\{h,2\}}}|\mathscr F(k-1)])^{\frac{1}{2^{\max\{h,2\}}}})\leq \rho_0~\text{a.s.}$
Denote
\ban
G(k)=\frac{\theta_1\theta_2}{2Nh\rho_0+N\theta_1}\min\{a((k+1)h),b((k+1)h)\}+\sum_{i=kh}^{(k+1)h-1}\lambda(i).
\ean
(I). If  $\sum_{k=0}^{\infty}G(k)=\infty$ and  $\sum_{k=0}^{\infty}(a^2(k)+b^2(k)+\lambda(k))<\infty$,
then $\lim_{k\to\infty}x_i(k)=x_0,~i\in \mathcal V$ a.s.\\
(II). If $ \sum_{k=0}^{\infty}G(k)=\infty$ and $a^2(kh)+b^2(kh)+\lambda(kh)=o(G(k))$,
then $\lim_{k\to\infty}\E[\|x_i(k)-x_0\|^2]=0,~i\in \mathcal V$.
  }
\end{lemma}
\vskip 0.2cm

\begin{proof}
Noting that $\{\mathcal G(k),k\ge 0\}\in \Gamma_1$, it follows that $\E[\widehat \L_{\G(k)}|\F(k-1)]$ is positive semi-definite. Noting that $\E[\widehat \L_{\G(k)}|\F(mh-1)]=\E[\E[\widehat \L_{\G(k)}|\F(k-1)]|\F(mh-1)]$, $k\ge mh$, then $\E[\widehat \L_{\G(k)}|\F(mh-1)]$ is also positive semi-definite, which shows
\bna\label{Anxsala}
&&\hspace{-0.35cm}\sum_{i=kh}^{(k+1)h-1}\mathbb E\left[b(i)\widehat {\mathcal L}_{\mathcal G(i)}\otimes I_n+a(i){\mathcal H}^T(i){\mathcal H}(i)|\F(kh-1)\right]\cr
&&\hspace{-0.7cm}\ge \min\{a((k+1)h),b((k+1)h)\}\sum_{i=kh}^{(k+1)h-1}\mathbb E\left[\widehat {\mathcal L}_{\mathcal G(i)}\otimes I_n+{\mathcal H}^T(i){\mathcal H}(i)|\F(kh-1)\right].
\ena
By Lemma $\ref{Alemma5}$, condition (ii) and (\ref{Anxsala}), we get
\ban
\Lambda_k^h \ge \frac{\theta_1\theta_2}{2Nh\rho_0+N\theta_1}\min\{a((k+1)h),b((k+1)h)\}~\text{a.s.},
\ean
which leads to
\bna\label{Alxsffff}
\Lambda_k^h+\sum_{i=kh}^{(k+1)h-1}\lambda(i)\ge G(k).
\ena
We first prove (I) of Lemma \ref{Alemma6}. Here, the algorithm gains guarantee that the sample path spatio-temporal persistence of excitation condition holds, %i.e.
%\ban
%\inf_{k\ge 0}\(\Lambda_k^h-\sum_{i=kh}^{(k+1)h-%1}\lambda^{\a}(i)\)\ge 0~\text{a.s.}
%\ean
%and
%\ban
%\sum_{k=0}^{\infty}\(\Lambda_k^h-\sum_{i=kh}^{(k+1)h-1}\lambda^{\a}(i)\)=\infty~\text{a.s.},
%\ean
then the algorithm $(\ref{Aalgorithm})$ converges almost surely by Theorem $\ref{ATheorem1}$.

%Lemma \ref{Alemma6}.A is proved.
Next, we will prove (II) of Lemma \ref{Alemma6}. By (\ref{Alxsffff}) and Lemma $\ref{Alemma4}$, there exists a positive integer $k_0$, such that
  \bna\label{Afiiwjffwrr}
\E[V((k+1)h)|\F(kh-1)]\leq (1+\Omega(k))V(kh)-2G(k)V(kh)+\Gamma(k)~\text{a.s.},~k\ge k_0,
  \ena
where $\Omega(k)+\Gamma(k)=\mathcal O(a^2(kh)+b^2(kh)+\lambda(kh))$. Noting that $G(k)=o(1)$ and $\Omega(k)=o(G(k))$, without loss of generality, we  suppose tha $0< \Omega(k)\leq G(k)<1$, $k\ge k_0$. Taking mathematical expectation on both sides of (\ref{Afiiwjffwrr}), we get
$
  \E[V((k+1)h)]\leq (1-G(k))\E[V(kh)]+\Gamma(k),~k\ge k_0.
$
On one hand, we know that $\Gamma(k)=o(G(k))$ and $\sum_{k=0}^{\infty}G(k)=\infty$, which together with Lemma A.3 in \cite{Alw} gives
  \bna\label{Aopopoop}
  \lim_{k\to\infty}\E[V(kh)]=0.
  \ena
On the other hand, by Lemma \ref{Alemma2}, we know that there exists a positive integer $k_1$, such that
\ban
\mathbb{E}[V(k)]\leq 2^h\mathbb{E}[V(m_kh)]+h2^h\gamma(m_kh),~k\ge k_1,
\ean
where $\gamma(k)=o(1)$ and $m_k=\lfloor \frac{k}{h}\rfloor$. Noting that $0\leq k-m_kh<h$, by (\ref{Aopopoop}), we get
$
  \lim_{k\to\infty}\E[V(k)]=0.
$
\end{proof}

\end{appendices}

\end{CJK}

\begin{thebibliography}{99}
  \bibitem{AEvgeniou} T. Evgeniou, M. Pontil, and T. Poggio, ``Regularization networks and support vector machines,'' \emph{Advances in Computational Mathematics}, vol. 13, no. 1, pp. 1-50, Apr. 2000.
  \bibitem{AGirosi} F. Girosi, ``An equivalence between sparse approximation and support vector machines neural computation,'' \emph{Neural Computation}, vol. 10, no. 6, pp. 1455-1480, Aug. 1998.
  \bibitem{APoggio} T. Poggio and S. Smale, ``The Mathematics of learning: dealing with data,'' \emph{Notices of the American Mathematical Society}, vol. 50, no. 5, pp. 537-544, May 2003.
  \bibitem{ATheodoridis} S. Theodoridis, \emph{Machine learning: a Bayesian and optimization perspective}. London, UK: Elsevier, 2015.
  \bibitem{Axue2020} H. Xue and Z. Ren, ``Sketch discriminatively regularized online gradient descent classification,'' \emph{Applied Intelligence}, vol. 50, pp. 1367-1378, Jan. 2020.

  \bibitem{Azhou2008} N. Zhou, D. J. Trudnowski, J. W. Pierre, and  W. A. Mittelstadt, ``Electromechanical mode online estimation using regularized robust RLS methods,'' \emph{IEEE Trans. Power Systems}, vol. 23, no. 4, pp. 1670-1680, Nov. 2008.

  \bibitem{Asun2019} Y. Sun, B. Wohlberg, and  U. S. Kamilov, ``An online plug-and-play algorithm for regularized image reconstruction,'' \emph{IEEE Trans. Computational Imaging}, vol. 5, no. 3, pp. 395-408, Sep. 2019.
  \bibitem{Aivanov2002} V. K. Ivanov, V. V. Vasin, and V. P. Tanana, \emph{Theory of linear ill-posed problems and its applications}, Berlin, Germany: Walter de Gruyter, 2002.

  \bibitem{AVito} E. D. Vito, L. Rosasco, A. Caponnetto, U. D. Giovannini, F. Odone, and P. Bartlett, ``Learning from examples as an inverse problem,'' \emph{J. Machine Learning Research}, vol. 6, no. 5, pp. 883-904, May 2005.

  \bibitem{ACesa} N. Cesa-bianchi, P. M. Long, and M. K. Warmuth, ``Worst-case quadratic loss bounds for prediction using linear functions and gradient descent,'' \emph{IEEE Trans. Neural Networks}, vol. 7, no. 3, pp. 604-619, May 1996.

  \bibitem{AGail2019} P. Gaillard, S. Gerchinovitz, M. Huard, and G. Stoltz, ``Uniform regret bounds over $\mathbb {R}^ d $ for the sequential linear regression problem with the square loss,'' in \emph{Proc. 30th Int. Conf. Algorithmic Learning Theory}, Chicago, USA, Mar. 2019, pp. 404-432.
  \bibitem{AGerch} S. Gerchinovitz, ``Sparsity regret bounds for individual sequences in online linear regression,'' \emph{J. Machine Learning Research}, vol. 14, no. 1, pp. 729-769, Mar. 2013.
  \bibitem{AJamil} W. Jamil and A. Bouchachia, ``Competitive regularised regression,'' \emph{Neurocomputing}, vol. 390, pp. 374-383, May 2020.
  \bibitem{AThramp} C. Thrampoulidis, S. Oymak, and B. Hassibi, ``Regularized linear regression: a precise analysis of the estimation error,'' in \emph{Proc. 28th Conf. Learning Theory}, Jun. 2015, pp.1683-1709.

  \bibitem{Ascaman} K. Scaman, F. Bach, S. Bubeck, L. Massoulie, and Y. T. Lee, ``Optimal algorithms for non-smooth distributed optimization in networks,'' in \emph{Proc. 32nd Conf. Neural Information Processing Systems}, Montr$\acute{\text{e}}$al, Canada, Dec. 2018, pp. 2745-2754.
  \bibitem{Apredd} J. B. Predd, S. R. Kulkarni, and H. V. Poor, ``A collaborative training algorithm for distributed learning,'' \emph{IEEE Trans. Information Theory}, vol. 55, no. 4, pp. 1856-1871, Mar. 2009.
  \bibitem{ALLB} Y. Liu, J. Liu, and T. Basar, ``Differentially private gossip gradient descent,'' in \emph{Proc. 57th IEEE Conf. Decision and Control}, Miami, USA, Dec. 2018, pp. 2777-2782.
  \bibitem{AYSVQ} F. Yan, S. Sundaram, S. Vishwanathan, and Y. Qi, ``Distributed autonomous online leaming: Regrets and intrinsic privacy-preserving properties,'' \emph{IEEE Trans. Knowledge and Data Engineering}, vol. 25, no. 11, pp. 2483-2493, Nov. 2013.
  \bibitem{Akar2011} S. Kar and J M. F. Moura, ``Convergence rate analysis of distributed gossip (linear parameter) estimation: fundamental limits and tradeoff,'' \emph{IEEE J. Sel. Topics Signal Processing}, vol. 5, no. 4, pp. 674-690, Aug. 2011.
  \bibitem{Akar2012} S. Kar, J. M. F. Moura, and K. Ramanan, ``Distributed parameter estimation in sensor networks: Nonlinear observation models and imperfect communication,'' \emph{IEEE Trans. Information Theory}, vol. 58, no. 6, pp. 3575-3605, Jun. 2012.
  \bibitem{Akar20132} S. Kar and J. M. F. Moura, ``Consensus+innovations distributed inference over networks: Cooperation and sensing in  networked systems,'' \emph{IEEE Signal Processing Magazine}, vol. 30, no. 3, pp. 99-109, May 2013.
  \bibitem{Akar2013} S. Kar, J M. F Moura, and  H. V Poor, ``Distributed linear parameter estimation: asymptotically efficient adaptive strategies,'' \emph{SIAM J. Control and Optmization}, vol. 51, no. 3, pp. 2200-2229, May 2013.
    \bibitem{Asahu2016} A. K. Sahu, S. Kar, J. M. F. Moura, and H. V. Poor, ``Distributed constrained recursive nonlinear least-squares estimation: algorithms and asymptotics,'' \emph{IEEE Trans. Signal and Information Processing over Networks}, vol. 2, no. 4, pp. 426-441, Dec. 2016.
  \bibitem{Axiesiyu1} S. Xie and  L. Guo, ``Analysis of normalized least mean squares-based consensus adaptive filters under a general information condition,'' \emph{SIAM J. Control and Optimization}, vol. 56, no. 5, pp. 3404-3431, Sep. 2018.
  \bibitem{Axiesiyu2} S. Xie and  L. Guo, ``Analysis of distributed adaptive filters based on diffusion strategies over sensor networks,'' \emph{IEEE Trans. Automatic Control}, vol. 63, no. 11, pp. 3643-3658, Nov. 2018.
  \bibitem{Asahu2018} A. K. Sahu, D. Jakovetic, and S. Kar, ``$\mathcal{CIRF\varepsilon}$: A distributed random fields estimator,'' \emph{IEEE Trans. Signal Processing}, vol. 66, no. 18, pp. 4980-4995, Sep. 2018.
   \bibitem{Acheny} Y. Chen, S. Kar, and J. M. F. Moura, ``Resilient distributed estimation: Sensor attacks,'' \emph{IEEE Trans. Automatic Control}, vol. 64, no. 9, pp. 3772-3779, Sep. 2019.
  \bibitem{Aydm} D. Yuan, A. Proutiere, and G. Shi, ``Distributed online linear regressions,'' \emph{IEEE Trans. Information Theory}, vol. 67, no. 1, pp. 616-639, Oct. 2020.
  \bibitem{Awjx} J. Wang, T. Li, and X. Zhang, ``Decentralized cooperative online estimation with random observation matrices, communication graphs and time delays,'' \emph{IEEE Trans. Information Theory}, vol. 67, no. 6, pp. 4035-4059, Jun. 2021.
  \bibitem{Alw} T. Li and J. Wang, ``Distributed averaging with random network graphs and noises,'' \emph{IEEE Trans. Information Theory}, vol. 64, no. 11, pp. 7063-7080, Nov. 2018.
  \bibitem{AZYD} Z. Zhang, Y. Zhang, D. Guo, S. Zhao, and X. L. Zhu, ``Communication-efficient federated continual learning for distributed learning system with Non-IID data,'' \emph{SCIENCE CHINA Information Sciences}, vol. 66, no. 2, pp. 122102, Dec. 2022.
  \bibitem{Aguolei} L. Guo, ``Estimating time-varying parameters by Kalman filter based algorithm: Stability and convergence,'' \emph{IEEE Trans. Automatic Control}, vol. 35, no. 2, pp. 141-147, Feb. 1990.
  \bibitem{AZJF} J. F. Zhang, L. Guo, and H. F. Chen, ``$L_p$-stability of estimation errors of Kalman filter for tracking time-varying parameters,'' \emph{Int. J. Adaptive Control and Signal Processing}, vol. 5, no. 3, pp. 155-174, May 1991.
\bibitem{AASBedi} A. S. Bedi, A. Koppel, and K. Rajawat, ``Asynchronous saddle point algorithm for stochastic optimization in heterogeneous networks,'' \emph{IEEE Trans. Signal Processing}, vol. 67, no. 7, pp. 1742-1757, Apr. 2019.
  \bibitem{ARDixit} R. Dixit, A. S. Bedi, and K. Rajawat, ``Online learning over dynamic graphs via distributed proximal gradient algorithm,'' \emph{IEEE Trans. Automatic Control}, vol. 66, no. 11, pp. 5065-5079, Nov. 2021.
  \bibitem{Azzjf} Q. Zhang and J. F. Zhang, ``Distributed parameter estimation over unreliable networks with markovian switching topologies,'' \emph{IEEE Trans. Automatic Control}, vol. 57, no. 10, pp. 2545-2560, Oct. 2012.
  \bibitem{Azhz} J. Zhang, X. He, and D. Zhou, ``Distributed filtering over wireless sensor networks with parameter and topology uncertainties,'' \emph{Int. J. Control}, vol. 93, no. 4, pp. 910-921, Apr. 2020.
  \bibitem{Awood}
  A. J. Wood and B. F. Wollenberg, \emph{Power Generation, Operation, and Control}. New York, NY, USA: Wiley, 2012.
%  \bibitem{bollobas} B. Bollobas, \emph{Modern Graph Theory}. New York, USA: Springer-Verlag, 1998.
   \bibitem{ADja}
    S. Djaidja and Q. Wu, ``An overview of distributed consensus of multi-agent systems with measurement/communication noises,'' in \emph{Proc. 34th Chin. Control Conf.}, Hangzhou, China, Jul. 2015, pp. 7285-7290.
    \bibitem{ADima}
    D. V. Dimarogonas and K. H. Johansson, ``Stability analysis for multi-agent systems using the incidence matrix: Quantized communication and formation control,'' \emph{Automatica}, vol. 46, no. 4, pp. 695-700, Apr. 2010.
    \bibitem{Ajwang}
    J. Wang and N. Elia, ``Mitigation of complex behavior over networked systems: Analysis of spatially invariant structures,'' \emph{Automatica}, vol. 49, no. 6, pp. 1626-1638, Jun. 2013.
      \bibitem{Asayed1}
  F. S. Cattivelli and A. H. Sayed, ``Diffusion LMS strategies for distributed estimation,'' \emph{IEEE Trans. Signal Processing}, vol. 58, no. 3, pp. 1035-1048, Mar. 2009.
\bibitem{Apiggott}
M. J. Piggott and V. Solo, ``Diffusion LMS with correlated regressors II: Performance,'' \emph{IEEE Trans. Signal Processing}, vol. 65, no. 15, pp. 3934-3947, Aug. 2017.
  \bibitem{ALeonard}
  N. E. Leonard and A. Olshevsky,  ``Cooperative learning in multiagent systems from intermittent measurements,'' \emph{SIAM J. Control and Optimization}, vol. 53, no. 1, pp. 1-29, Jan. 2015.
  \bibitem{Arobbinssiegmund}
  H. Robbins and D. Siegmund, ``A convergence theorem for non negative almost supermartingales and some applications,'' in \emph{Selected Papers}, T. L. Lai, and D. Siegmund, Eds., New York, USA: Springer-Verlag, 1985.
%  \bibitem{AHuang}
%  M. Huang, S. Dey, G. Nair and J. Manton, ``Stochastic consensus over noisy networks with Markovian and arbitrary switches,'' \emph{Automatica}, vol. 46, no. 10, pp. 1571-1583, 2010.
  %\bibitem{ALong}
%  Y. Long, S. Liu and L. Xie, ``Distributed consensus of discrete-time multi-agent systems with multiplicative noises,'' \emph{Int. J. Robust Nonlinear Control}, vol. 25, no. 16, pp. 3113-3131, 2014.
%  \bibitem{ANi}
%  Y. Ni and X. Li, ``Consensus seeking in multi-agent systems with multiplicative measurement noises,'' \emph{Systems \& Control Letters}, vol. 62, no. 5, pp. 430-437, 2013.
\end{thebibliography}
\end{document}